\documentclass[12pt]{msml2021} 

\title[]{Phase Retrieval with Holography and Untrained Priors: Tackling the Challenges of Low-Photon Nanoscale Imaging}
\usepackage{times}

\usepackage{amsmath,amsfonts,bm}

\def\eqref#1{equation~\ref{#1}}

\def\1{\bm{1}}

\def\eps{{\epsilon}}

\def\mB{{\bm{B}}}

\def\mR{{\bm{R}}}

\def\mX{{\bm{X}}}
\def\mY{{\bm{Y}}}
\def\mZ{{\bm{Z}}}

\DeclareMathAlphabet{\mathsfit}{\encodingdefault}{\sfdefault}{m}{sl}
\SetMathAlphabet{\mathsfit}{bold}{\encodingdefault}{\sfdefault}{bx}{n}

\newcommand{\R}{\mathbb{R}}

\usepackage{hyperref}
\usepackage{xcolor}
\usepackage{url}
\usepackage{graphicx} 
\usepackage{paperstyle}

\usepackage[normalem]{ulem}

\newcommand{\changes}[1]{{#1}}

\msmlauthor{%
 \Name{Hannah Lawrence*$^{1}$} \Email{hanlaw@mit.edu}\\
 \Name{David A. Barmherzig*$^2$} \Email{dbarmherzig@flatironinstitute.org}\\
 \Name{Henry Li$^{3}$} \Email{henry.li@yale.edu}\\
 \Name{Michael Eickenberg$^2$} \Email{meickenberg@flatironinstitute.org}\\
 \Name{Marylou Gabrié$^{\dagger 2,4}$} \Email{mgabrie@flatironinstitute.org}\\
 \addr 1. Massachusetts Institute of Technology, Cambridge, Massachusetts, USA\\
 \addr 2. Center for Computational Mathematics, Flatiron Institute, New York, New York, USA\\
 \addr 3. Program in Applied Mathematics, Yale University, Connecticut, USA \\
 \addr 4. Center for Data Science, New York University, New York, New York, USA\\
 \addr $\dagger$ Corresponding author,
 \addr * These authors contributed equally
}

\newcommand{\latentim}{z}

\makeatletter
 \let\Ginclude@graphics\@org@Ginclude@graphics
\makeatother
\begin{document}
\addtolength{\textfloatsep}{-0.3in}
\maketitle

\begin{abstract}
Phase retrieval is the inverse problem of recovering a signal from magnitude-only Fourier measurements, and underlies numerous imaging modalities, such as Coherent Diffraction Imaging (CDI). A variant of this setup, known as holography, includes a reference object that is placed adjacent to the specimen of interest before measurements are collected. The resulting inverse problem, known as holographic phase retrieval, is well-known to have improved problem conditioning relative to the original. This innovation, i.e. Holographic CDI, becomes crucial at the nanoscale, where imaging specimens such as viruses, proteins, and crystals require low-photon measurements. This data is highly corrupted by Poisson shot noise, and often lacks low-frequency content as well. In this work, we introduce a dataset-free deep learning framework for holographic phase retrieval adapted to these challenges. The key ingredients of our approach are the explicit and flexible incorporation of the physical forward model into an automatic differentiation procedure, the Poisson log-likelihood objective function, and an optional untrained deep image prior. We perform extensive evaluation under realistic conditions. Compared to competing classical methods, our method recovers signal from higher noise levels and is more resilient to suboptimal reference design, as well as to large missing regions of low frequencies  in the observations. 
\changes{Finally, we show that these properties carry over to experimental data acquired on optical wavelengths.}
To the best of our knowledge, this is the first work to consider a dataset-free machine learning approach for holographic phase retrieval.

\end{abstract}
\section{Introduction}
\label{sec:intro}

\begin{figure} [!htbp]  
     \centering
     {\includegraphics[width=\textwidth]{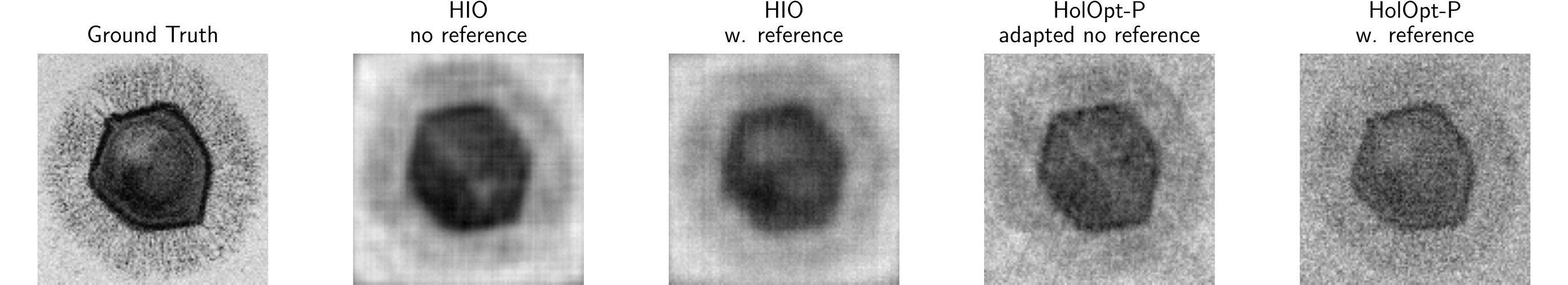}}
     \caption{{The advantage of using a reference 
     for phase retrieval at $N_p = 1$ photon/pixel.
      Two algorithms, HIO and our \hp, are applied 
      to reconstruct 
      from Fourier magnitude data of
      image alone (classical phase retrieval), and image with reference (holographic phase retrieval). The setting and reference are given on Figure \ref{subfig:FH-CDI-B}.
     Comparing between algorithms, observe the
     higher-quality reconstruction when the reference is present.
     (VIRUS image courtesy of \cite{Mimivirus})} \label{fig:summary} }
 \end{figure}

Phase retrieval is a nonlinear inverse problem that arises ubiquitously in imaging sciences, and has gained much recent attention ~\citep{eldar-review}. In this work we focus on a practical instance of the problem 
that arises in Coherent Diffraction Imaging (CDI). Here,
\textit{holographic phase retrieval}
consists of 
recovering an image $\mX_0\in\mathbb R^{m\times n}$ from a set of squared Fourier transform magnitudes
\begin{equation}
   \mY = \left|F(\mX_0 + \mR_0)\right|^2,
\end{equation}
where \(F\) denotes an oversampled Fourier transform operator and $\mR_0\in\mathbb R^{m\times n}$ is a known reference image whose support does not intersect the support of $\mX_0$. 
The known reference image $\mR_0$ distinguishes holographic phase retrieval from the (classical) phase retrieval setting, where the goal is to recover $\mX_0$ from $|F(\mX_0)|^2$ alone. 
We
focus 
entirely
on the holographic version of the problem in realistic conditions: with high noise levels and missing low-frequency data. The advantage provided by the
 holographic
 reference is briefly illustrated 
on
 Figure \ref{fig:summary}.

In the remainder of this introduction we will situate this problem in the context of Coherent Diffraction Imaging, review related works, and list our contributions.
Section \ref{sec:setup} describes our setup in detail, and Section \ref{sec:strategy} our reconstruction strategy.
In Section \ref{sec:experiments}, we describe extensive experiments 
and 
compare to
several baseline methods. \changes{Section \ref{sec:realdata} presents a validation in an optical laser CDI experiment.}
Section \ref{sec:discussion} concludes with a general discussion.

\subsection{Holographic Coherent Diffraction Imaging and phase retrieval}
\label{sec:CDI-and-PR}
\begin{figure} 
    \centering
    \subfigure[][Experimental setup.]
    {\includegraphics[trim = 10 0 10 10,  clip,width=0.45\textwidth]{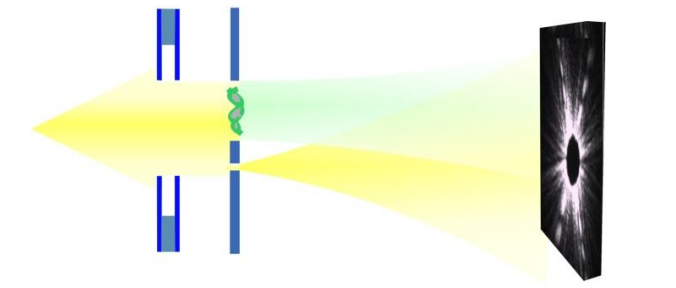}\label{subfig:FH-CDI-A}}
    \hspace{0.01\textwidth}
    \subfigure[][The zero-padded data and its measured Fourier magnitudes.]
    {\includegraphics[trim = 100 20 100 30,  clip, width=0.3\textwidth]{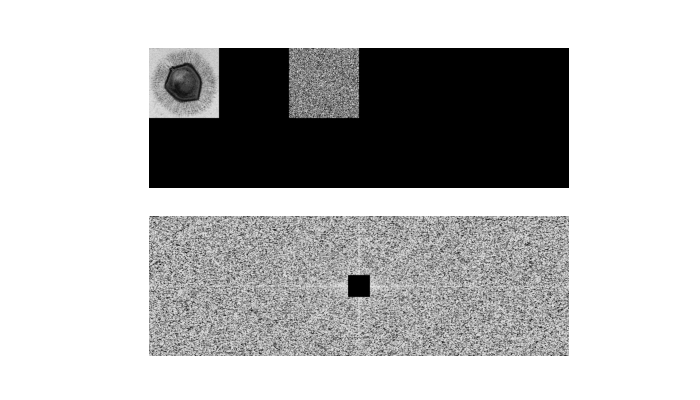}\label{subfig:FH-CDI-B}} 
    \caption{{Holographic CDI schematic. The upper portion of the diffraction area contains the specimen of interest $\mX_0$, and the adjacent portion consists of a known ``reference'' $\mR_0$. The recorded data $\mY$ has its low frequencies occluded by a beamstop. (Image courtesy of~\cite{FT-Cambridge}.\label{fig:FH-CDI})}}
\end{figure}

Coherent Diffraction Imaging (CDI) is a scientific imaging technique used for resolving nanoscale scientific specimens, such as viruses, proteins, and crystals~\citep{CDI-orig}. In CDI, an image is sought to be reconstructed from X-ray diffraction measurements recorded on a CCD detector plane. By the
\textit{far-field approximation} of optical theory, these measurements are approximately proportional to the squared magnitude values of the Fourier transform of the electric field within the diffraction area. Thus, the specimen structure (e.g., its electron density) can be determined, in principle, by solving the phase retrieval problem.
Holographic CDI is a popular setup to perform CDI experiments in which 
the 
object undergoing diffraction physically consists of a specimen 
together with a ``reference'', i.e. a portion of the object 
\emph{a priori} known. This setup is illustrated in Figure \ref{fig:FH-CDI}. The inclusion of a reference in the CDI setup 
both 
enhances the quality of image reconstruction, and 
greatly simplifies the analysis and solution 
of the corresponding phase retrieval problem \citep{MarchesiniURA,HERALDO,BarmherzigEtAl2019Holographic}.

Nevertheless, holographic CDI remains challenging in practice.
Due to the quantum mechanical nature of photon emission and 
absorption, CDI measurements are inherently corrupted by \textit{Poisson shot noise}. The severity of this noise corruption is inversely proportional to
the
strength of the X-ray source in use, which is in turn quantified via $N_p$, the number of photons per pixel reaching the detector plane. Nanoscale applications of CDI often necessitate imaging in the \textit{low-photon regime}, where measurements are highly corrupted by noise. CDI measurements are also typically lacking low-frequency data, due to the presence of a \textit{beamstop} apparatus which occludes direct measurement of these values \citep{Cossairt2015,Tatiana2019,barmherzig2020recovering}.

The holographic phase retrieval problem is 
commonly
solved by
\textit{inverse filtering} \citep{Holog1,Holog2}, which 
amounts to solving a structured system of linear equations. While straightforward, this method is not well-suited for noisy data. \textit{Wiener filtering} ~\citep{Gorkhover2018} is a variant on this method with some denoising ability.
Yet Wiener filtering is derived to account for an \textit{additive} noise model --- an assumption which is not true for Poisson shot noise at low photon counts, and only holds as an approximation at high photon count levels ~\citep{CDI-Book}. 
Moreover, these methods do not account explicitly for missing low-frequency data, and require a minimum separation
between the specimen and the reference objects, the \textit{holographic separation condition} (see Section \ref{sec:sep-exp}).
The most popular algorithm 
for the classical phase retrieval problem is the
Hybrid Input-Output (HIO) algorithm ~\citep{FienupHIO}, which is based on an alternating projection scheme. This method can be modified to the holographic setting by adding an additional projection step to enforce the reference constraints, which greatly improves the algorithm's performance \cite{BarmherzigEtAl2019Holographic}.

\subsection{Related work}
\label{sec:related}

\textbf{Machine learning for inverse problems}
Increasing
research effort has been devoted to addressing inverse problems, even beyond phase retrieval, with deep learning approaches (see \cite{Ongie2020} for a recent review). 
Supervised strategies can be broadly divided into four main categories: 
end-to-end methods, (e.g. \cite{McCann2017}), ``unrolling" algorithms (e.g. \cite{Meinhardt2017}),
pretrained image denoisers (e.g. \cite{Romano2017}), and learned generative models as highly informative priors (e.g. \cite{Tramel2016}).
All of 
these
approaches require a training set, containing either matched signal-observation pairs or simply typical signals, 
with a large number of data points.
The reconstruction improves drastically when this information is available. 
However,
it is often unrealistic to assume such prior knowledge on the measured signal.
Recently,
for the reconstruction of images, it was found that untrained generative neural networks with appropriate architectures can still be efficient priors 
\citep{Ulyanov2018,Heckel2018}. 
By adjusting their 
parameters
to fit a single output observation,
they do not require a training set
, and instead encourage naturalistic images due to architecture alone.
 In this paper we focus on the last approach, as it is 
 widely applicable 
 to image data
 and 
satisfies our requirement of applicability in a realistic setting.

\textbf{Machine learning for phase retrieval }
More specifically, several variants of the phase retrieval problem have received 
attention in the context of machine learning for inverse problems.
Compressive Gaussian phase retrieval, where one observes the amplitude of random complex Gaussian projections of the signal, is a popular setting in the machine learning community. 
It is 
easier than the Fourier phase retrieval problem and often more amenable to theoretical analysis (see e.g. \cite{Aubin2020}). 
For this version of the problem, trained generative models such as Generative Adversarial Networks \citep{Shamshad2018, Hand2018}, as well as untrained priors \citep{Jagatap2019}, were found to be very effective on machine learning toy 
datasets. 
An increasing number of works now consider the more realistic
problem of 
Fourier phase retrieval. Using pre-trained Gaussian denoisers and iterative algorithms, Deep-prior-based sparse representation \citep{Shi2020}, prDeep \citep{Metzler2018} and Deep-ITA \citep{Wang2020}, are solutions robust to noise 
in the case where
the corruption is small enough to be approximately Gaussian. The end-to-end solution investigated by \cite{Uelwer2019} features some robustness to Poisson shot noise, but struggles to generalize to complicated datasets. Meanwhile, the ``physics-informed" architecture of \cite{Goy2018}, 
which includes
information about the data generating process, is shown to perform well on 
realistic
signals at very low photon counts, but requires a few thousand training examples. Closer to our work, \cite{Wang2020a} proposed a U-net and \cite{Bostan2020} tested the 
deep decoder,
both untrained neural networks, for Fourier phase retrieval,
but did not consider the holographic setting.
%
To the best of our knowledge, \cite{Rivenson2018} is the only proposition considering holography, showing that deep neural networks trained end-to-end
on a dataset of a few hundred images
lead to state-of-the-art performance. 
%

\textbf{Other optimization approaches}
The use of auto-differentiation for Fourier phase retrieval was initiated by \cite{Jurling2014}, while the convenience of deep learning packages was exploited later \citep{Nashed2017, Kandel2019}. 
\cite{Thibault2012} derived conjugate gradients to optimize the likelihood for classical (non-holographic) CDI, following either the Poisson or the Euclidean metric.
Recently, \cite{Barmherzig20LowPhoton} 
pointed at the potential of likelihood optimization for holographic phase retrieval.

\subsection{Our contributions}
We 
address the holographic phase retrieval problem in the low-photon regime using a Poisson maximum likelihood framework and recent insights in machine learning for inverse problems.  
Our strategy combines three key ideas: 
(i) a realistic physical 
noise model for CDI,
(ii) 
auto-differentiation and efficient optimization 
readily available 
in a package like PyTorch \citep{pytorch2019}, 
and lastly 
(iii) the option to add a neural network prior. 
%
We 
(a) compare these methods to baselines, exploring several experiment challenges;
(b) demonstrate significant improvements at different noise levels;
(c) investigate the impact of missing low-frequency data on our methods, and show that ours are more robust than baseline methods;
%
(d) investigate the impact of the distance between object image and reference image on reconstruction quality,
showing
that our proposed method can easily deal with distances below the \textit{holographic separation condition}; 
(e) vary the oversampling rate of observation, observing a graceful degradation of reconstruction with decreasing samples;
\changes{and (f) perform a comparison of our different methods on an experimental data set in the optical range.}
Finally, we 
(g) provide a Python package\footnote{anonymous URL to download \href{https://drive.google.com/file/d/1gRxdVgjbV-HEJd-SiU4eXRFwJVEmVn1K/view?usp=sharing}{code} during the review process} to run our implementation.

\section{Holographic CDI Setup}
\label{sec:setup}

The data generation process mimics the key components of a holographic diffraction experiment
    as realistically as possible,
    namely by including
    two crucial ingredients:
    the Poisson shot noise model and
    the beamstop occluding low-frequency measurements.

\textbf{Coherent diffraction imaging }
Let $\mZ\in\R^{m\times n}$ represent a real $m\times n$-pixel image.
As explained in Section \ref{sec:CDI-and-PR}, the recorded CDI measurements can be approximated by the square of the oversampled Fourier transform magnitude of the object image. Here, we assume an oversampling factor of two, which is the minimum oversampling factor theoretically required for perfect reconstruction in the noiseless setting ~\citep{Hayes}.
%
%
Let $F: \R^{m\times n}\rightarrow \mathbb{C}^{2m\times 2n}$ be the doubly oversampled discrete Fourier transform operator.
    $F(\mZ)$ can be implemented as the discrete Fourier transform of 
    a
    zero-padded version of $\mZ$.
        Let $\tilde \mZ = \left(\begin{array}{c|c}\mZ & \mathbf{0}\\\hline \mathbf{0}&\mathbf{0}\end{array}\right)\in\R^{2m\times 2n}$.
        Then let $F(\mZ) = \mathcal F(\tilde \mZ),$ where $\mathcal F$ is the discrete Fourier transform operator.
The intensity distribution at the detector is defined as $I(\mZ) = |F(\mZ)|^2$ (where the absolute value here is understood in the pointwise sense).

\textbf{Beam stopping }
A beam stopping mask, or ``beamstop", is defined as $\mB\in\{0, 1\}^{2m\times 2n}$ such that it takes the value 0 in a region 
of low
frequency and 1 everywhere else.
The beam-stopped intensity image can then be written as $I(\mZ)\odot \mB,$ where $\odot$ represents pointwise multiplication.
%

\textbf{Measurement process } Let $N_p > 0$ represent the expected mean number of photons incident per detector pixel.
    Then $(2m\times 2n)N_p$ is the expected total number of photons incident on the detector.
The measurement data vector $\mY\in\R^{2m\times 2n}$ is set to 
\begin{align}
    \label{eq:obs-poisson}
    \mY\sim\frac{C}{N_p}\mathrm{Poisson}\left(\frac{N_p}{C}I(\mZ)\odot \mB\right).
\end{align}

The constant $C$ is equal to the sum of all 
square Fourier magnitudes
over the detector.
The inner normalization constant $N_p/C$ ensures that the simulated setting corresponds on average to the measurement of $N_p$ expected photons per pixel. 
The outer normalization constant is applied to make $\mY$ be of the same order of magnitude as $I(\mZ)\odot \mB$
.

\textbf{Holography setup }
We structure
$\mZ$
    into an unknown object $\mX$,
        and
        a known reference $\mR$.
The setting we will use throughout this paper 
is as follows (see also Figure \ref{fig:FH-CDI}).
Let $\mX, \mR\in \R^{m\times m}$, and set $\mX_0=(\mX|\mathbf{0}_{m\times m}|\mathbf{0}_{m\times m}), \mR_0=(\mathbf{0}_{m\times m}|\mathbf{0}_{m\times m}|\mR)$ with $\mX_0, \mR_0\in\R^{m\times 3m}$. Then $\mZ = \mX_0 + \mR_0 = (\mX|\mathbf{0}_{m\times m}|\mR)$.
The region of zeros $\mathbf{0}_{m\times m}$ separating object and reference represents the \textit{holographic separation condition}. It is not necessary for our proposed methods (see Section \ref{sec:sep-exp})
, but required
for several baseline methods.
Thus to ensure a fair comparison, the separation setting will be our standard setting.

\section{A reconstruction strategy adapted to low-photon CDI}

\label{sec:strategy}
\subsection{\hp~and \hpdd}
We
propose to maximize the likelihood of the measurements $\mY$ given the underlying image $\mX$ and the CDI model above. This objective involves the likelihood of the Poisson-distributed measurements, 
accounting for the nature of noise in the low-photon regime, as well as the full forward model (including reference and beamstop): $\hat\mX = \underset{\mX}{\arg\max} \log p(\mY | \mX, N_p)$,
where the distribution of $\mY$ conditional on $\mX$ is given by Equation \ref{eq:obs-poisson}. Replacing the expression of the Poisson distribution and 
dropping
constants yields 
\begin{align} \label{eq:poisson_loss}
    \hat\mX 
     = \underset{\mX}{\arg\max} \sum_{ij | (\mB)_{ij} = 1}
    Y_{ij} \log I\left((\mX|\mathbf{0}_{m\times m}| \mR) \right)_{ij} -  I((\mX|\mathbf{0}_{m\times m}| \mR))_{ij} 
\end{align} 
where the sum is taken 
only 
over non-zero entries of the beamstop mask. 
The optimization of this objective is performed directly using gradient ascent in PyTorch. 
We
investigate two strategies. The optimization is done either directly on the pixels of $\mX$, as in Equation \ref{eq:poisson_loss},
or on the parameters of a 
deep decoder
neural network prior encoding $\mX$ 
\citep{Heckel2018}. 
We refer to these two variants as \hp~for holographic Poisson likelihood optimization, and \hpdd~for holographic Poisson likelihood optimization with a deep decoder. 

\subsection{\changes{The deep decoder}}
\label{app:dd}
The deep decoder belongs to the class of untrained image priors:
neural networks
with image-shaped outputs trained by gradient descent
    to output one single image. 
The architecture of the network imposes an inductive bias favoring natural image statistics. 
\changes{In a sense, the architecture itself has been trained by decades of engineering in image processing.}

\textbf{Deep decoder architecture}
The deep decoder essentially consists of an alternation of two operations ---
    convolutions with filter size of $1\times 1$ pixels,
    and upsampling by a factor of 2 using bilinear interpolation.
The input is a randomly initialized image of smaller size.
In order to end up with a specific output image size,
    either the input image size or the number of layers $d$ is adjusted.  
\changes{It should be noted that pixels are spatially coupled only through the upsampling layers, while the $1\times 1$ convolutions are pixel-wise linear transforms shared among all pixels. This weight-sharing allows to reduce the number of free parameter to mitigate overfitting  \citep{Heckel2018}.
}
    
Let $c_i$ represent the number of channels at layer $i$ and
let $\vartheta_i\in\R^{c_{i+1}\times c_i\times 1\times 1}$ represent the convolution kernels at layer $i$.
Denote by $\mathrm{conv}_{\vartheta_i}$ the typical deep learning convolution with $\vartheta_i$,
by $\mathrm{up_2}$ the bilinear upsampling operation, and by $\mathrm{relu}(x) = x\1_{x > 0}(x)$.
Then we can define one component as
$$\mathrm{block}_i :=\mathrm{up}_2 \circ\mathrm{relu}\circ\mathrm{conv}_{\vartheta_i}$$
and the full network as
$$\mathrm{net}:= \mathrm{block}_d\circ\cdots\circ\mathrm{block}_1.$$
For an input image $\latentim\in\R^{1\times c_1\times \kappa\times \lambda}$, let
$$\sigma(\vartheta, \latentim) = \sigma(\vartheta_d, \dots, \vartheta_1, \latentim) = \mathrm {net}(\latentim),$$
where $\vartheta = (\vartheta_d, \dots, \vartheta_1)$ collects all the convolution parameters.
For \hpdd, we set $\mX = \sigma(\vartheta, \latentim)$ and train all $\vartheta_i$ in $\vartheta$.
Here the objective can be re-written as
\begin{align} \label{eq:poisson_loss_dd}
    \hat\vartheta = \underset{\vartheta}{\arg\max} \sum_{ij / (\mB)_{ij} = 1}
    Y_{ij} \log I((\sigma(\vartheta, \latentim)|\mathbf{0}_{m\times m}| \mR))_{ij} -  I((\sigma(\vartheta, \latentim)|\mathbf{0}_{m\times m}| \mR))_{ij} .
\end{align} 
The reconstructed image is then the output of the deep decoder $\hat\mX = \sigma(\hat{\vartheta}, \latentim)$ 
after
training on
a single magnitude image $\mY$.

\textbf{Deep decoder depth}\label{sec:dd-hyperparameters} 
\begin{figure} 
    \centering
    {\includegraphics[trim= 10 10 10 0 , clip, width=0.45\textwidth]{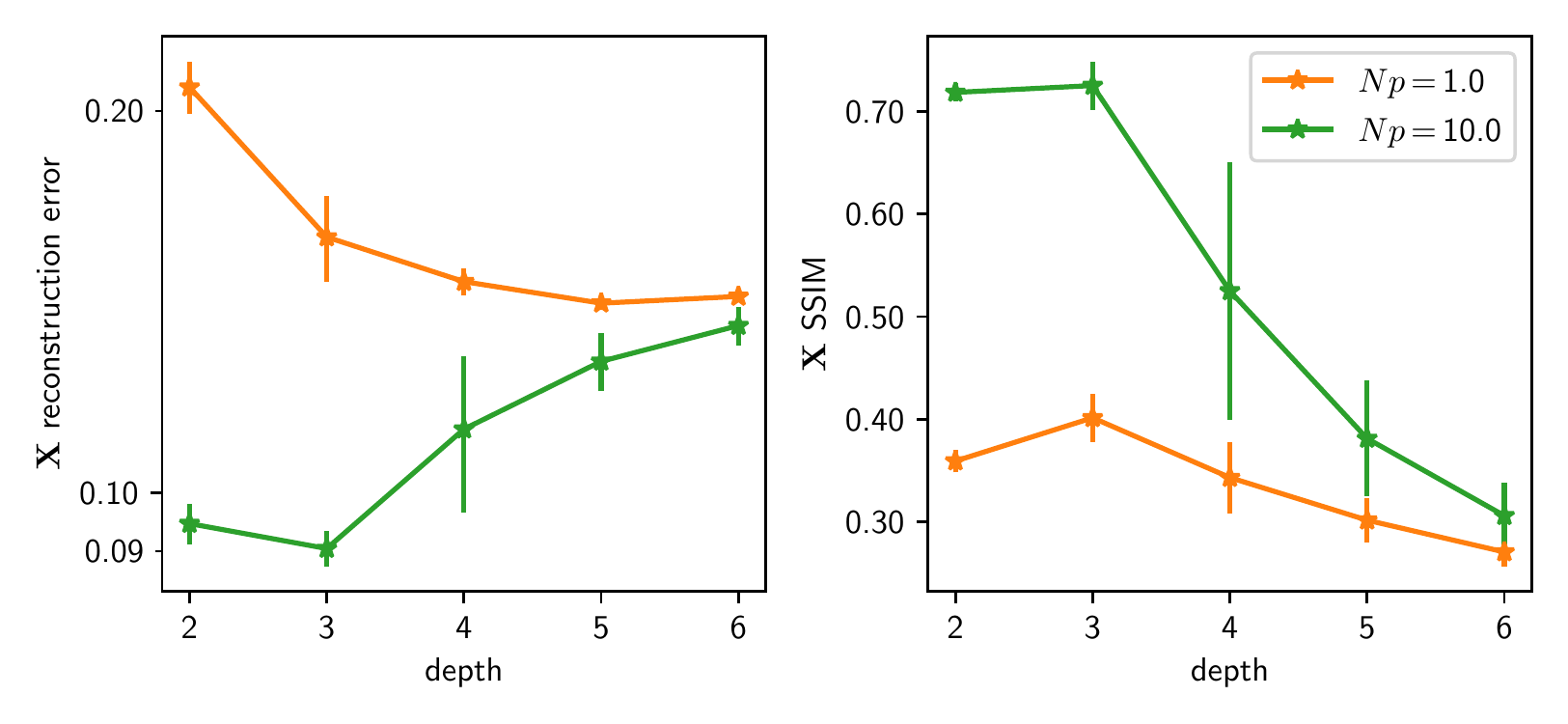}}
    \hspace{0.02\textwidth}
    {\includegraphics[trim= 0 -10 0 0 , clip,width=0.47\textwidth]{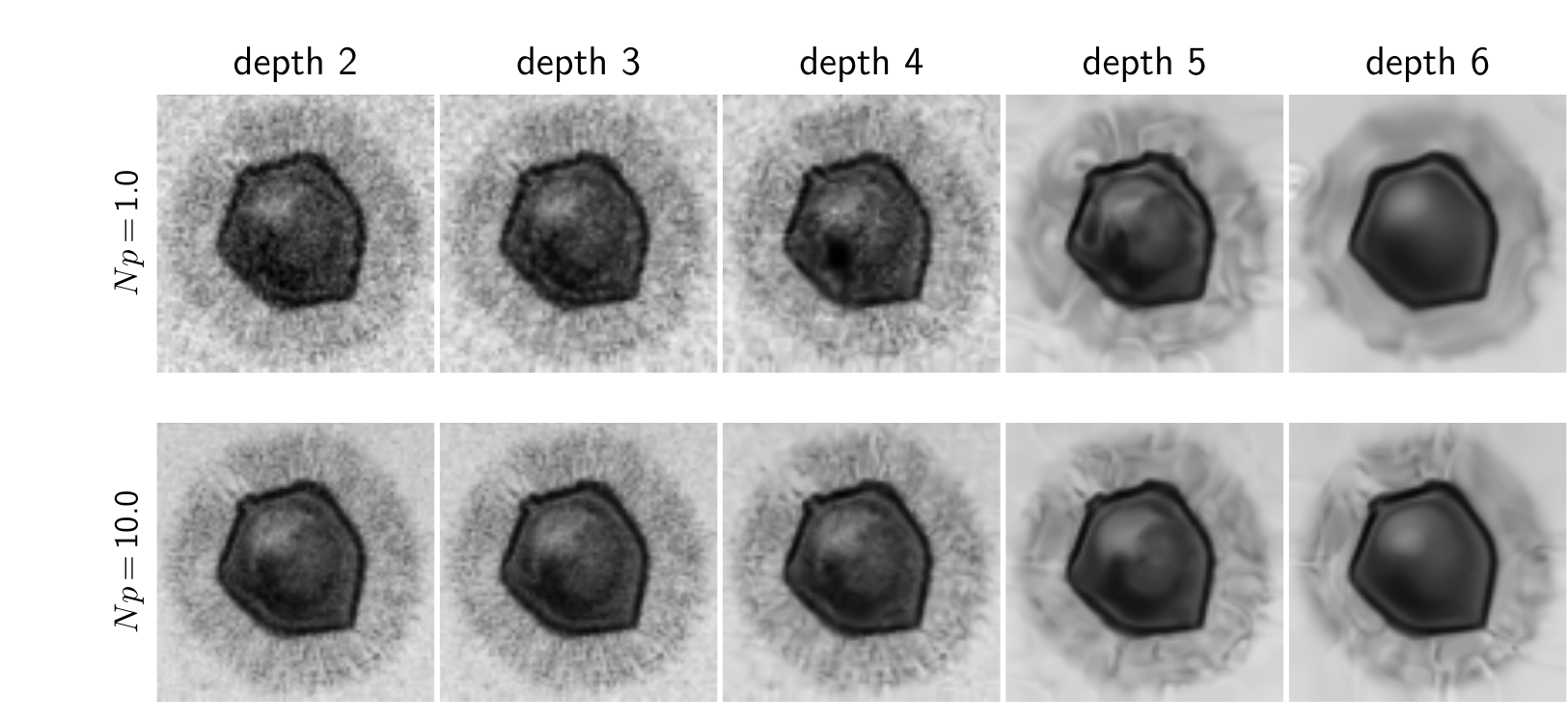}}
    \caption{\changes{Impact of depth of a deep decoder prior on the reconstruction illustrated on VIRUS. Left: Relative reconstruction mean squared errors and structure similarity indices with ground truth as a function of depth. Right: Best reconstruction out of 4 runs. \label{fig:depth-virus}}}
\end{figure}
\label{app:dd-param}
In our reconstruction experiments, we observe the number of channels of the convolutional filters to only marginally change the outcome of the reconstruction, while varying the depth of the prior trades off precision of the reconstructed edges and finer details 
(shallower) with spatial regularity (deeper).
To fine-tune a specific reconstruction it can be useful to adjust this parameter, e.g. to limit the fitting power of the model at high noise levels, as pointed out by \cite{Heckel2018}, yet its impact is much more subtle than that of depth. 
\changes{
We observe that the scaling of the distribution of the random input vector $\latentim$ does not have a significant impact on the reconstructions. Following the heuristic of \cite{Heckel2018},
$\latentim$ was drawn from a uniform distribution between $0$ and $0.1$.  Using different scalings seem to be compensated for by the training of convolutional layer weights, which were initialized with the PyTorch default.
}

In Figure \ref{fig:depth-virus}, we illustrate the impact of depth by reporting errors and reconstructed images of the VIRUS as a function of depth for different noise levels in the setting of the experiment presented in Section \ref{sec:denoising} below. Visually, deeper decoders render smoother images. This is a direct consequence of the upsampling layers which correlates the neighboring pixels. The deeper the decoder, the smaller the latent representation $\latentim$ and the less independent the output pixels. 
At high noise levels ($N_p = 1$) the smoothing reduces the reconstruction error estimated using Euclidean distance. 
Perceptually however, the smoothing is only beneficial to some extent, and a practitioner would likely prefer a shallow network. 
At lower noise levels ($N_p = 10$), the excess of depth can be spotted directly in the reconstruction error as we observe a dip in the curve.

\begin{figure} 
    \centering
    {\includegraphics[width=0.82\textwidth]{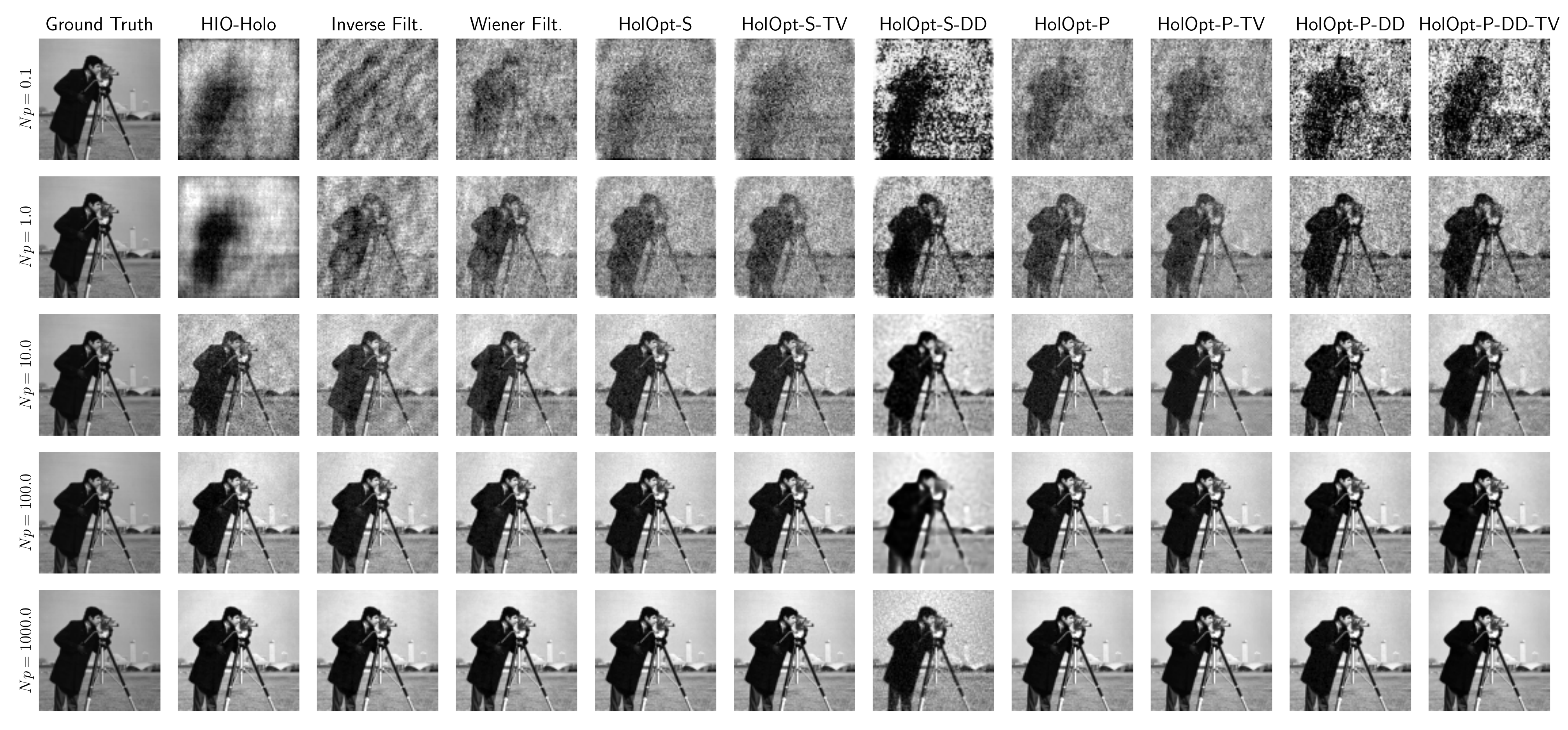}}
    \centering
    {\includegraphics[width=0.82\textwidth]{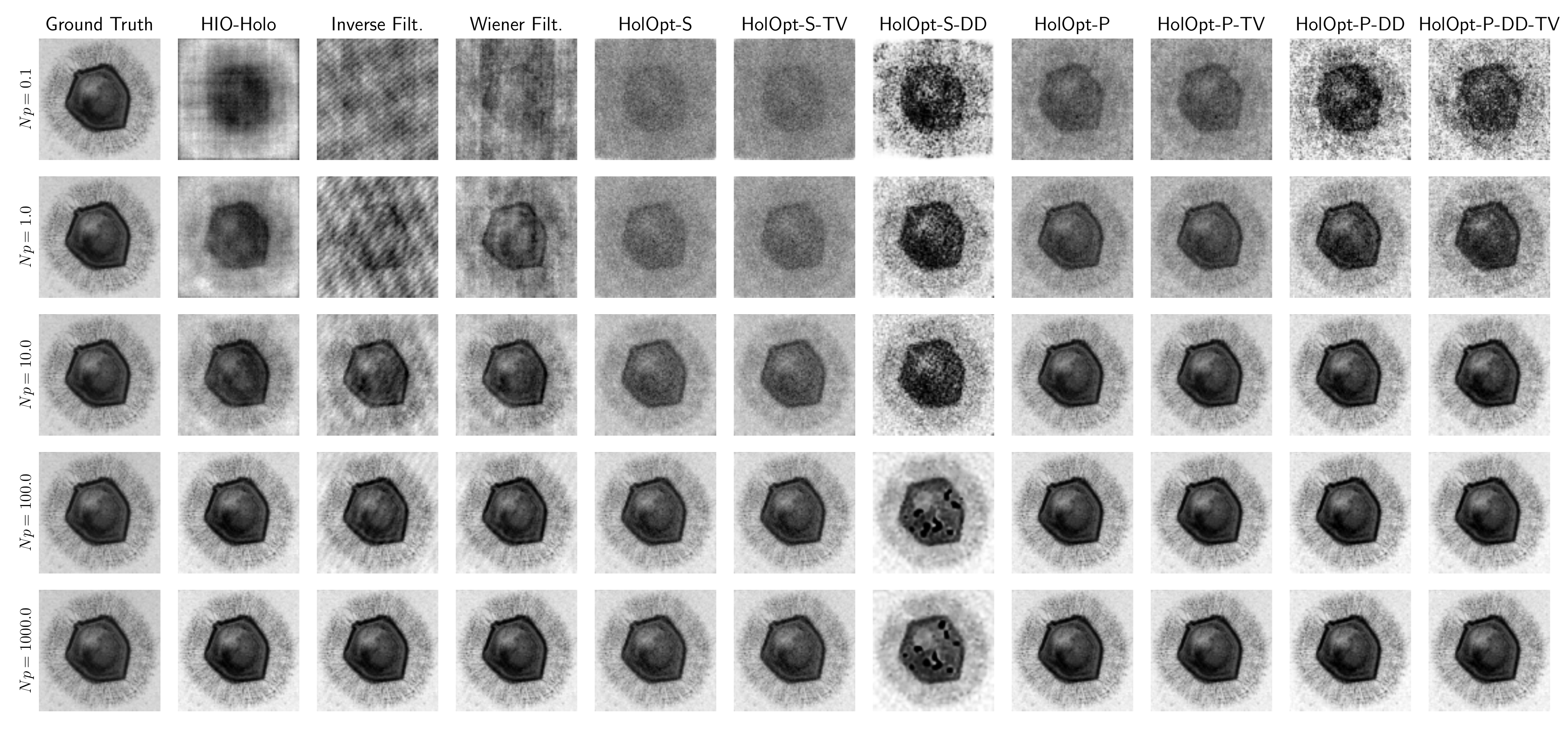}}
    \centering 
    {\includegraphics[width=0.82\textwidth]{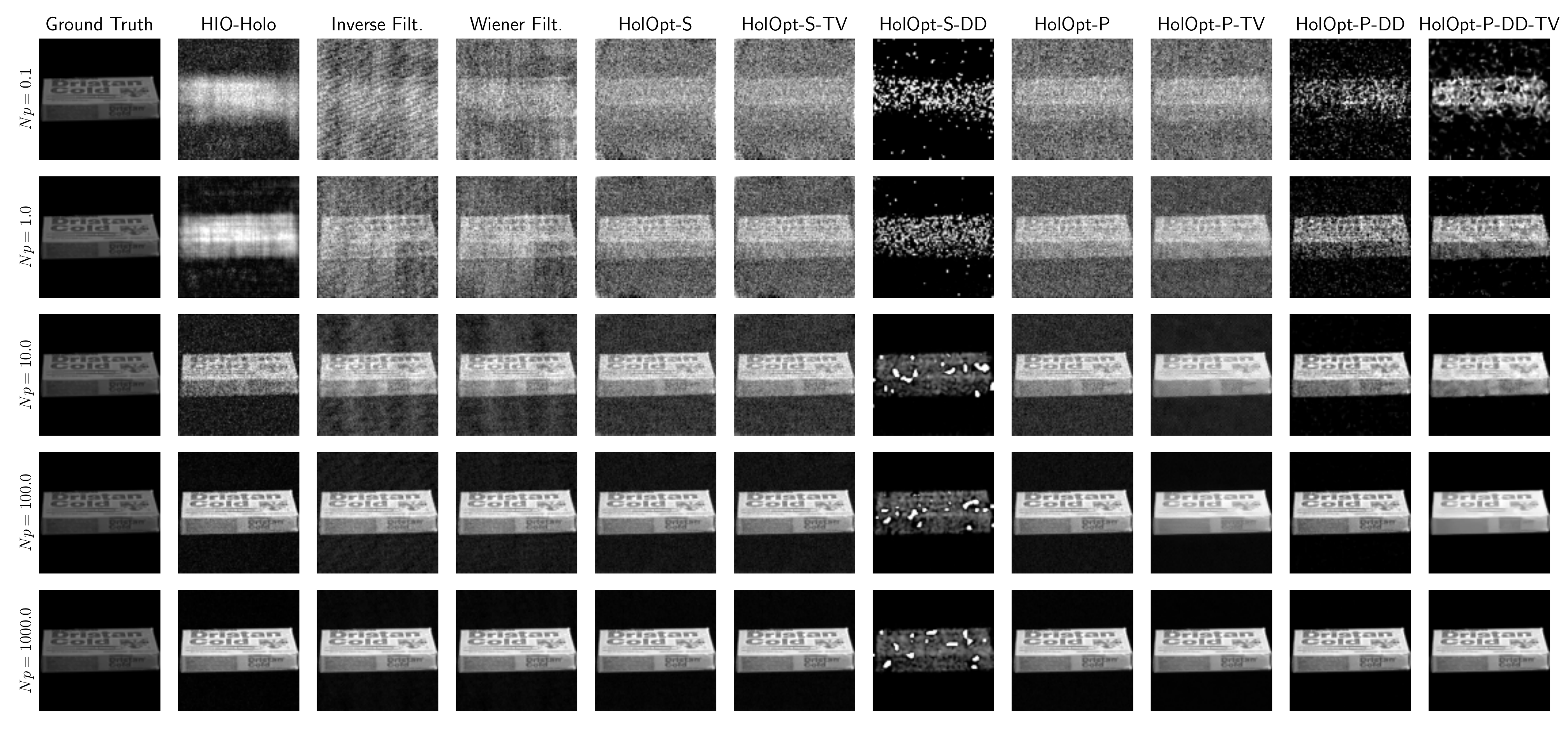}}
 \caption{ \label{fig:denoising-camera-visual}
    Comparing reconstructions across algorithms and noise levels on a sample image from SET12 (CAMERA, top), a sample image from BIO10 (VIRUS, middle) and a sample image from COIL100 (bottom) with a 
    binary random reference and without beamstop.
    Figure
    \ref{fig:denoising-coil100-visual} presents 3 more images. Corresponding SSIM scores can be found in Figure \ref{fig:denoising-graph-ssim}.
    \changes{To improve contrast black and white are set respectively to first and last (99th) percentile of all pixels within each image. This convention is adopted for all visuals in the paper.}
    \changes{Revision: Results with TV regularization where added.}}
\end{figure}

\begin{figure} 
    \centering
    {\includegraphics[width=0.9\textwidth]{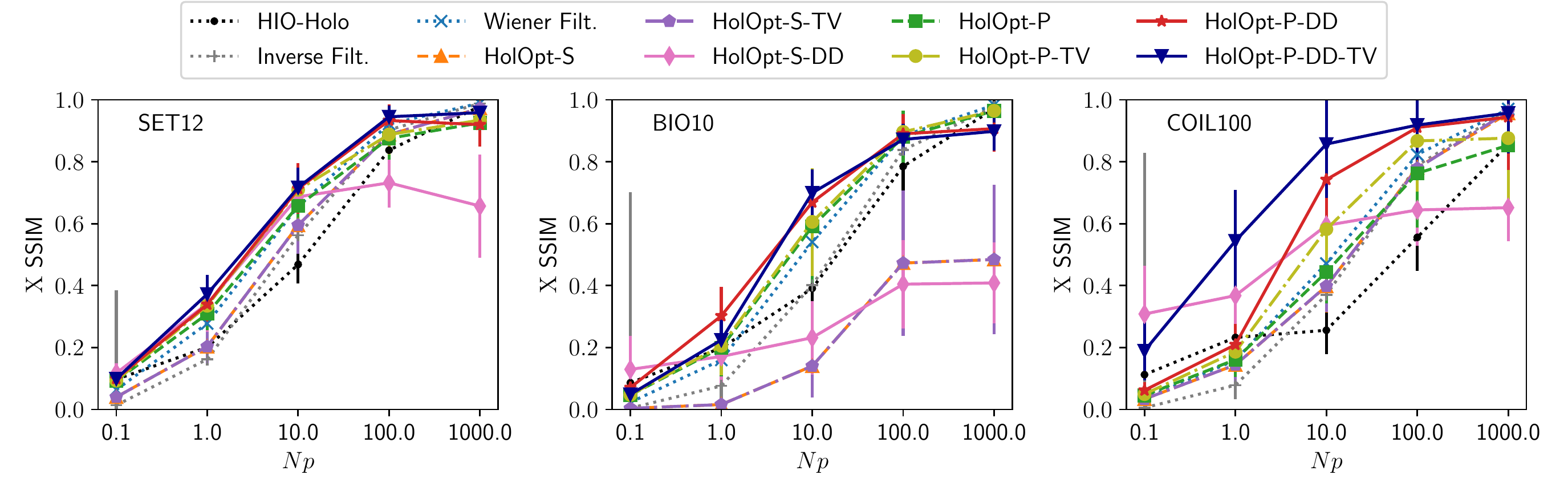}}
    \caption{\label{fig:denoising-graph-ssim} Reconstruction scores (SSIM) for SET12 (left), BIO10 (center), and COIL100 (right) as a function of the photon count $N_p$ with a 
    binary random reference and without beamstop. Corresponding reconstruction MSEs and residual observations MSEs are given in Figure \ref{fig:denoising-graph-l2} of the Appendix. \changes{Results with TV regularization where added.}}
\end{figure}

\section{Experiments}\label{sec:experiments}

{
\textbf{Data } Our strategy is demonstrated
on the following datasets. SET12
is a dataset consisting of 12 images used as a traditional benchmark in image processing,
 here resized to $128 \times 128$ pixels, while BIO10 
, resized to $256 \times 256$ pixels, contains 10 more realistic biological samples (these two datasets are available with the paper code).
We also consider the COIL100 dataset \citep{Nene1996} which contains 100 objects on a black background with $128 \times 128$ pixels.
We explicitly zero-out the background such that the support of the objects is not perfectly known. In contrast to non-holographic phase retrieval, a good 
reference should disambiguate the position of the sample within the frame.  %
Hence, COIL100 allows us to test the robustness of the different algorithms to reference design.
All images are converted to gray scale, and examples are presented in Figure \ref{fig:datasets} of the Appendix.}

\textbf{Benchmark setup } The 
strategy proposed in this work is compared against three 
algorithms for holographic CDI: inverse filtering, Wiener filtering and
Hybrid Input-Output modified for holographic phase retrieval, here referred to as HIO-Holo. We further augment HIO-Holo by selecting the best residual reconstruction over all iterations, to give it the fairest possible chance in our comparison.
As discussed in Section \ref{sec:related}, there is no comparable machine learning method that can be used here as a benchmark. 
However, we test variants of \hp~and \hpdd, termed \hs~and \hsdd~respectively, in which mean squared error (MSE) is optimized instead of the Poisson likelihood.
At high photon counts (low noise), Poisson noise is 
well-approximated by Gaussian noise and the two objectives are 
expected to perform similarly.
However, their difference becomes significant at lower photon counts (high noise).
The benefits of taking into account the Poisson nature of the noise are 
shown in
experiments below.
\changes{We also test variants with total variation (TV) regularization, a common strategy in image reconstruction.}

\textbf{Hyperparameters} 
\changes{Without access to a ground truth, the selection of hyperparameters of any of the compared algorithms inevitably relies on heuristics. 
The heuristic adopted here is to tune the hyperparameters by visual inspection on one of the specimens of each dataset at each noise level, and leave these parameters fixed for all subsequent specimens.
\changes{This would correspond to determining parameters once for each specific experimental setup.}
Practitioners are encouraged to proceed similarly choosing preferably the specimen for which they have the strongest prior knowledge for the hyperparameter selection.}
\changes{In Section \ref{sec:frechet} we propose a direction to formalize this procedure based on statistics of collection of images.}

For gradient descent, we use the Adam optimizer \citep{Kingma2015} and learning rates varying between 0.01 and 0.1, depending on the loss and prior. \changes{Maximum number of iterations is also fixed depending on noise level and the reconstruction at the best magnitude residual value along the iterations is retained as the final reconstruction.} 
Deep decoder parameters are gathered in Table \ref{tab:dd-params}
which also includes number of steps in the gradient descent. Some early stopping was found beneficial in order to avoid overfitting at high noise levels. As a result we adapt the number of iterations to the photon count.
\changes{TV regularization coefficients are chosen at the highest possible value that yields good visual reconstruction and does not deteriorate the final residual error. Selected values ranged between $10^2$ and $10^3$.}

\textbf{Evaluation } Algorithms are compared in terms of Structural Similarity Index (SSIM)  \cite{Wang2004}.
Comparisons in terms of  
relative reconstruction error --- Euclidean distance between the reconstructed $\hat \mX$ and ground truth specimen $\mX$ normalized by the $\ell_2$-norm of $\mX$ --- and relative residual error --- understood as the Euclidean distance between the observations $\mY$ and the noiseless output of the forward model for a specimen $\hat \mX$ normalized by the $\ell_2$-norm of $\mY$ --- are also reported in the Appendix.
Errors are averaged over images of each dataset as well as a few different random seeds for BIO10 and SET12 to increase the statistics of the small datasets. Error bars correspond to standard deviations.

\subsection{Noisy reconstruction with and without the deep decoder}
\label{sec:denoising}

In a first series of experiments, we examine the robustness to noise of \hp~and \hpdd~and \changes{their TV variants}. We set the reference $\mR$ to a $m \times m$ binary array with entries $0$ or $1$ sampled uniformly and independently, a reference design generally very favorable to the reconstruction \citep{WirtingerFlow,MarchesiniURA}. No beamstop mask is included. 

Figure \ref{fig:denoising-camera-visual} shows a clear qualitative improvement of
Poisson likelihood optimization methods 
over the baselines as noise increases (observed consistently over the different datasets, see also Figure \ref{fig:denoising-coil100-visual} in the Appendix).
In Figure \ref{fig:denoising-graph-ssim}, \hpdd~consistently reaches higher SSIM scores in the noise range $Np=1$ to $Np=100$ across datasets. \changes{For SET12 and COIL, which both feature images with large constant regions, the TV regularized version \hpdd-TV yields the best SSIMs.}
At very low noise $Np=1000$, the ordering of the methods varies, yet visual inspection confirms that all algorithms perform similarly and well, except for \hsdd~combining squared loss and deep decoder (see discussion below).
On the other hand, at $Np=0.1$, little information is left for the algorithms to retrieve: all methods reconstruct images with SSIM scores close to $0.1$, visually failing in different ways.

Among the variants of
HolOpt, we observe the following trends.
The reconstruction loss and visual quality of the samples are in almost all cases better with \hp~than with \hs, validating our adoption of the most realistic noise model and Poisson likelihood objective.
The difference between MSE and Poisson likelihood objectives is most drastic when including the deep decoder prior.
 In particular, the MSE loss sometimes leads to artifacts in the images reconstructed by \hsdd~(see Figures \ref{fig:denoising-camera-visual} and \ref{fig:denoising-coil100-visual} in Appendix \ref{app:exp}). 

 Thus, we focus only on 
the Poisson likelihood objective going forward. 
Regarding the use of a deep decoder, we distinguish several performance regimes for our method.
At low noise ($N_p \simeq 1000$ to $100$), there is no need for 
regularization by a deep decoder, and \hpdd~achieves reconstructions of similar quality to \hp. 
Here, the SSIM is typically slightly worse with a deep decoder than without (except for COIL images). 
Including a prior is not harmful, but often unnecessary.
At higher levels of noise, the denoising power of the deep decoder is beneficial.
{These observations are in accordance with the intuition that a prior helps when information is scarce, but is less helpful when observations of high quality are available.}

\changes{Focusing on visuals, Figures \ref{fig:denoising-camera-visual} and \ref{fig:denoising-coil100-visual}, we observe that the relative performances of \hp-TV, \hpdd~and \hpdd-TV are image dependent. The deep decoder usually allows better contrast. TV regularization smooths constant backgrounds, but sometimes also erases details (see e.g. the writing of the COIL box). Adding TV regularization to the deep decoder can hurt (COIL box, $Np=10$ ) or help (COIL box, $Np=1$). In the next experiments we focus on the behavior of \hp~and \hpdd, noting that a practitioner should also check for improvements with a TV regularization tuned to their setup. }

\changes{Finally, we note that training the deep decoder incurs an additional computational cost. For our implementations, HoloOpt-P-DD is  about  3  times  slower  than  the  classical  HIO-Holo, an  acceptable  slow-down in scientific imaging applications, if traded with reconstruction improvement.}\footnote{\changes{We report computational costs for runs on an NVIDIA Tesla V100SXM2 GPU reconstructing a 128$\times$128 pixel image:  HIO-Holo (1000 iterations) $\sim$27s, HoloOpt (2500  iterations) $\sim$3s  and  HoloOpt-P-DD  (2-layer)  (2500  iterations) $\sim$92s. Inverse filtering and Wiener filtering running times are negligible. Running times have not been optimized.}}

\begin{figure}[!htbp] 
    \centering
    \includegraphics[width=0.8\textwidth]{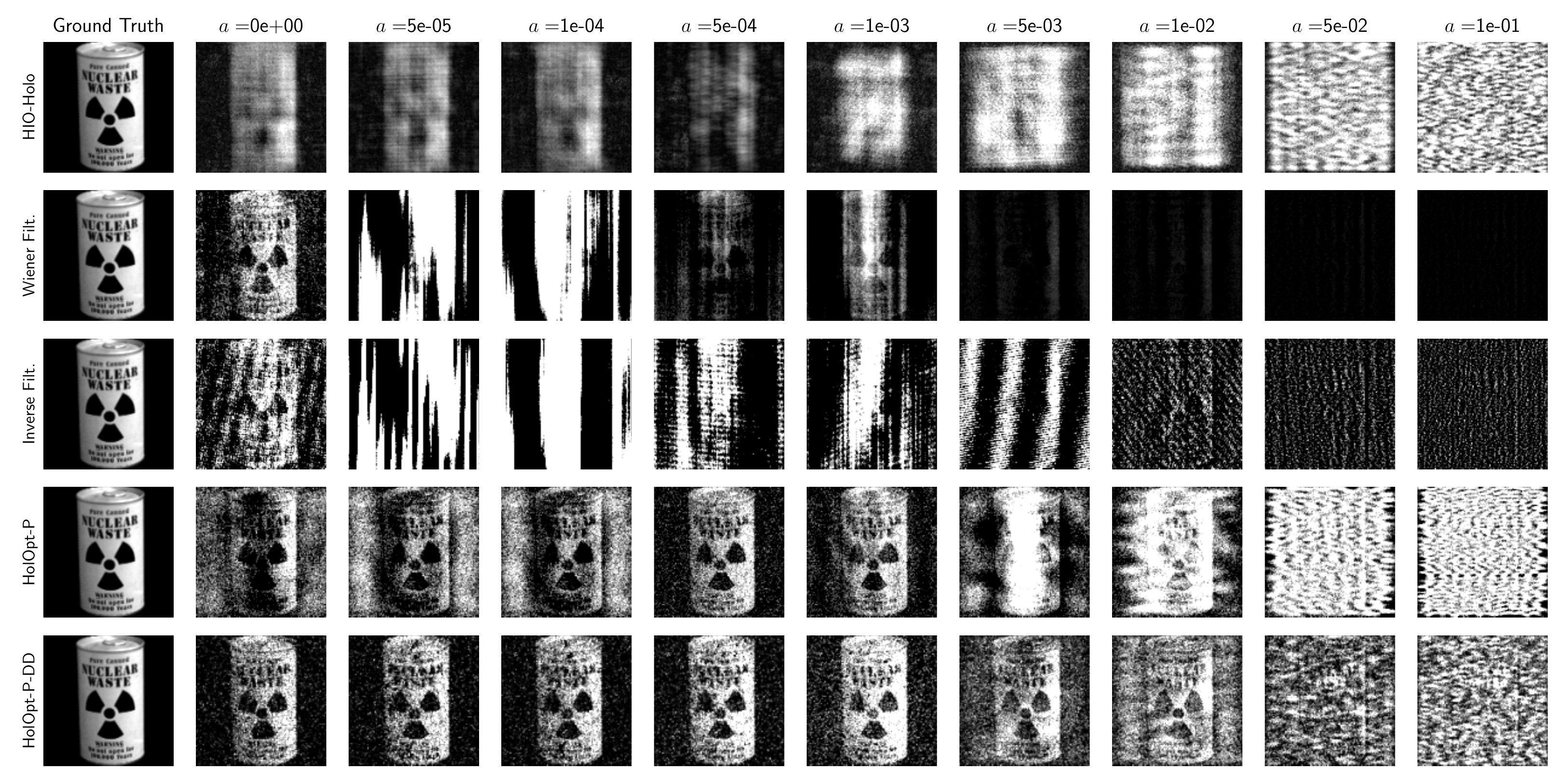}
    \caption{\label{fig:beamstop-SET12-images} { An example reconstructed image from 
    COIL dataset
    as a function of beamstop area fraction $a$ for fixed photon count $N_p = 1$. See also Figure \ref{fig:beamstop-SET12-images-orig} in the Appendix for a differently scaled visualization.
    }}
\end{figure}

\vspace{-6pt}
\subsection{Reconstructing with missing low frequencies} \label{sec:beamstop}
As discussed in Section \ref{sec:setup}, a universal 
feature of CDI experiments is a beamstop which obscures low-frequency magnitudes. In Wiener and inverse filtering, one simply sets the missing magnitudes to zero ~\citep{HERALDO}, whereas HIO-Holo can be made agnostic to the missing magnitudes.
We show here that our optimization-based methods can 
effectively incorporate an arbitrary beamstop in the forward model, as defined in Equation \ref{eq:poisson_loss}. 
Moreover, we expect and indeed observe that the deep decoder prior can be particularly useful in compensating for the missing magnitudes. 

\begin{figure} 
        \centering
        \includegraphics[trim= 0 0 0 2 , clip, width=\textwidth]{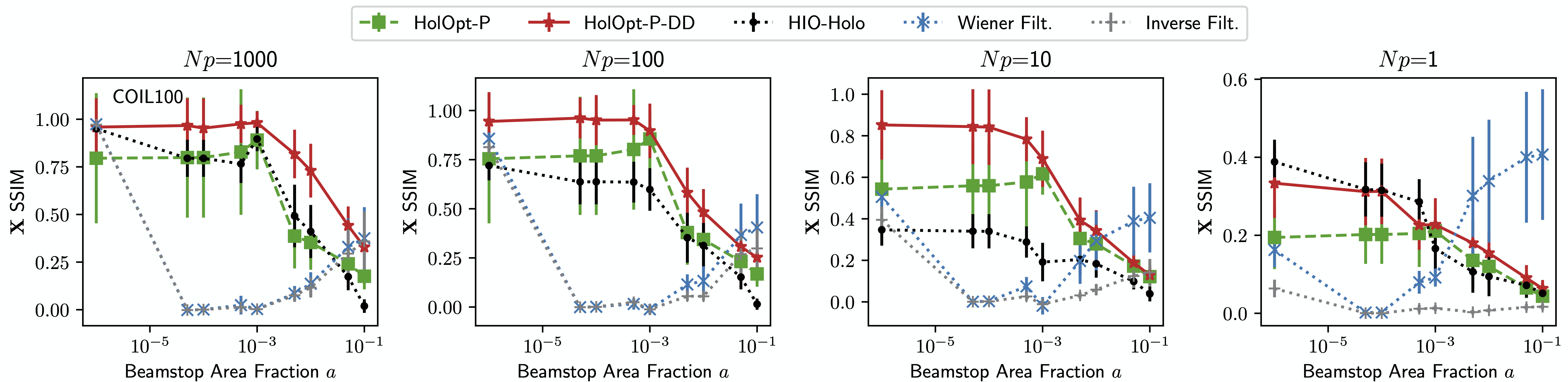}
        \caption{\label{fig:beamstop-error-plots} {Reconstruction SSIM as a function of beamstop area fraction. Baseline methods are run for 5 trials per image 
        while for tractability our methods are run 
        1 trial each on COIL100. Average SSIM and one standard deviation error bars are shown. 
        The leftmost datapoints
        correspond to no missing data,
         i.e. formally $a=0$ at the leftmost points. See Figure \ref{fig:beamstop-error-plots-all} for a comparision over all datasets.}}
\end{figure}

We evaluate our methods \hp~and \hpdd~at several noise levels and beamstop sizes, example reconstructions and SSIMs plots for COIL100 are  in Figures \ref{fig:beamstop-SET12-images} and \ref{fig:beamstop-error-plots} respectively. Supplemental plots for all datasets and noise levels are available in Figures \ref{fig:beamstop-area-fractions}-\ref{fig:beamstop-virus-1000} of the Appendix, including mean-squared error, which 
corroborate
the trends observed here. We consider square beamstop masks centered at the $0$ frequency, identified by their area fraction $a$: the fraction of the total measured magnitudes which are lost (visualized in Figure \ref{fig:beamstop-area-fractions} of the Appendix). 

We find that both \hp~and \hpdd~vastly outperform HIO-Holo, Wiener and inverse filtering at near-all noise levels, test images, and beamstop area fractions. The advantage in terms of SSIM of the deep decoder prior is image-dependent: \hpdd~yields most significantly improved SSIM relative to \hp~on BIO10 at low photon counts, and at all noise levels for COIL100, but makes little difference in SSIM on the SET12 dataset (see Figure \ref{fig:beamstop-error-plots}). Yet visually, as in Figure \ref{fig:beamstop-SET12-images}, the reconstructions via \hpdd~ on all datasets have enhanced contrast and details relative to \hp; furthermore, both \hp~ and \hpdd~ visually outperform baselines. We note that even at settings where baselines achieve higher SSIM, notably with largest beamstop size and low photon counts on the COIL100 dataset, visual inspection of Figures \ref{fig:beamstop-SET12-images} and \ref{fig:beamstop-coil-10} illustrate that no method performs recognizable reconstruction at this beamstop size; the supposed improvement thus appears to be an idiosyncrasy of the SSIM\footnote{We note that the performance of Wiener and inverse filtering suffers at small beamstop sizes due to the ``ringing" effect of low frequencies, an effect which occurs regardless of the method for filling the low magnitudes. Evident as a striping effect in Figure \ref{fig:beamstop-SET12-images}, this explains the occasional increase in SSIM with increasing beamstop size evident in Figure \ref{fig:beamstop-error-plots}.}. 
In sum, the performances of our methods smoothly degrade with increasing fraction of missing magnitudes $a$, and enable reconstructions with lower error (Figure \ref{fig:beamstop-error-plots}) and visually improved features (Figure \ref{fig:beamstop-SET12-images}) relative to baselines. This provides powerful evidence that 
our method
can enable refined reconstructions given even large fractions of lost magnitude data at the highest noise levels. 

\subsection{Robustness to reference separation}
\label{sec:sep-exp}

The separation distance between the specimen and the reference limits the smallest resolution that can be achieved \citep{CDI-Book}, and thus would ideally be minimized. 
For a specimen of size $m \times m$ pixels, 
the \textit{holographic separation condition} dictates that
inverse filtering and Wiener filtering require a full separation 
$(\mX | \mathbf{0}_{m \times m} | \mR)$.
In contrast, 
our approach only requires that the forward model 
be differentiable, which is the case for any reference placement.

\begin{figure}
    \centering
    {\includegraphics[trim = 0 0 0 15,  clip,width=0.49\textwidth]{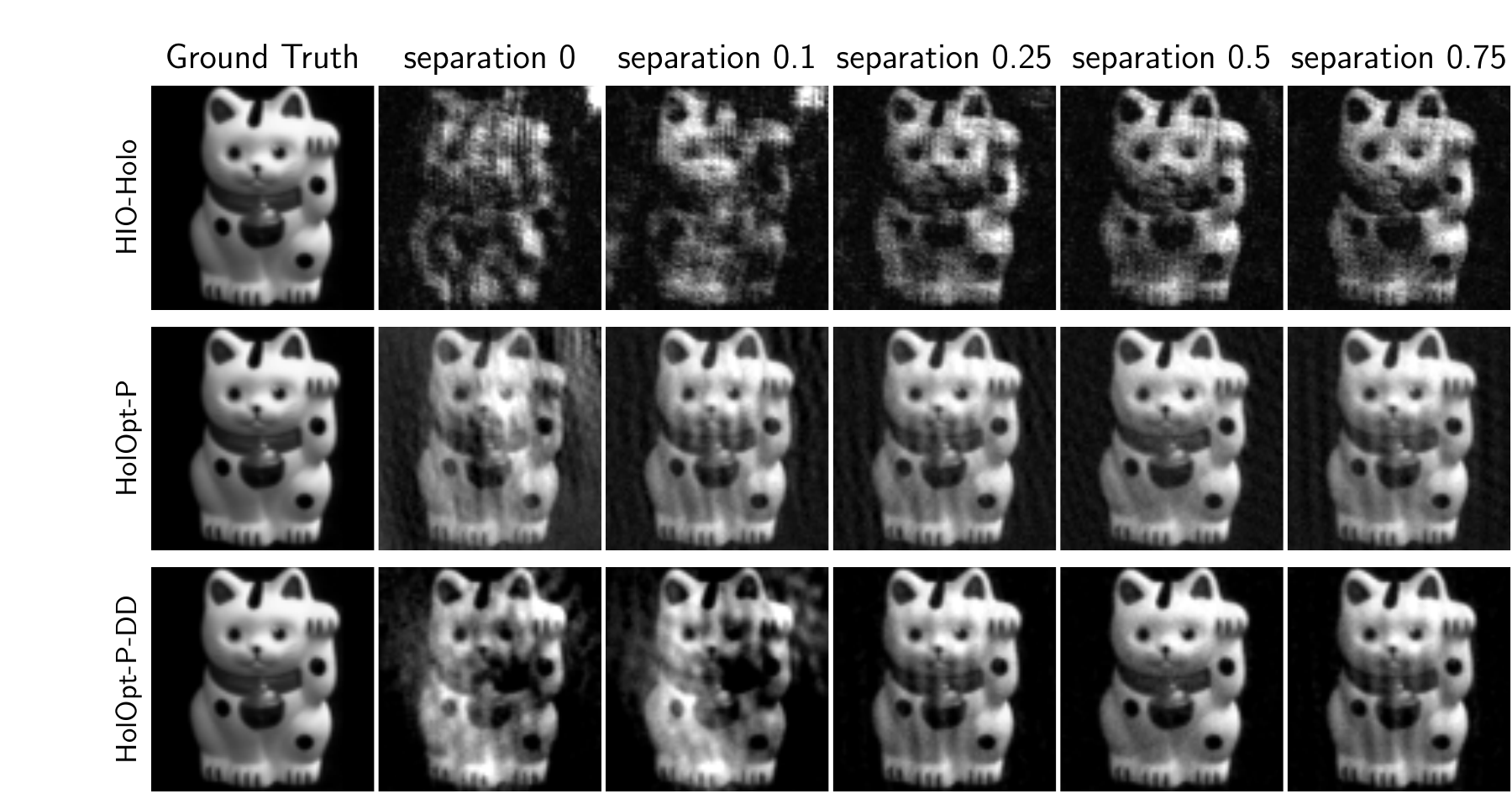}}
    {\includegraphics[trim = 0 0 0 15,  clip,width=0.49\textwidth]{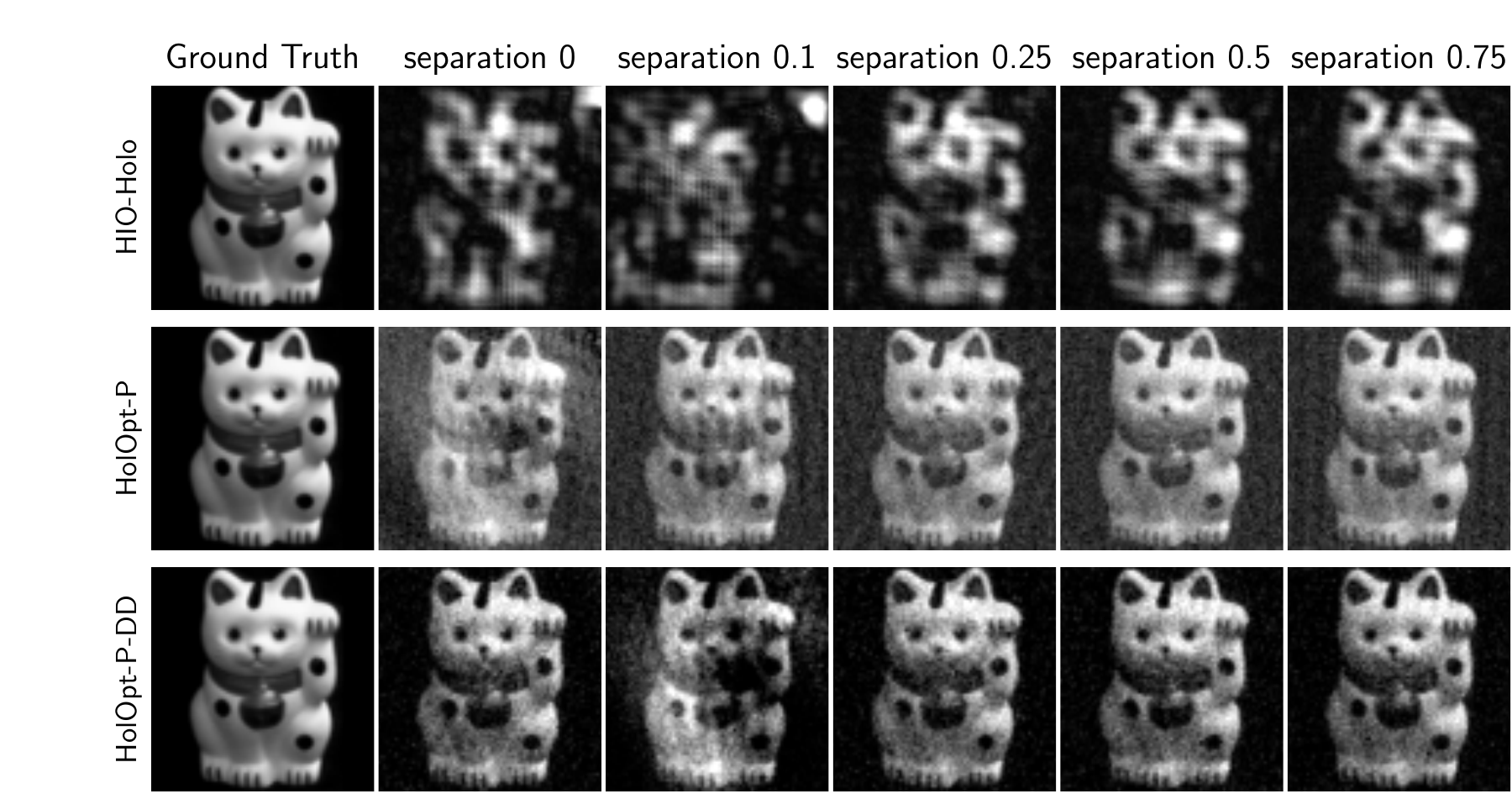}}\\
    \caption{\label{fig:sep-visual-np1-np10} {Reconstructions for photon counts $N_p = 10$ (left) and $N_p = 1$ (right) with a $0.1 m \times 0.1 m $ binary random reference as a function of the relative separation. A separation of $0.5$ implies that the left-most non-zero pixel of the reference is $0.5n$ pixels away form the image. See Figure \ref{fig:sep-visual-np1-np10-all}-\ref{fig:sep-visual-np01-np100} in Appendix for $N_p = 100$ and $N_p = 0.1$ and different samples.}}
\end{figure}

\begin{figure}
    {\includegraphics[trim = 0 0 0 0,  clip,width=1.\textwidth]{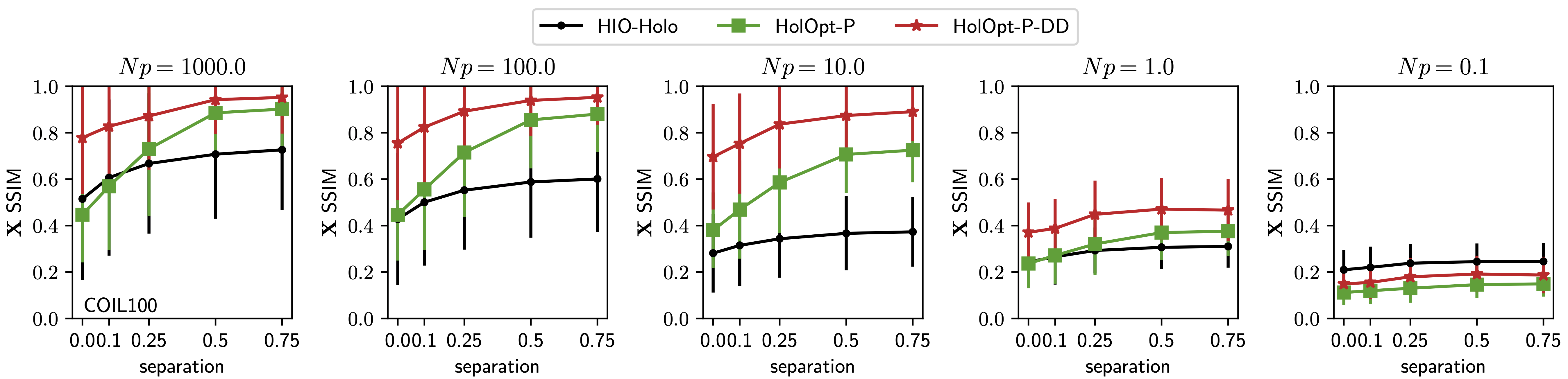}}
    \caption{\label{fig:sep-ssim-graph} Reconstruction SSIM for decreasing photon counts $N_p$ as a function of the relative separation (see caption of Figure \ref{fig:sep-visual-np1-np10}).
    Similar plot for all datasets is given in Figure \ref{fig:sep-ssim-graph-all}. Corresponding MSEs are plotted in Figure \ref{fig:sep-graph} of the Appendix.}
    \vspace{-6pt}
\end{figure}

Here, 
we explore signal-reference association of the form $(\mX| \mR| \mathbf{0}_{m \times m})$, where $\mR$ is a random binary block 
of small size: it is non-zero only on a box of size $0.1 m \times 0.1 m$
with
uniformly and independently chosen 
entries in $\{0,1\}$,
and 
the box
position is varied between experiments. 
No beamstop is included.
Figure \ref{fig:sep-visual-np1-np10} displays example
reconstructions
 for $N_p = 10$ and $N_p = 1$
and Figure \ref{fig:sep-ssim-graph}
 reports achieved SSIM scores as a function of the relative specimen-reference separation for the three datasets and multiple level of noises.
The only applicable baseline 
is HIO-Holo, illustrating the challenging nature of this setting.

For the three compared methods, the quality of the reconstruction depends slightly on the separation, only HIO-Holo at 0 separation generates clearly poorer images. 
Visually, HIO-Holo yields the less sharp images and \hpdd~is almost always the best algorithm.  
Without the deep decoder, \hp~ reconstructions are also of good quality  but sometimes show less contrast. 
The neural network prior makes a difference in particular at high levels of noise and on COIL100 images, for which the deep decoder captures well the solid black background. 
These observations are consistent with the average SSIM, but we note that the variance across images in the dataset can be significant. 
At very high noise $Np = 0.1$ photon/pixel, HIO-Holo has better SSIM scores than \hp~methods, yet this is arguably the limit of reconstruction possibilities for any of the methods (SSIM $< 0.2$ and visuals in Figure \ref{fig:sep-visual-np01-np100}).

Overall, this more challenging 
experimental setting confirms the efficiency of our proposed method, allowing reconstruction 
even 
if the holographic separation condition is broken. Moreover, a deep decoder prior provides clear benefits even over direct optimization in certain cases.

\subsection{Varying the oversampling rate}
\label{sec:ovs}
\begin{figure}[!htbp] 
    \centering
    {\includegraphics[width=0.49\textwidth]{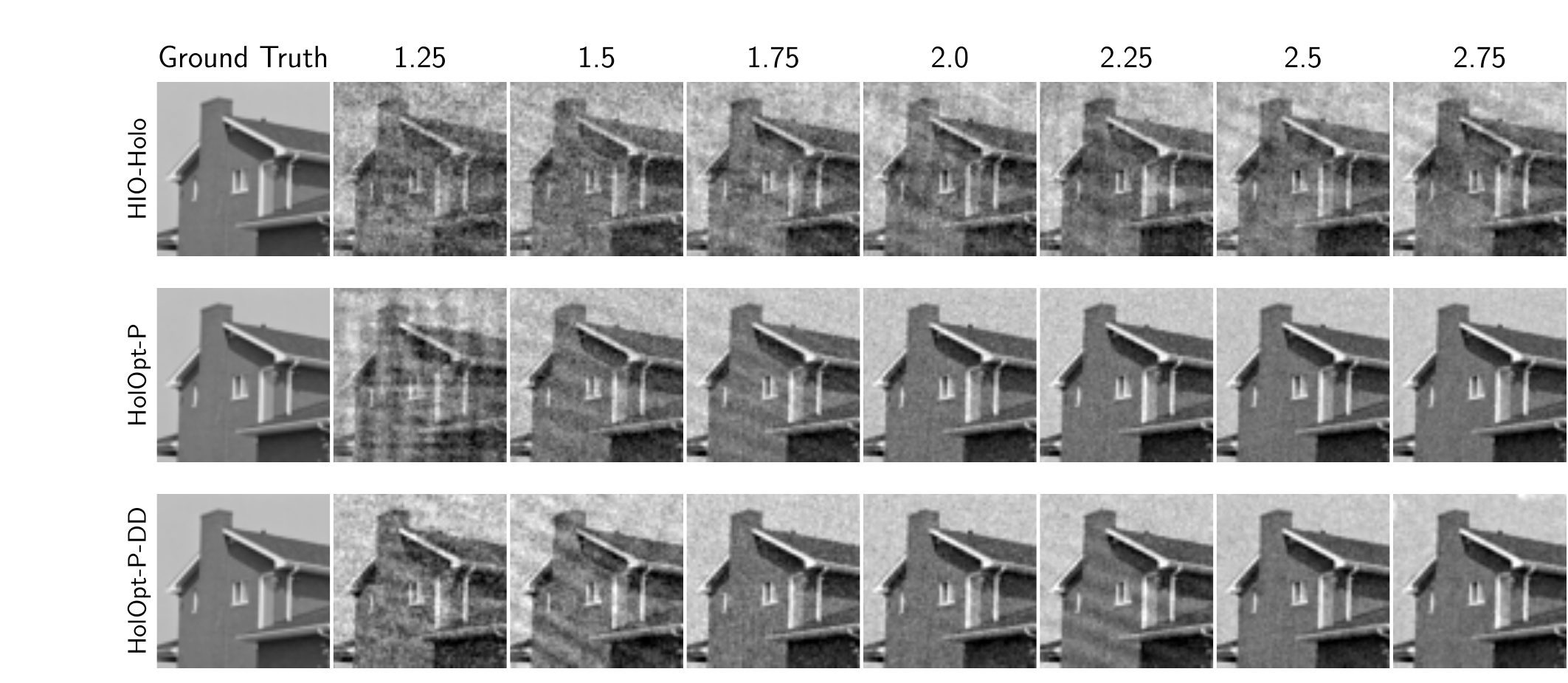}}
    {\includegraphics[width=0.49\textwidth]{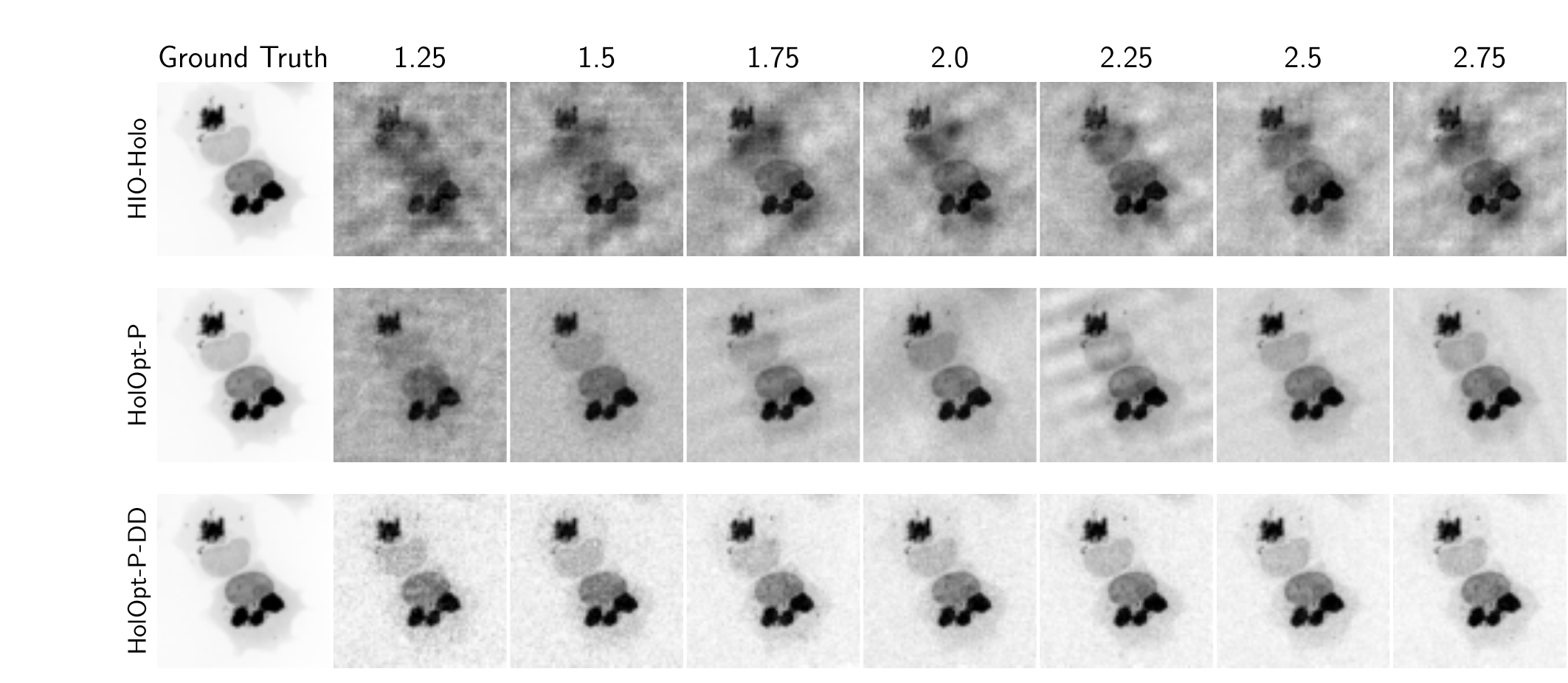}}\\
    \setlength{\belowcaptionskip}{-20pt}
    \caption{Reconstructed images for samples from SET12, BIO10
    datasets with varying oversampling factors (numbers above each column) at $N_p=10$ photon/pixel. See Figure \label{fig:ovs-visuals-COIL} for a figure incuding COIL images.}
    \label{fig:ovs-visuals}
\vspace*{\floatsep}
    \centering
    \includegraphics[trim= 0 30 0 0 , clip, width=\textwidth]{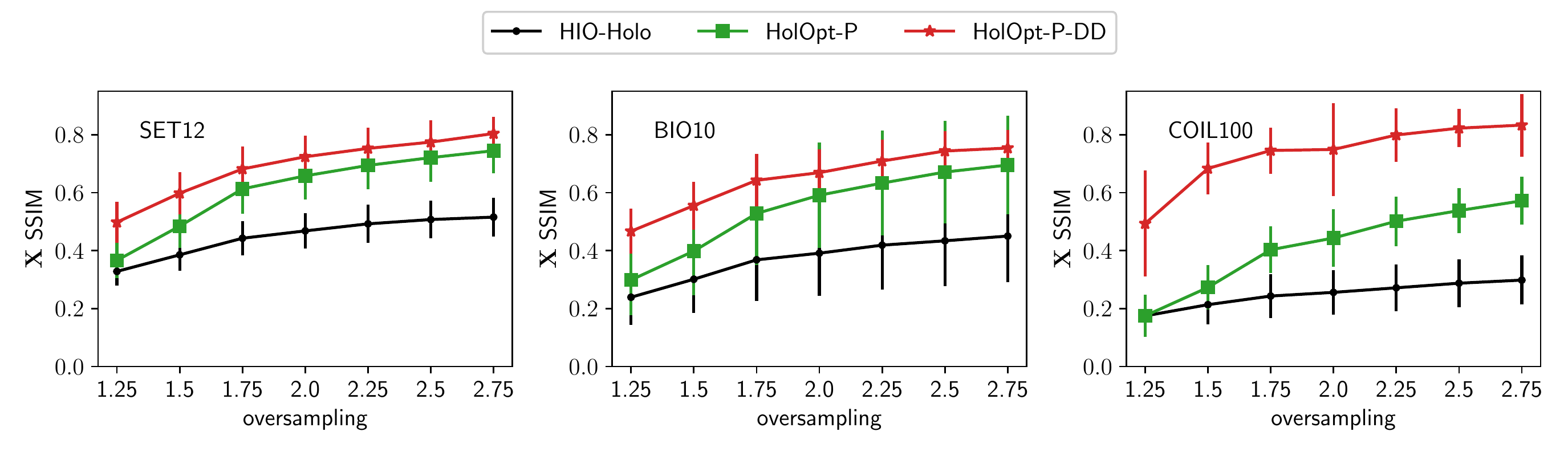}
    \setlength{\belowcaptionskip}{-20pt}
    \caption{Average SSIM scores for SET12, BIO10 and COIL100 with varying oversampling factor at $N_p=10$ photon/pixel. Errors bars represent standard deviations.}
    \label{fig:ovs-graph}
\vspace*{\floatsep}
    \centering
    \includegraphics[width=0.32\textwidth]{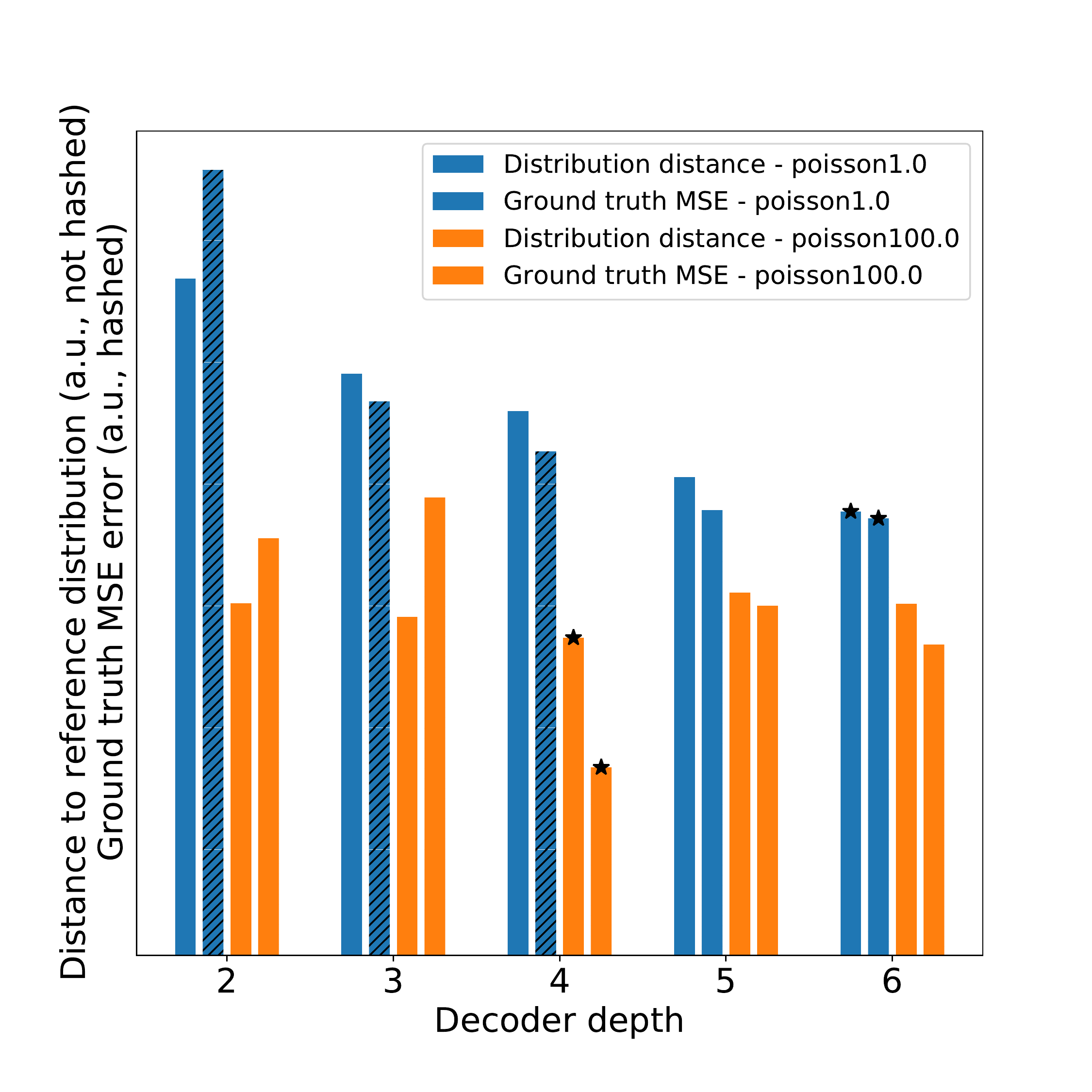}
    \includegraphics[width=0.32\textwidth]{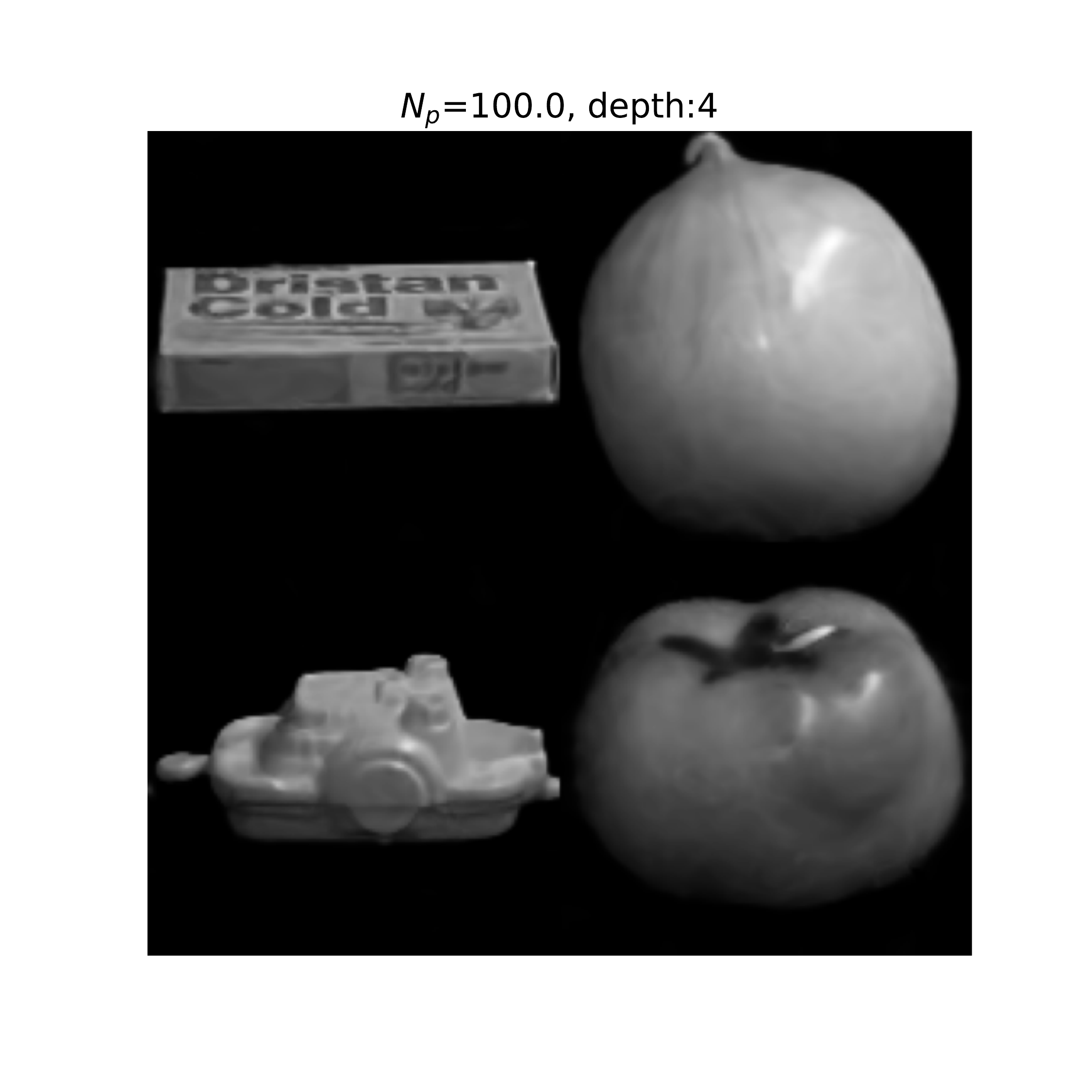}
    \includegraphics[width=0.32\textwidth]{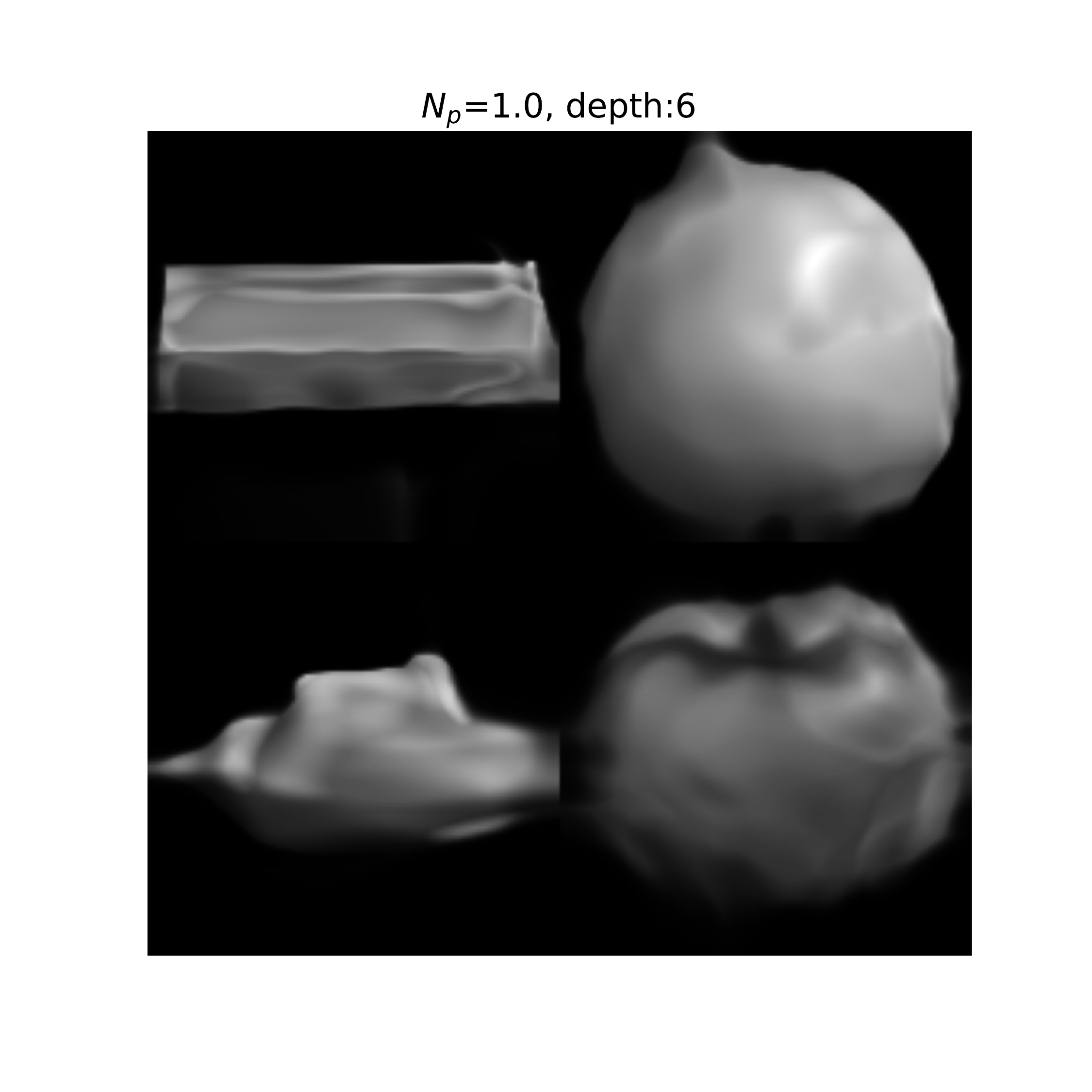}
    \caption{\changes{Evaluation of deep decoder depth selection using Fr\'echet Inception Distance (FID) to a natural image prior. Different colors represent different noise levels. The unhashed bars represent Fr\'echet Inception Distance, and the hashed bars represent MSE error compared to ground truth. Asterisks are placed at the decoder depth representing minimal FID and minimal MSE error. They correspond for all noise levels. Reconstruction examples at the selected decoder depths are shown on the right. While  $N_p=100$ shows sharp reconstructions at depth 4, the noisy $N_p=1.0$ leads to selection of depth 6, which blurrs the finer features but preserves the shape.}}
    \label{fig:frechet}
\end{figure}

In a last experiment, we consider yet another challenge for reconstruction by lowering the oversampling rate of the observation. For oversampling factors larger or equal to 2, the inverse problem can be solved exactly in the absence of noise ~\citep{Hayes}. In the presence of noise, increased oversampling is all the more beneficial, as collecting more information can compensate for the noise corruption. We test how our methods perform relative to HIO-Holo for oversampling ratios around the critical value of 2. Note that inverse and Wiener filtering are not well-defined for oversampling rates smaller than 2. Figures~\ref{fig:ovs-visuals} and \ref{fig:ovs-graph} display results for a well-separated binary reference and without beamstop at $N_p = 10$ photon/pixel (see Appendix \ref{app:ovs} for identical figures at $N_p = 1$). All images reconstructed by \hpdd~are visually superior to images reconstructed by HIO-Holo at oversampling factors both smaller and larger than 2. \hpdd~ also produces sharper images than HIO-Holo, although sometimes with less contrast. This phenomenology is largely captured by the SSIM.

\subsection{Selecting hyperparameters using group statistics}
\label{sec:frechet}
\changes{In order for systematic hyperparameter selection to be practically useful, it is required that it can be performed without access to the ground truth image sought to be reconstructed from measurements.

For a reconstructed image or a group of reconstructed images with unknown ground truth stemming from a specific hyperparameter setting, 
an idea is
to quantify the extent to which the reconstructions look \textit{in-distribution} or expected with respect to the general class of images under investigation. We propose to do this using the Fr\'echet Inception Distance (FID,  \cite{Heusel2017frechet}).
Since the distribution of natural images is extremely high-dimensional and complicated, it needs to be approximated. 
This can be done by selecting a collection of images, extracting features, and using the statistic of the features on the collection to model a Gaussian distribution.
The feature extraction should ideally be Gaussianizing, contracting for images within the collection and mapping non-images, \emph{out-of-distribution} samples, to outlying points.

As a proof of concept, we study parameter selection on the COIL100 data set using features from a pretrained VGG16 convolutional neural network \citep{simonyan2015deep}.
Let $f_\textrm{VGG}(x)$ denote the function returning last-layer logits from a pretrained VGG16 network\footnote{We use the one available in the \texttt{torchvision} python package}.
In order to create a suitable image distribution, we divide the COIL100 dataset into two halves. The first 50 images will serve as an image prior, and the second half will be used to perform hyperparameter selection without ground truth.
We compute
\begin{equation}
    \mu_\textrm{prior} =\frac{1}{50}\sum_{i=1}^{50} f_\textrm{VGG}(x_i),\quad\quad
    \Sigma_\textrm{prior} = \frac{1}{49}\sum_{i=1}^{50}(f_\textrm{VGG}(x_i) - \mu_\textrm{prior})(f_\textrm{VGG}(x_i) - \mu_\textrm{prior})^T.
\end{equation}
For a given hyperparameter setting, let $\hat x_i, i=51,\dots,100$ be the reconstructed images. Analogously to the computation of the prior distribution parameters, we compute
\begin{equation}
    \mu_\textrm{recon}=\frac{1}{50}\sum_{i=51}^{100} f_\textrm{VGG}(\hat x_i),\quad\quad 
    \Sigma_\textrm{recon} = \frac{1}{49}\sum_{i=51}^{100}(f_\textrm{VGG}(\hat x_i) - \mu_\textrm{recon})(f_\textrm{VGG}(\hat x_i) - \mu_\textrm{recon})^T.
\end{equation}

The Gaussians defined by $(\mu_\textrm{prior}, \Sigma_\textrm{prior})$ and $(\mu_\textrm{recon}, \Sigma_\textrm{recon})$ can now be compared using the Wasserstein 2-distance between distributions, which reduces to
\begin{equation}
    d_\textrm{prior, recon} = \|\mu_\textrm{prior} - \mu_\textrm{recon}\|^2 + \textrm{trace}(\Sigma_\textrm{prior} + \Sigma_\textrm{recon} - 2\sqrt{\Sigma_\textrm{prior}\Sigma_\textrm{recon}}).
\end{equation}

Using this distance, we can evaluate reconstructions with different hyperparameter settings and select the one with the smallest distance to the collection of prior images.

We exemplify this strategy for setting depth of deep decoders.
Figure \ref{fig:frechet} shows FIDs to the natural image prior for different depths at two different noise levels. For a given noise level, the FID is minimal at the same decoder depth as the MSE
with ground truth. We have thus created a model selection heuristic that can select the decoder depth leading to minimal squared reconstruction error. While these statitics offer a promising direction of systematic hyperparameter selection, the FID evaluated here exhibits the same selection biases as MSE (e.g. privileging low-frequency accuracy over high frequencies). A practitioner may prefer shallower networks, as  in the experiments from the sections above. 
}

\section{Optical Laser Coherent Imaging Experimental Data}
\label{sec:realdata}
\begin{figure} 
    \centering
    {\includegraphics[trim = 270 450 50 430,  clip,width=8.5in]{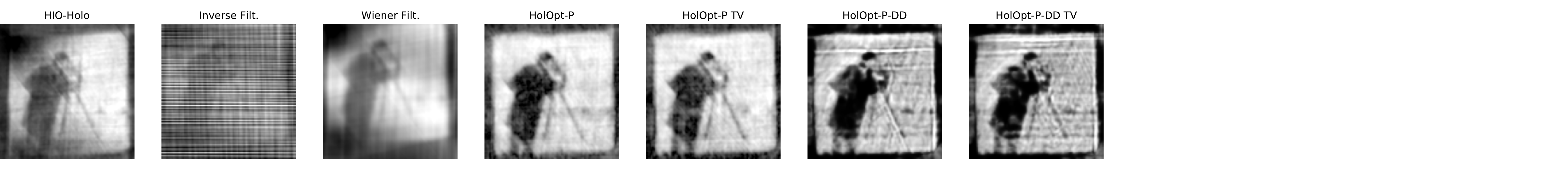}/}
    \caption{\changes{Reconstructions from experimental data. Further details on the experimental setup and methods can be found in Appendix
\ref{app:realdata}.}\label{fig:realdata-crop}}
\end{figure}

\changes{In this section, we evaluate 
\hp~ and \hpdd, with and without TV regularization, on the experimental data used in \citet{HERALDO_experimental}, obtained with permission from the authors. In this experimental setup, a slide printed with the CAMERA image and an adjacent triangular reference object were exposed to a collimated wave from a He-Ne laser (632.8 nm wavelength), and the resultant Fourier intensity pattern was measured. As this was not an X-ray experiment, no beamstop was necessary.
We refer to \cite{HERALDO_experimental} for precise experimental details and to Appendix \ref{app:realdata} for detail of our methods.  We compare variants of \hp~to baselines of Wiener filtering, inverse filtering, and HIO-Holo.\footnote{\changes{While the HERALDO method can yield sharper reconstructions, it is not directly comparable to the techniques presented here. In particular, HERALDO requires specific assumptions on the reference geometry and uses a tailored reconstruction operator for each new reference shape, whereas our method is one-size-fits-all \citep{HERALDO}.}}
The results are shown in Figure \ref{fig:realdata-crop}.
\hp~
produces superior reconstructions, with the qualitative best by either \hp~ or \hpdd~TV. Those produced by the Wiener and inverse filtering baselines have important artifacts; that of HIO-Holo looks better than the former two, but still suffers from significant
shadowing artifacts in the upper and lower left corners. We also observe that TV-regularization alone noticeably smooths out the horizon line, effectively removing it, while the addition of a deep decoder prevents this phenomenon. 
}

\section{Discussion and Conclusion}\label{sec:discussion}

In this paper, we have shown that recent progress at the intersection of machine learning and inverse problems can yield 
highly successful algorithms which also account for realistic experimental challenges. 
Our novel optimization framework for holographic phase retrieval improves on state-of-the-art reconstruction, even in the most difficult experimental settings and without external training data. 
Untrained image priors are confirmed to be powerful tools, especially when a significant amount of information is missing from the measured data due to low photon counts, beamstop-obscured frequencies and small oversampling. 
Due to its practicality and flexibility, we believe our methodology should prove 
valuable for 
practitioners. 
\changes{We confirm the success of our approach on the experimental data in the optical range. For reasons of data availability, we leave an evaluation of the method on X-ray holographic CDI to future work.}
Finally, our framework is easily adaptable to different variants of the problem --- even the mathematically distinct non-holographic setting --- and should enable similarly improved reconstruction with other imaging modalities than Holographic CDI, such as optical holography ~\citep{Cossairt2015},  magnetic holographic imaging \citep{WenHuMagneticImaging}, and ptychography \citep{StefanoPtychography}. 

\bibliography{iclr2021_conference}

\begin{thebibliography}{50}
\providecommand{\natexlab}[1]{#1}
\providecommand{\url}[1]{\texttt{#1}}
\expandafter\ifx\csname urlstyle\endcsname\relax
  \providecommand{\doi}[1]{doi: #1}\else
  \providecommand{\doi}{doi: \begingroup \urlstyle{rm}\Url}\fi

\bibitem[Aubin et~al.(2020)Aubin, Loureiro, Baker, Krzakala, and
  Zdeborova]{Aubin2020}
Benjamin Aubin, Bruno Loureiro, Antoine Baker, Florent Krzakala, and Lenka
  Zdeborova.
\newblock {Precise asymptotics for phase retrieval and compressed sensing with
  random generative priors}.
\newblock In \emph{Proceedings of The First Mathematical and Scientific Machine
  Learning Conference, PMLR}, volume 107, pages 55--73, 2020.

\bibitem[Barmherzig and Sun(2020)]{Barmherzig20LowPhoton}
David~A. Barmherzig and Ju~Sun.
\newblock Low-photon holographic phase retrieval.
\newblock In \emph{Imaging and Applied Optics Congress}, page JTu4A.6. Optical
  Society of America, 2020.
\newblock URL
  \url{http://www.osapublishing.org/abstract.cfm?URI=COSI-2020-JTu4A.6}.

\bibitem[Barmherzig et~al.(2019)Barmherzig, Sun, Li, Lane, and
  Cand{\`{e}}s]{BarmherzigEtAl2019Holographic}
David~A Barmherzig, Ju~Sun, Po-Nan Li, T~J Lane, and Emmanuel~J Cand{\`{e}}s.
\newblock Holographic phase retrieval and reference design.
\newblock \emph{Inverse Problems}, 35\penalty0 (9):\penalty0 094001, Aug 2019.
\newblock \doi{10.1088/1361-6420/ab23d1}.
\newblock URL \url{https://doi.org/10.1088%2F1361-6420%2Fab23d1}.

\bibitem[Barmherzig et~al.(2020)Barmherzig, Barnett, Epstein, Greengard,
  Magland, and Rachh]{barmherzig2020recovering}
David~A. Barmherzig, Alex~H. Barnett, Charles~L. Epstein, Leslie~F. Greengard,
  Jeremy~F. Magland, and Manas Rachh.
\newblock Recovering missing data in coherent diffraction imaging, 2020.

\bibitem[Bostan et~al.(2020)Bostan, Heckel, Chen, Kellman, and
  Waller]{Bostan2020}
Emrah Bostan, Reinhard Heckel, Michael Chen, Michael Kellman, and Laura Waller.
\newblock {Deep phase decoder: self-calibrating phase microscopy with an
  untrained deep neural network}.
\newblock \emph{Optica}, 7\penalty0 (6):\penalty0 559, jun 2020.
\newblock ISSN 2334-2536.
\newblock \doi{10.1364/OPTICA.389314}.
\newblock URL
  \url{https://www.osapublishing.org/abstract.cfm?URI=optica-7-6-559}.

\bibitem[{Candès} et~al.(2015){Candès}, {Li}, and
  {Soltanolkotabi}]{WirtingerFlow}
E.~J. {Candès}, X.~{Li}, and M.~{Soltanolkotabi}.
\newblock Phase retrieval via wirtinger flow: Theory and algorithms.
\newblock \emph{IEEE Transactions on Information Theory}, 61\penalty0
  (4):\penalty0 1985--2007, 2015.

\bibitem[Fienup(1978)]{FienupHIO}
J.~R. Fienup.
\newblock Reconstruction of an object from the modulus of its fourier
  transform.
\newblock \emph{Opt. Lett.}, 3\penalty0 (1):\penalty0 27--29, Jul 1978.
\newblock \doi{10.1364/OL.3.000027}.
\newblock URL \url{http://ol.osa.org/abstract.cfm?URI=ol-3-1-27}.

\bibitem[Gabor(1948)]{Holog1}
D.~Gabor.
\newblock A new microscopic principle.
\newblock \emph{Nature}, 161:\penalty0 777--778, 1948.

\bibitem[Ghigo et~al.(2008)Ghigo, Kartenbeck, Lien, Pelkmans, Capo, Mege, and
  Raoult]{Mimivirus}
Eric Ghigo, J{\"{u}}rgen Kartenbeck, Pham Lien, Lucas Pelkmans, Christian Capo,
  Jean-Louis Mege, and Didier Raoult.
\newblock {Ameobal Pathogen Mimivirus Infects Macrophages through
  Phagocytosis}.
\newblock \emph{PLOS Pathogens}, 4\penalty0 (6):\penalty0 1--17, 2008.
\newblock \doi{10.1371/journal.ppat.1000087}.
\newblock URL \url{https://doi.org/10.1371/journal.ppat.1000087}.

\bibitem[Gorkhover et~al.(2018)Gorkhover, Ulmer, Ferguson, Bucher, Maia,
  Bielecki, Ekeberg, Hantke, Daurer, Nettelblad, Andreasson, Barty, Bruza,
  Carron, Hasse, Krzywinski, Larsson, Morgan, M{\"u}hlig, M{\"u}ller, Okamoto,
  Pietrini, Rupp, Sauppe, van~der Schot, Seibert, Sellberg, Svenda, Swiggers,
  Timneanu, Westphal, Williams, Zani, Chapman, Faigel, M{\"o}ller, Hajdu, and
  Bostedt]{Gorkhover2018}
Tais Gorkhover, Anatoli Ulmer, Ken Ferguson, Max Bucher, Filipe R. N.~C. Maia,
  Johan Bielecki, Tomas Ekeberg, Max~F. Hantke, Benedikt~J. Daurer, Carl
  Nettelblad, Jakob Andreasson, Anton Barty, Petr Bruza, Sebastian Carron, Dirk
  Hasse, Jacek Krzywinski, Daniel S.~D. Larsson, Andrew Morgan, Kerstin
  M{\"u}hlig, Maria M{\"u}ller, Kenta Okamoto, Alberto Pietrini, Daniela Rupp,
  Mario Sauppe, Gijs van~der Schot, Marvin Seibert, Jonas~A. Sellberg, Martin
  Svenda, Michelle Swiggers, Nicusor Timneanu, Daniel Westphal, Garth Williams,
  Alessandro Zani, Henry~N. Chapman, Gyula Faigel, Thomas M{\"o}ller, Janos
  Hajdu, and Christoph Bostedt.
\newblock Femtosecond x-ray fourier holography imaging of free-flying
  nanoparticles.
\newblock \emph{Nature Photonics}, 12\penalty0 (3):\penalty0 150--153, Mar
  2018.
\newblock ISSN 1749-4893.
\newblock \doi{10.1038/s41566-018-0110-y}.
\newblock URL \url{https://doi.org/10.1038/s41566-018-0110-y}.

\bibitem[Goy et~al.(2018)Goy, Arthur, Li, and Barbastathis]{Goy2018}
Alexandre Goy, Kwabena Arthur, Shuai Li, and George Barbastathis.
\newblock {Low Photon Count Phase Retrieval Using Deep Learning}.
\newblock \emph{Physical Review Letters}, 121\penalty0 (24):\penalty0 1--8,
  2018.
\newblock ISSN 10797114.
\newblock \doi{10.1103/PhysRevLett.121.243902}.

\bibitem[Guizar-Sicairos and Fienup(2007)]{HERALDO}
Manuel Guizar-Sicairos and James~R. Fienup.
\newblock Holography with extended reference by autocorrelation linear
  differential operation.
\newblock \emph{Opt. Express}, 15\penalty0 (26):\penalty0 17592--17612, Dec
  2007.
\newblock \doi{10.1364/OE.15.017592}.
\newblock URL
  \url{http://www.opticsexpress.org/abstract.cfm?URI=oe-15-26-17592}.

\bibitem[Guizar-Sicairos and Fienup(2008)]{HERALDO_experimental}
Manuel Guizar-Sicairos and James~R. Fienup.
\newblock Direct image reconstruction from a fourier intensity pattern using
  heraldo.
\newblock \emph{Opt. Lett.}, 15\penalty0 (33(22)):\penalty0 2668--70, Nov 2008.
\newblock \doi{10.1364/ol.33.002668}.
\newblock URL \url{https://pubmed.ncbi.nlm.nih.gov/19015703/}.

\bibitem[Hand et~al.(2018)Hand, Leong, and Voroninski]{Hand2018}
Paul Hand, Oscar Leong, and Vladislav Voroninski.
\newblock {Phase Retrieval Under a Generative Prior}.
\newblock In \emph{Neural Information Processing Systems 2018}, number NeurIPS,
  2018.
\newblock URL \url{http://arxiv.org/abs/1807.04261}.

\bibitem[{Hayes}(1982)]{Hayes}
M.~{Hayes}.
\newblock The reconstruction of a multidimensional sequence from the phase or
  magnitude of its fourier transform.
\newblock \emph{IEEE Transactions on Acoustics, Speech, and Signal Processing},
  30\penalty0 (2):\penalty0 140--154, 1982.

\bibitem[He et~al.(2015)He, Sharma, and Cossairt]{Cossairt2015}
Kuan He, Manoj~Kumar Sharma, and Oliver Cossairt.
\newblock High dynamic range coherent imaging using compressed sensing.
\newblock \emph{Opt. Express}, 23\penalty0 (24):\penalty0 30904--30916, Nov
  2015.
\newblock \doi{10.1364/OE.23.030904}.
\newblock URL
  \url{http://www.opticsexpress.org/abstract.cfm?URI=oe-23-24-30904}.

\bibitem[Heckel and Hand(2019)]{Heckel2018}
Reinhard Heckel and Paul Hand.
\newblock {Deep Decoder: Concise Image Representations from Untrained
  Non-convolutional Networks}.
\newblock \emph{International Conference on Learning Representations}, 2019.
\newblock URL \url{http://arxiv.org/abs/1810.03982}.

\bibitem[Heusel et~al.(2017)Heusel, Ramsauer, Unterthiner, Nessler, and
  Hochreiter]{Heusel2017frechet}
Martin Heusel, Hubert Ramsauer, Thomas Unterthiner, Bernhard Nessler, and Sepp
  Hochreiter.
\newblock Gans trained by a two time-scale update rule converge to a local nash
  equilibrium.
\newblock In I.~Guyon, U.~V. Luxburg, S.~Bengio, H.~Wallach, R.~Fergus,
  S.~Vishwanathan, and R.~Garnett, editors, \emph{Advances in Neural
  Information Processing Systems}, volume~30. Curran Associates, Inc., 2017.
\newblock URL
  \url{https://proceedings.neurips.cc/paper/2017/file/8a1d694707eb0fefe65871369074926d-Paper.pdf}.

\bibitem[Hu et~al.(2019)Hu, Mazzoli, and Wilkins]{WenHuMagneticImaging}
Wen Hu, Claudio Mazzoli, and Stuart Wilkins.
\newblock {New advances at CSX beamline in magnetic imaging (Conference
  Presentation)}.
\newblock In Barry Lai and Andrea Somogyi, editors, \emph{X-Ray Nanoimaging:
  Instruments and Methods IV}, volume 11112. International Society for Optics
  and Photonics, SPIE, 2019.
\newblock \doi{10.1117/12.2528058}.
\newblock URL \url{https://doi.org/10.1117/12.2528058}.

\bibitem[Jagatap and Hegde(2019)]{Jagatap2019}
Gauri Jagatap and Chinmay Hegde.
\newblock {Phase Retrieval using Untrained Neural Network Priors}.
\newblock \emph{NeurIPS workshop on Inverse Problems}, 2019.

\bibitem[Jurling and Fienup(2014)]{Jurling2014}
Alden~S. Jurling and James~R. Fienup.
\newblock {Applications of algorithmic differentiation to phase retrieval
  algorithms}.
\newblock \emph{Journal of the Optical Society of America A}, 31\penalty0
  (7):\penalty0 1348, 2014.
\newblock ISSN 1084-7529.
\newblock \doi{10.1364/josaa.31.001348}.

\bibitem[Kandel et~al.(2019)Kandel, Maddali, Allain, Hruszkewycz, Jacobsen, and
  Nashed]{Kandel2019}
Saugat Kandel, S.~Maddali, Marc Allain, Stephan~O. Hruszkewycz, Chris Jacobsen,
  and Youssef S.~G. Nashed.
\newblock {Using automatic differentiation as a general framework for
  ptychographic reconstruction}.
\newblock \emph{Optics Express}, 27\penalty0 (13):\penalty0 18653, 2019.
\newblock ISSN 1094-4087.
\newblock \doi{10.1364/oe.27.018653}.

\bibitem[Kikuta et~al.(1972)Kikuta, Aoki, Kosaki, and Kohra]{Holog2}
S.~Kikuta, S.~Aoki, S.~Kosaki, and K.~Kohra.
\newblock X-ray holography of lensless fourier-transform type.
\newblock \emph{Optics Communications}, 5\penalty0 (2):\penalty0 86 -- 89,
  1972.
\newblock ISSN 0030-4018.
\newblock \doi{https://doi.org/10.1016/0030-4018(72)90005-3}.
\newblock URL
  \url{http://www.sciencedirect.com/science/article/pii/0030401872900053}.

\bibitem[Kingma and Ba(2015)]{Kingma2015}
Diederik~P. Kingma and Jimmy~Lei Ba.
\newblock {Adam: A method for stochastic optimization}.
\newblock In \emph{3rd International Conference on Learning Representations,
  ICLR 2015 - Conference Track Proceedings}, 2015.

\bibitem[Latychevskaia(2019)]{Tatiana2019}
Tatiana Latychevskaia.
\newblock Reconstruction of missing information in diffraction patterns and
  holograms by iterative phase retrieval.
\newblock \emph{Optics Communications}, 452:\penalty0 56 -- 67, 2019.
\newblock ISSN 0030-4018.
\newblock \doi{https://doi.org/10.1016/j.optcom.2019.07.021}.
\newblock URL
  \url{http://www.sciencedirect.com/science/article/pii/S0030401819306054}.

\bibitem[Lempitsky et~al.(2018)Lempitsky, Vedaldi, and Ulyanov]{Ulyanov2018}
Victor Lempitsky, Andrea Vedaldi, and Dmitry Ulyanov.
\newblock {Deep Image Prior}.
\newblock In \emph{2018 IEEE/CVF Conference on Computer Vision and Pattern
  Recognition}, pages 9446--9454. IEEE, jun 2018.
\newblock ISBN 978-1-5386-6420-9.
\newblock \doi{10.1109/CVPR.2018.00984}.
\newblock URL
  \url{https://box.skoltech.ru/index.php/s/ib52BOoV58ztuPM{\#}pdfviewer
  https://ieeexplore.ieee.org/document/8579082/}.

\bibitem[Marchesini et~al.(2008)Marchesini, Boutet, Sakdinawat, Bogan, Bajt,
  Barty, Chapman, Frank, Hau-Riege, Sz{\"o}ke, Cui, Shapiro, Howells, Spence,
  Shaevitz, Lee, Hajdu, and Seibert]{MarchesiniURA}
Stefano Marchesini, S{\'e}bastien Boutet, {Anne E.} Sakdinawat, {Michael J.}
  Bogan, Sǎa Bajt, Anton Barty, {Henry N.} Chapman, Matthias Frank, {Stefan
  P.} Hau-Riege, Abraham Sz{\"o}ke, Congwu Cui, {David A.} Shapiro, {Malcolm
  R.} Howells, John Spence, {Joshua W.} Shaevitz, {Joanna Y.} Lee, Janos Hajdu,
  and {Marvin M.} Seibert.
\newblock Massively parallel x-ray holography.
\newblock \emph{Nature Photonics}, 2\penalty0 (9):\penalty0 560--563, September
  2008.
\newblock ISSN 1749-4885.
\newblock \doi{10.1038/nphoton.2008.154}.

\bibitem[McCann et~al.(2017)McCann, Jin, and Unser]{McCann2017}
Michael~T. McCann, Kyong~Hwan Jin, and Michael Unser.
\newblock {Convolutional neural networks for inverse problems in imaging: A
  review}.
\newblock \emph{IEEE Signal Processing Magazine}, 34\penalty0 (6):\penalty0
  85--95, 2017.
\newblock ISSN 10535888.
\newblock \doi{10.1109/MSP.2017.2739299}.

\bibitem[Meinhardt et~al.(2017)Meinhardt, Moeller, Hazirbas, and
  Cremers]{Meinhardt2017}
Tim Meinhardt, Michael Moeller, Caner Hazirbas, and Daniel Cremers.
\newblock {Learning Proximal Operators: Using Denoising Networks for
  Regularizing Inverse Imaging Problems}.
\newblock In \emph{2017 IEEE International Conference on Computer Vision
  (ICCV)}, number~12, pages 1799--1808. IEEE, oct 2017.
\newblock ISBN 978-1-5386-1032-9.
\newblock \doi{10.1109/ICCV.2017.198}.
\newblock URL \url{http://ieeexplore.ieee.org/document/8237460/}.

\bibitem[Metzler et~al.(2018)Metzler, Schniter, Veeraraghavan, and
  Baraniuk]{Metzler2018}
Christopher~A. Metzler, Philip Schniter, Ashok Veeraraghavan, and Richard~G.
  Baraniuk.
\newblock {prDeep: Robust phase retrieval with a flexible deep network}.
\newblock \emph{35th International Conference on Machine Learning, ICML 2018},
  8:\penalty0 5654--5663, 2018.

\bibitem[Miao et~al.(1999)Miao, Charalambous, Kirz, and Sayre]{CDI-orig}
Jianwei Miao, Pambos Charalambous, Janos Kirz, and David Sayre.
\newblock {Extending the methodology of X-ray crystallography to allow imaging
  of micrometre-sized non-crystalline specimens}.
\newblock \emph{Nature}, 400:\penalty0 342--344, jul 1999.
\newblock \doi{10.1038/22498}.

\bibitem[Nashed et~al.(2017)Nashed, Peterka, Deng, and Jacobsen]{Nashed2017}
Youssef~S.G. Nashed, Tom Peterka, Junjing Deng, and Chris Jacobsen.
\newblock {Distributed Automatic Differentiation for Ptychography}.
\newblock \emph{Procedia Computer Science}, 108:\penalty0 404--414, 2017.
\newblock ISSN 18770509.
\newblock \doi{10.1016/j.procs.2017.05.101}.
\newblock URL \url{http://dx.doi.org/10.1016/j.procs.2017.05.101}.

\bibitem[Nene et~al.(1996)Nene, Nayar, and Murase]{Nene1996}
S.~A. Nene, S.~K. Nayar, and H.~Murase.
\newblock {Columbia Object Image Library}.
\newblock \emph{Technical Report CUCS-006-96}, February, 1996.

\bibitem[Ongie et~al.(2020)Ongie, Jalal, Metzler, Baraniuk, Dimakis, and
  Willett]{Ongie2020}
Gregory Ongie, Ajil Jalal, Christopher~A. Metzler, Richard~G. Baraniuk,
  Alexandros~G. Dimakis, and Rebecca Willett.
\newblock {Deep Learning Techniques for Inverse Problems in Imaging}.
\newblock \emph{IEEE Journal on Selected Areas in Information Theory},
  1\penalty0 (1):\penalty0 39--56, may 2020.
\newblock ISSN 2641-8770.
\newblock \doi{10.1109/JSAIT.2020.2991563}.
\newblock URL \url{https://ieeexplore.ieee.org/document/9084378/}.

\bibitem[Paszke et~al.(2019)Paszke, Gross, Massa, Lerer, Bradbury, Chanan,
  Killeen, Lin, Gimelshein, Antiga, Desmaison, Kopf, Yang, DeVito, Raison,
  Tejani, Chilamkurthy, Steiner, Fang, Bai, and Chintala]{pytorch2019}
Adam Paszke, Sam Gross, Francisco Massa, Adam Lerer, James Bradbury, Gregory
  Chanan, Trevor Killeen, Zeming Lin, Natalia Gimelshein, Luca Antiga, Alban
  Desmaison, Andreas Kopf, Edward Yang, Zachary DeVito, Martin Raison, Alykhan
  Tejani, Sasank Chilamkurthy, Benoit Steiner, Lu~Fang, Junjie Bai, and Soumith
  Chintala.
\newblock Pytorch: An imperative style, high-performance deep learning library.
\newblock In H.~Wallach, H.~Larochelle, A.~Beygelzimer, F.~d\' Alch\'{e}-Buc,
  E.~Fox, and R.~Garnett, editors, \emph{Advances in Neural Information
  Processing Systems 32}, pages 8024--8035. Curran Associates, Inc., 2019.
\newblock URL
  \url{http://papers.neurips.cc/paper/9015-pytorch-an-imperative-style-high-performance-deep-learning-library.pdf}.

\bibitem[Rivenson et~al.(2018)Rivenson, Zhang, G{\"{u}}naydın, Teng, and
  Ozcan]{Rivenson2018}
Yair Rivenson, Yibo Zhang, Harun G{\"{u}}naydın, Da~Teng, and Aydogan Ozcan.
\newblock {Phase recovery and holographic image reconstruction using deep
  learning in neural networks}.
\newblock \emph{Light: Science and Applications}, 7\penalty0 (2):\penalty0
  17141, 2018.
\newblock ISSN 20477538.
\newblock \doi{10.1038/lsa.2017.141}.

\bibitem[Romano et~al.(2017)Romano, Elad, and Milanfar]{Romano2017}
Yaniv Romano, Michael Elad, and Peyman Milanfar.
\newblock {The little engine that could: Regularization by Denoising (RED)}.
\newblock \emph{SIAM Journal on Imaging Sciences}, 10\penalty0 (4):\penalty0
  1804--1844, 2017.
\newblock ISSN 19364954.
\newblock \doi{10.1137/16M1102884}.

\bibitem[Salditt et~al.(2020)Salditt, Egner, and Luke]{CDI-Book}
Tim Salditt, Alexander Egner, and D.~Russell Luke, editors.
\newblock \emph{Nanoscale Photonic Imaging}.
\newblock Springer International Publishing, 2020.
\newblock \doi{10.1007/978-3-030-34413-9}.
\newblock URL \url{https://doi.org/10.1007/978-3-030-34413-9}.

\bibitem[Saliba et~al.(2012)Saliba, Latychevskaia, Longchamp, and
  Fink]{FT-Cambridge}
M~Saliba, T~Latychevskaia, J~Longchamp, and H~Fink.
\newblock {Fourier Transform Holography: A Lensless Non-Destructive Imaging
  Technique}.
\newblock \emph{Microscopy and Microanalysis}, 18\penalty0 (S2):\penalty0
  564--565, 2012.
\newblock \doi{10.1017/S1431927612004679}.

\bibitem[Shamshad and Ahmed(2018)]{Shamshad2018}
Fahad Shamshad and Ali Ahmed.
\newblock {Robust Compressive Phase Retrieval via Deep Generative Priors}.
\newblock \emph{arXiv preprint}, 1808.05854:\penalty0 1--19, 2018.
\newblock URL \url{http://arxiv.org/abs/1808.05854}.

\bibitem[Shechtman et~al.(2015)Shechtman, Eldar, Cohen, Chapman, Miao, and
  Segev]{eldar-review}
Yoav Shechtman, Yonina~C Eldar, Oren Cohen, Henry~Nicholas Chapman, Jianwei
  Miao, and Mordechai Segev.
\newblock {Phase retrieval with application to optical imaging: a contemporary
  overview}.
\newblock \emph{IEEE signal processing magazine}, 32\penalty0 (3):\penalty0
  87--109, 2015.
\newblock \doi{10.1109/MSP.2014.2352673}.

\bibitem[Shi et~al.(2020)Shi, Lian, and Chang]{Shi2020}
Baoshun Shi, Qiusheng Lian, and Huibin Chang.
\newblock {Deep prior-based sparse representation model for diffraction
  imaging: A plug-and-play method}.
\newblock \emph{Signal Processing}, 168:\penalty0 107350, 2020.
\newblock ISSN 01651684.
\newblock \doi{10.1016/j.sigpro.2019.107350}.
\newblock URL \url{https://doi.org/10.1016/j.sigpro.2019.107350}.

\bibitem[Simonyan and Zisserman(2015)]{simonyan2015deep}
Karen Simonyan and Andrew Zisserman.
\newblock Very deep convolutional networks for large-scale image recognition.
\newblock In \emph{3rd International Conference on Learning Representations,
  ICLR 2015 - Conference Track Proceedings}, 2015.

\bibitem[Thibault and Guizar-Sicairos(2012)]{Thibault2012}
P.~Thibault and M.~Guizar-Sicairos.
\newblock {Maximum-likelihood refinement for coherent diffractive imaging}.
\newblock \emph{New Journal of Physics}, 14, 2012.
\newblock ISSN 13672630.
\newblock \doi{10.1088/1367-2630/14/6/063004}.

\bibitem[Tramel et~al.(2016)Tramel, Manoel, Caltagirone, Gabri{\'{e}}, and
  Krzakala]{Tramel2016}
Eric~W. Tramel, Andre Manoel, Francesco Caltagirone, Marylou Gabri{\'{e}}, and
  Florent Krzakala.
\newblock {Inferring sparsity: Compressed sensing using generalized restricted
  Boltzmann machines}.
\newblock In \emph{2016 IEEE Information Theory Workshop (ITW)}, pages
  265--269. IEEE, sep 2016.
\newblock ISBN 978-1-5090-1090-5.
\newblock \doi{10.1109/ITW.2016.7606837}.
\newblock URL \url{http://ieeexplore.ieee.org/document/7606837/}.

\bibitem[Uelwer et~al.(2019)Uelwer, Oberstra{\ss}, and Harmeling]{Uelwer2019}
Tobias Uelwer, Alexander Oberstra{\ss}, and Stefan Harmeling.
\newblock {Phase Retrieval using Conditional Generative Adversarial Networks}.
\newblock \emph{arXiv preprint}, 1912.04981, 2019.
\newblock URL \url{http://arxiv.org/abs/1912.04981}.

\bibitem[Wang et~al.(2020{\natexlab{a}})Wang, Bian, Wang, Lyu, Pedrini, Osten,
  Barbastathis, and Situ]{Wang2020a}
Fei Wang, Yaoming Bian, Haichao Wang, Meng Lyu, Giancarlo Pedrini, Wolfgang
  Osten, George Barbastathis, and Guohai Situ.
\newblock {Phase imaging with an untrained neural network}.
\newblock \emph{Light: Science and Applications}, 9\penalty0 (1),
  2020{\natexlab{a}}.
\newblock ISSN 20477538.
\newblock \doi{10.1038/s41377-020-0302-3}.
\newblock URL \url{http://dx.doi.org/10.1038/s41377-020-0302-3}.

\bibitem[Wang et~al.(2020{\natexlab{b}})Wang, Sun, and Fleischer]{Wang2020}
Yaotian Wang, Xiaohang Sun, and Jason~W. Fleischer.
\newblock {When deep denoising meets iterative phase retrieval}.
\newblock In \emph{International Conference on Machine Learning},
  2020{\natexlab{b}}.
\newblock URL \url{http://arxiv.org/abs/2003.01792}.

\bibitem[Wang et~al.(2004)Wang, Bovik, Sheikh, and Simoncelli]{Wang2004}
Z.~Wang, A.C. Bovik, H.R. Sheikh, and E.P. Simoncelli.
\newblock {Image Quality Assessment: From Error Visibility to Structural
  Similarity}.
\newblock \emph{IEEE Transactions on Image Processing}, 13\penalty0
  (4):\penalty0 600--612, apr 2004.
\newblock ISSN 1057-7149.
\newblock \doi{10.1109/TIP.2003.819861}.
\newblock URL \url{http://ieeexplore.ieee.org/document/1284395/}.

\bibitem[Wen et~al.(2012)Wen, Yang, Liu, and Marchesini]{StefanoPtychography}
Zaiwen Wen, Chao Yang, Xin Liu, and Stefano Marchesini.
\newblock Alternating direction methods for classical and ptychographic phase
  retrieval.
\newblock \emph{Inverse Problems}, 28\penalty0 (11):\penalty0 115010, oct 2012.
\newblock \doi{10.1088/0266-5611/28/11/115010}.
\newblock URL \url{https://doi.org/10.1088%2F0266-5611%2F28%2F11%2F115010}.

\end{thebibliography}

\newpage
\appendix

\section{Optical Laser Coherent Imaging Experiment Details}
\label{app:realdata}
\begin{figure} 
    \centering
    \subfigure[\label{fig:realdata-meas}]{\includegraphics[trim= 70 0 70 0, width=0.3\textwidth]{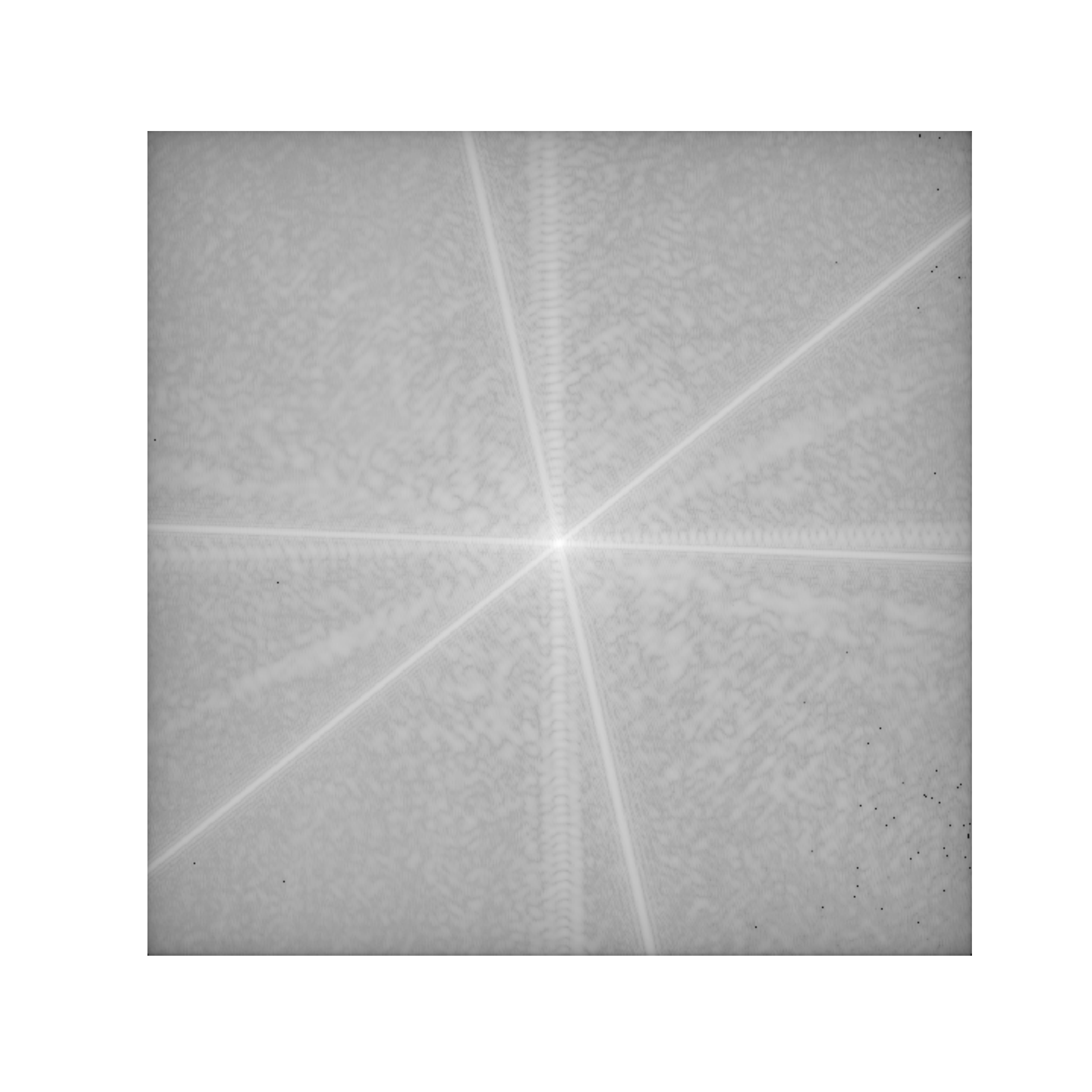}}
    \hspace{1cm}
    \subfigure[ \label{fig:realdata-ref}]{\includegraphics[trim= 100  0 100 0, clip, width=0.3\textwidth]{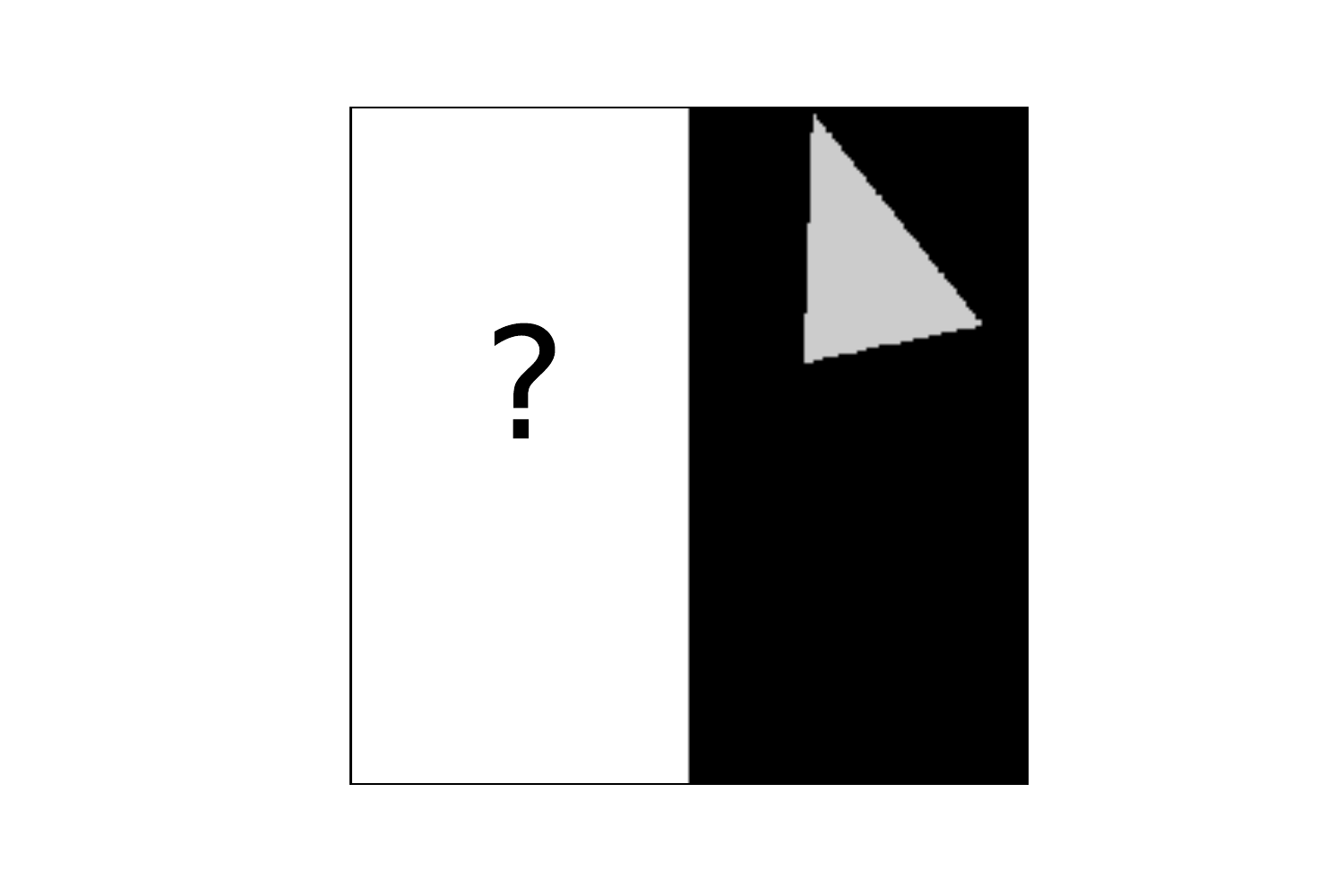}}
    \hspace{0.1cm}
    \subfigure[\label{fig:realdata-full}]{\includegraphics[trim= -65  0 -60 0, clip, width=0.3\textwidth]{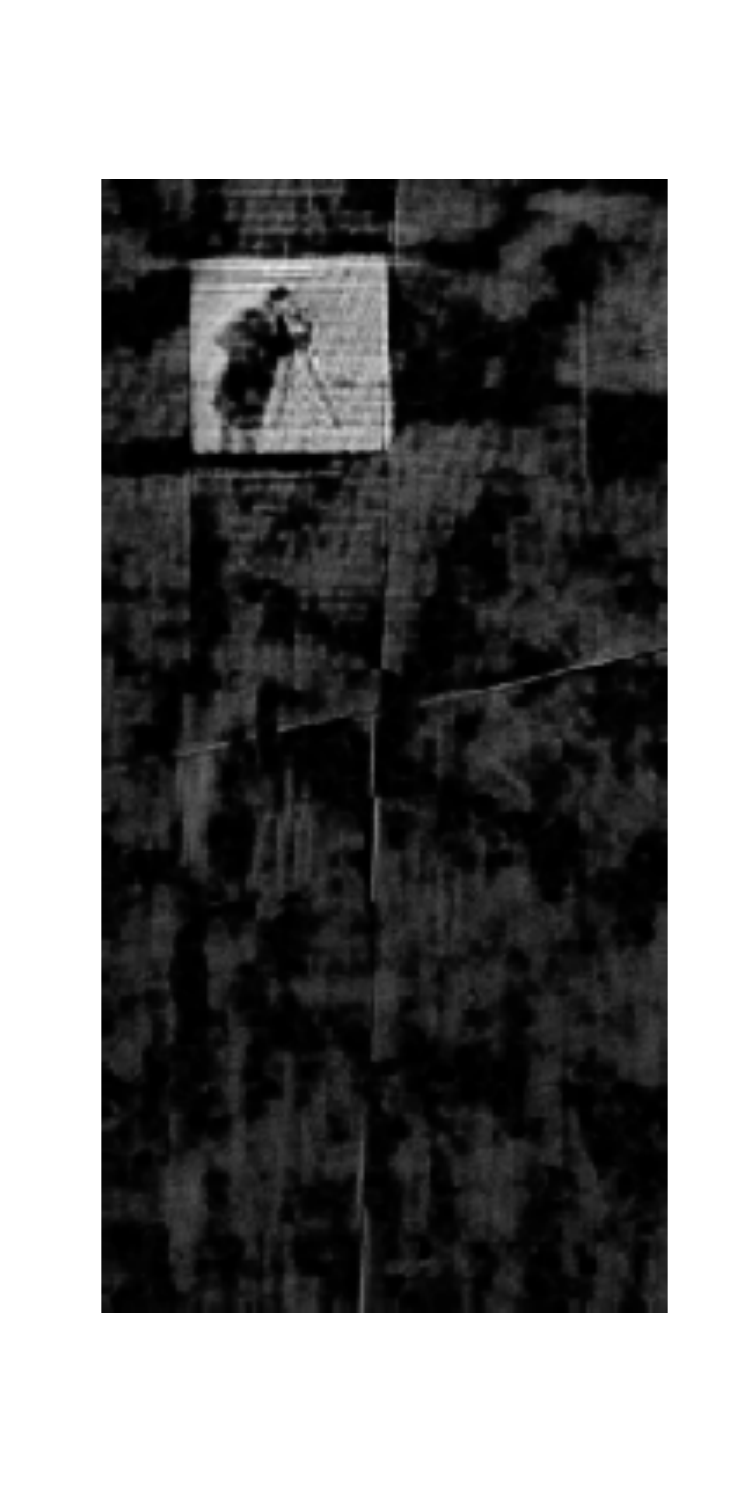}}
    \caption{\changes{Reconstruction from experimental data.
    (a) Measurements  ($1024\times1024$ pixels), (b) Reference positioning ($512\times512$ pixels), (c) Full reconstruction for \hpdd-TV 
    ($512\times256$ pixels) 
    \label{fig:realdata}}}
\end{figure}

\changes{We follow the same preprocessing steps on the raw data as in \cite{HERALDO_experimental}: namely, averaging over multiple exposures and frames, removing detector artifacts, and attenuating higher frequencies. The $1024\times1024$ processed magnitude measurements are displayed in Figure \ref{fig:realdata-meas}. Considering an oversampling ratio equal to 2, we assume half of a $512\times512$-pixel frame known, including the triangular reference, and reconstruct the remaining half. The deep decoder includes 3 layers and 256 channels. Figure \ref{fig:realdata-crop} presents a zoom taken on the CAMERA image reconstructed close to the top left corner of the frame for each methodm (see Figure \ref{fig:realdata-full} for the full frame on the example of \hpdd-TV). }

\section{Additional information for numerical experiments}
\label{app:exp}
\begin{figure} 
    \centering
    {\includegraphics[width=\textwidth]{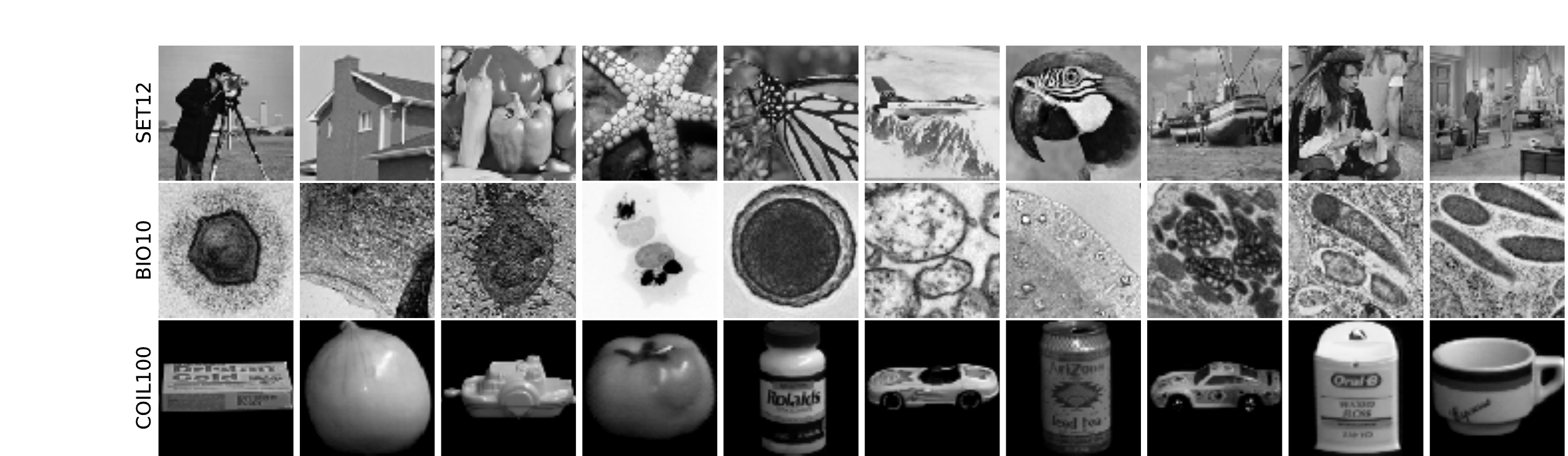}}
    \caption{Example images from datsets used in the experiments of this paper: SET12 (128x128 pixels), BIO10 (256x256 pixels) and COIL100 (128x128pixels)\label{fig:datasets}}
\end{figure}

In this section we provide additional figures for experiments presented in the main text as well as hyperparameters settings. 
Figure~\ref{fig:datasets} display sample images of the three datasets used for experiments. Table \ref{tab:dd-params} gathers deep decoder hyperparameters selected.

\begin{table}
    \centering
    \begin{tabular}{|l|c|c|c|c|c|}
  \hline
  Noise & $N_p =  1000$ & $N_p = 100$ & $N_p = 10$ & $N_p = 1$ & $ N_p = 0.1$\\
  \hline
  SET 12 & d2 c128 & d3 c128 & d2 c128 & d1 c128 & d1 c128 \\
  COIL100 & d2 c128 & d2 c128 & d2 c128 & d1 c128 & d1 c128 \\
  BIO10 & d3 c128 & d3 c128 & d2 c128 & d2 c128 & d2 c128 \\
  \hline
  $\#$ iterations & 10000 & 10000 & 5000 & 2500 & 1250 \\
  \hline
\end{tabular}
    \caption{Depth, number of channels, and number of optimizer steps used for the deep decoder in all experiments.}
    \label{tab:dd-params}
\end{table}

\begin{figure} 
    {\includegraphics[width=0.82\textwidth]{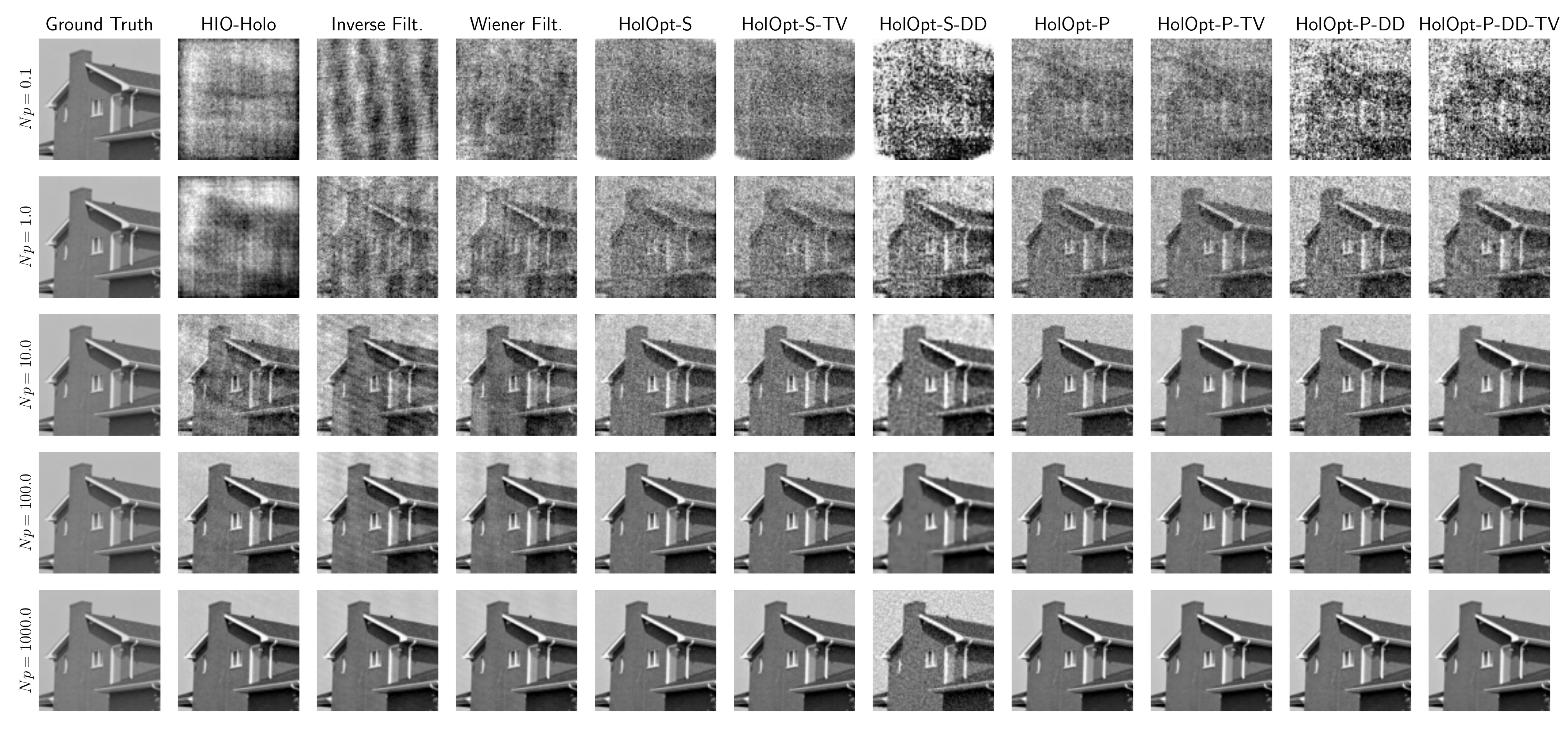}}
    \centering
    {\includegraphics[width=0.82\textwidth]{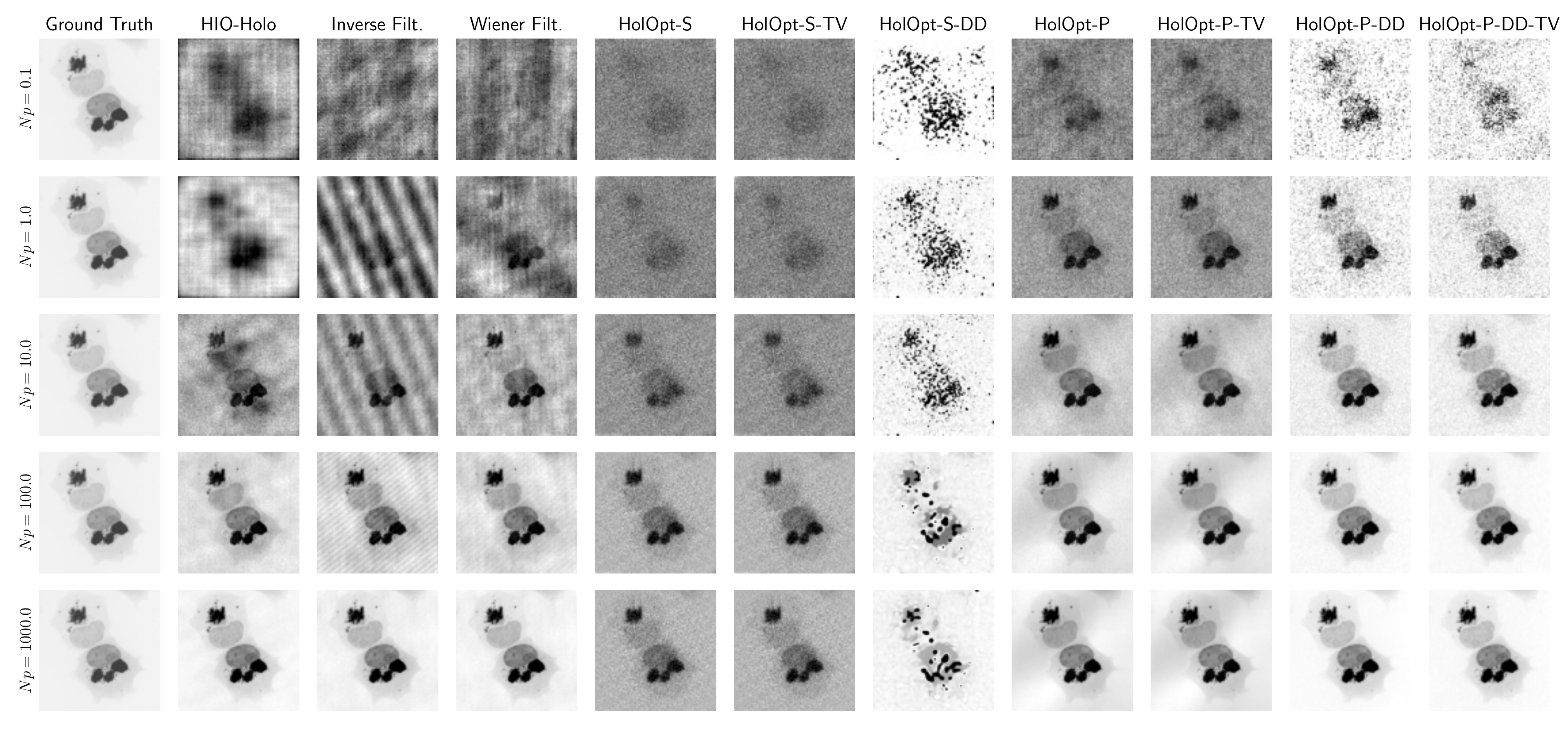}}
    \centering 
    {\includegraphics[width=0.82\textwidth]{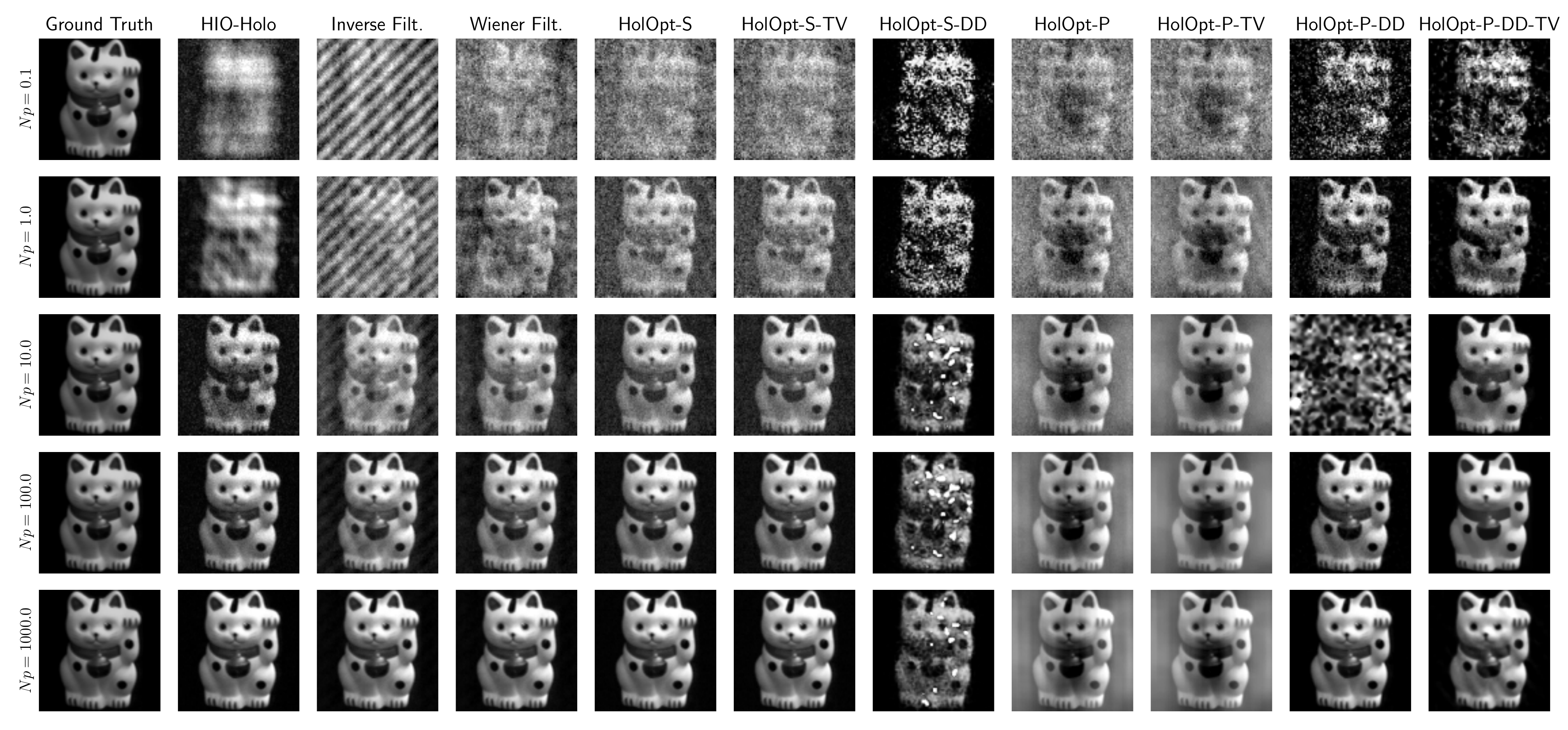}}
 \caption{ \label{fig:denoising-coil100-visual}
     Comparing reconstructions across algorithms and noise levels on a sample of images from SET12, BIO10 and COIL100 with a 
    binary random reference and without beamstop. Same as Figure \ref{fig:denoising-camera-visual}. \changes{Results with TV regularization were added.}}
\end{figure}

\begin{figure} 
    \centering
    {\includegraphics[width=\textwidth]{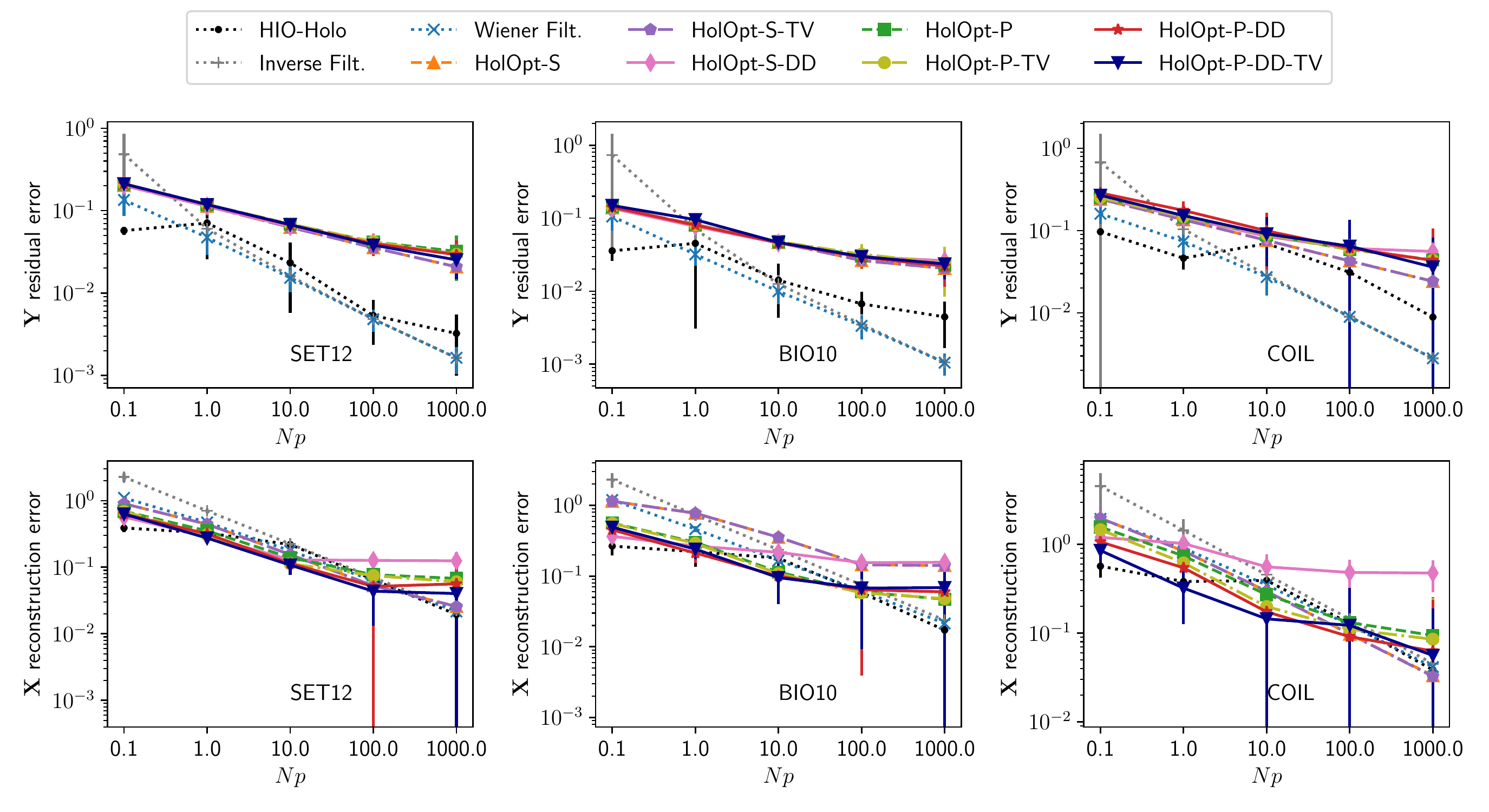}}
    \caption{\label{fig:denoising-graph-l2} Relative mean squared errors on the reconstructions and residual mean squared error averaged over the datasets. These are different metrics from the SSIM scores presented in Figure \ref{fig:denoising-graph-ssim} to analyze the results of the experiment presented of Section \ref{sec:denoising}. \changes{Results with TV regularization where added.}}
\end{figure}

\subsection{Noisy reconstruction experiment}
Figures \ref{fig:denoising-coil100-visual} and \ref{fig:denoising-graph-l2} present results for the experiment described in \ref{sec:denoising} of the main text. The superior visual quality of our algorithms is confirmed on samples form BIO10 and COIL100. 
In Figure \ref{fig:denoising-graph-l2} we compare methods in terms of residual $\ell2$ error minimization and MSE. These measure appear less informative of the perceived similarity of ground truth and reconstructed images than SSIM. 
In particular, it should be observed that the residual error reflects performance
neither perceptually nor according to the other error metrics.

\begin{figure} 
    \centering
    {\includegraphics[trim = 0 0 0 15,  clip,width=0.49\textwidth]{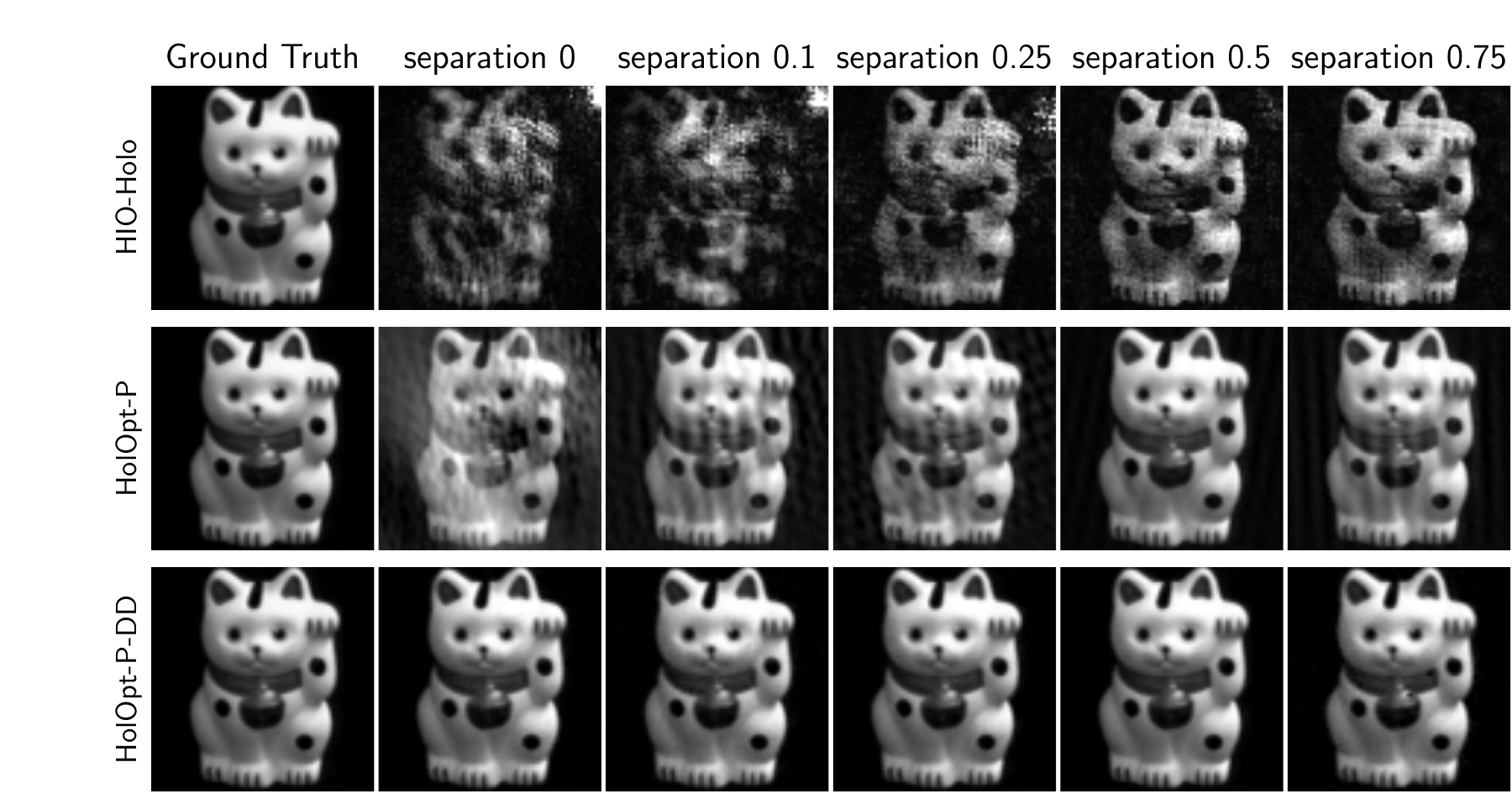}}
    {\includegraphics[trim = 0 0 0 15,  clip,width=0.49\textwidth]{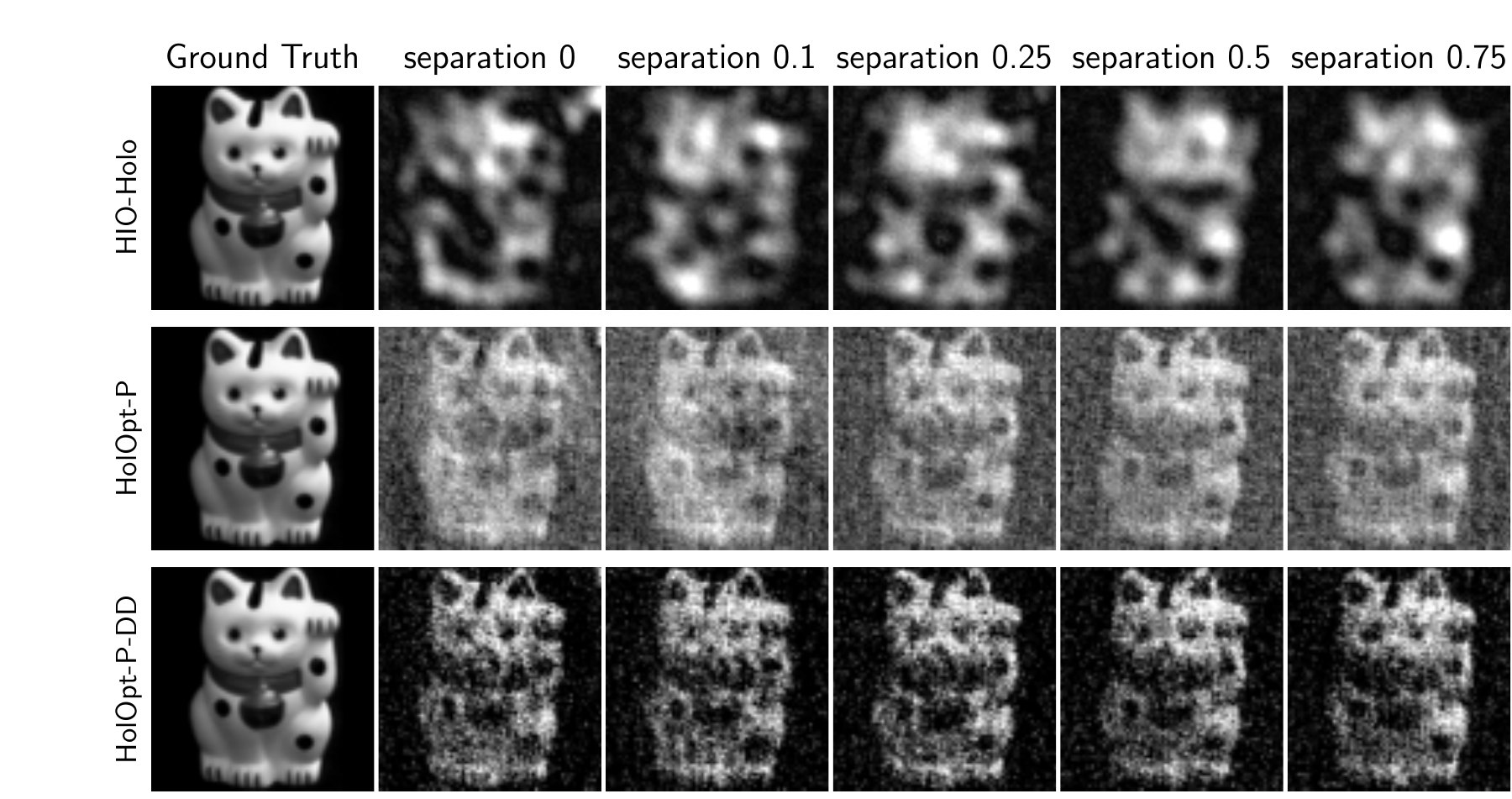}}\\
    \vspace{0.01\textwidth}
    {\includegraphics[trim = 0 0 0 15,  clip,width=0.49\textwidth]{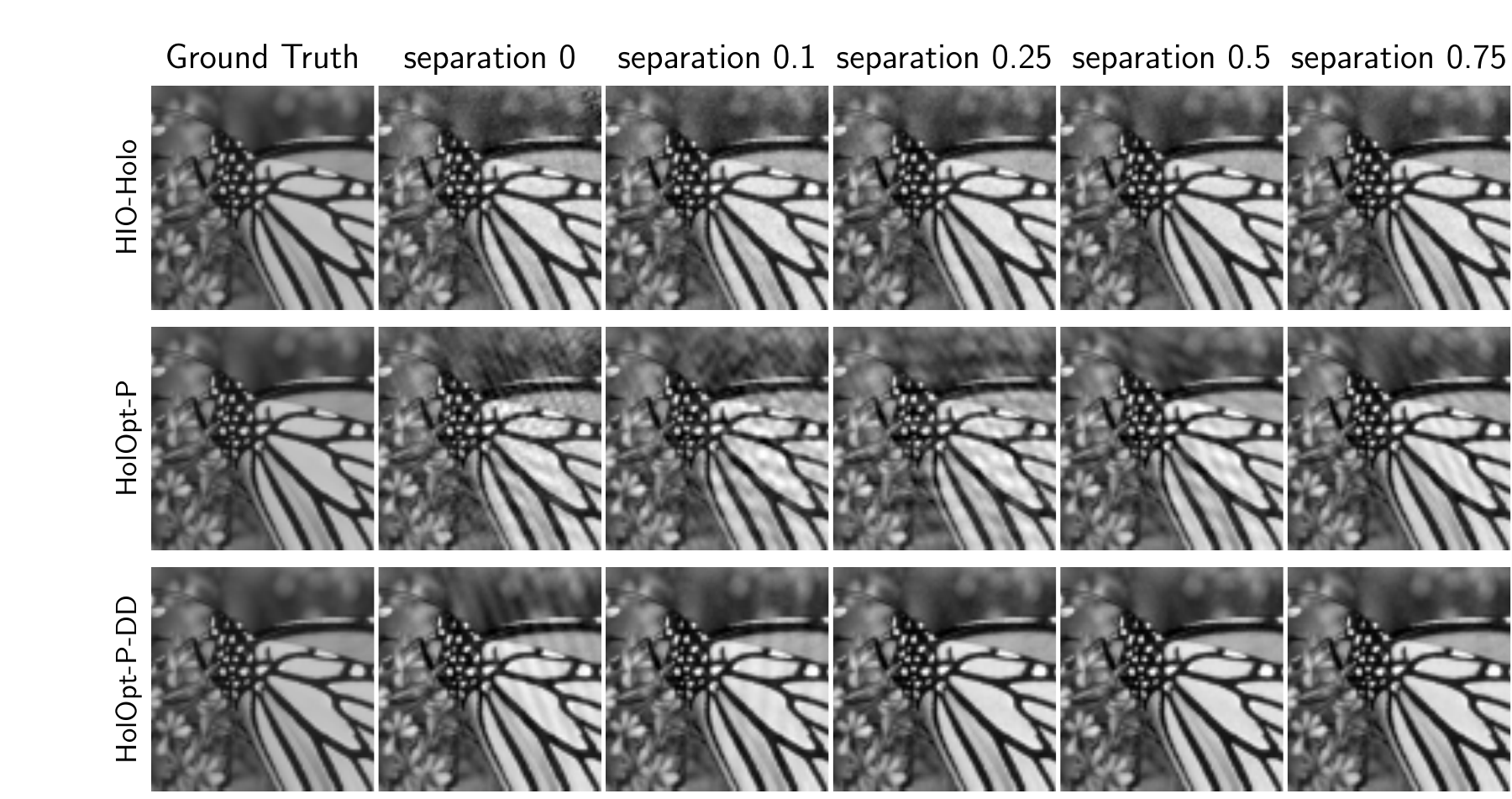}}
    {\includegraphics[trim = 0 0 0 15,  clip,width=0.49\textwidth]{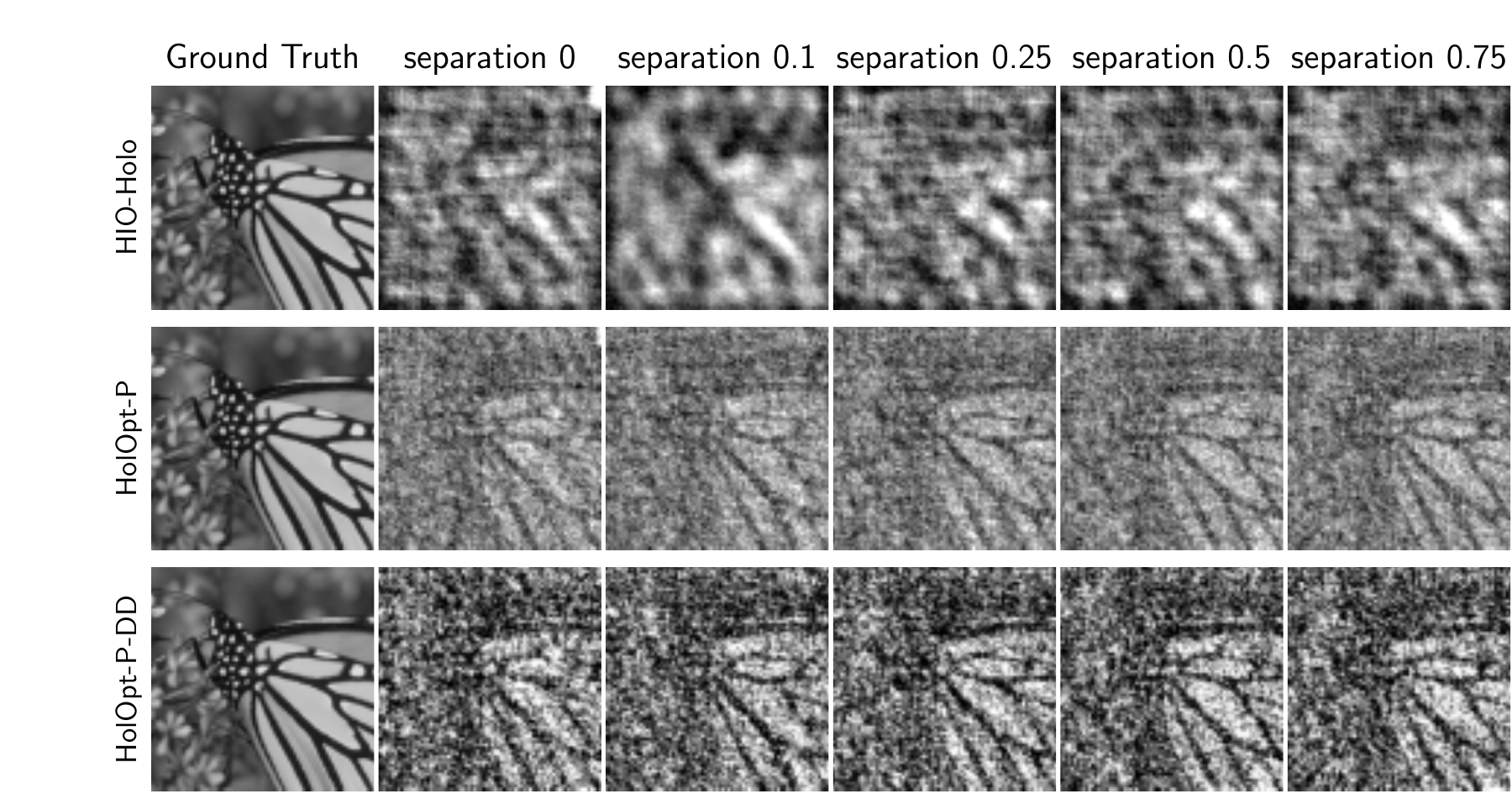}}\\
    \vspace{0.01\textwidth}
    \subfigure[$N_p = 100$ photons / pixel]
    {\includegraphics[trim = 0 0 0 15,  clip,width=0.49\textwidth]{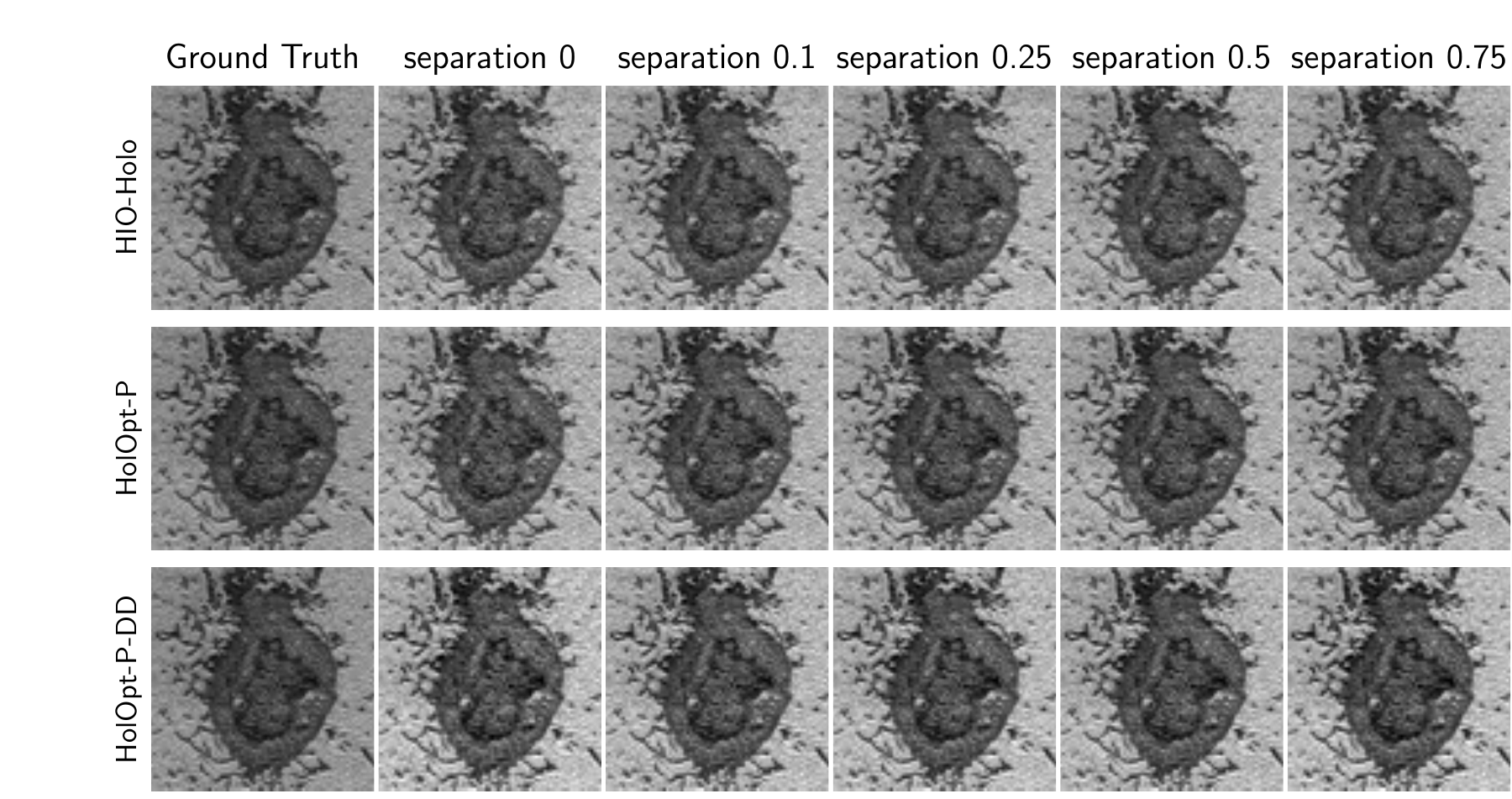}}
    \subfigure[$N_p = 0.1$ photons / pixel]
    {\includegraphics[trim = 0 0 0 15,  clip,width=0.49\textwidth]{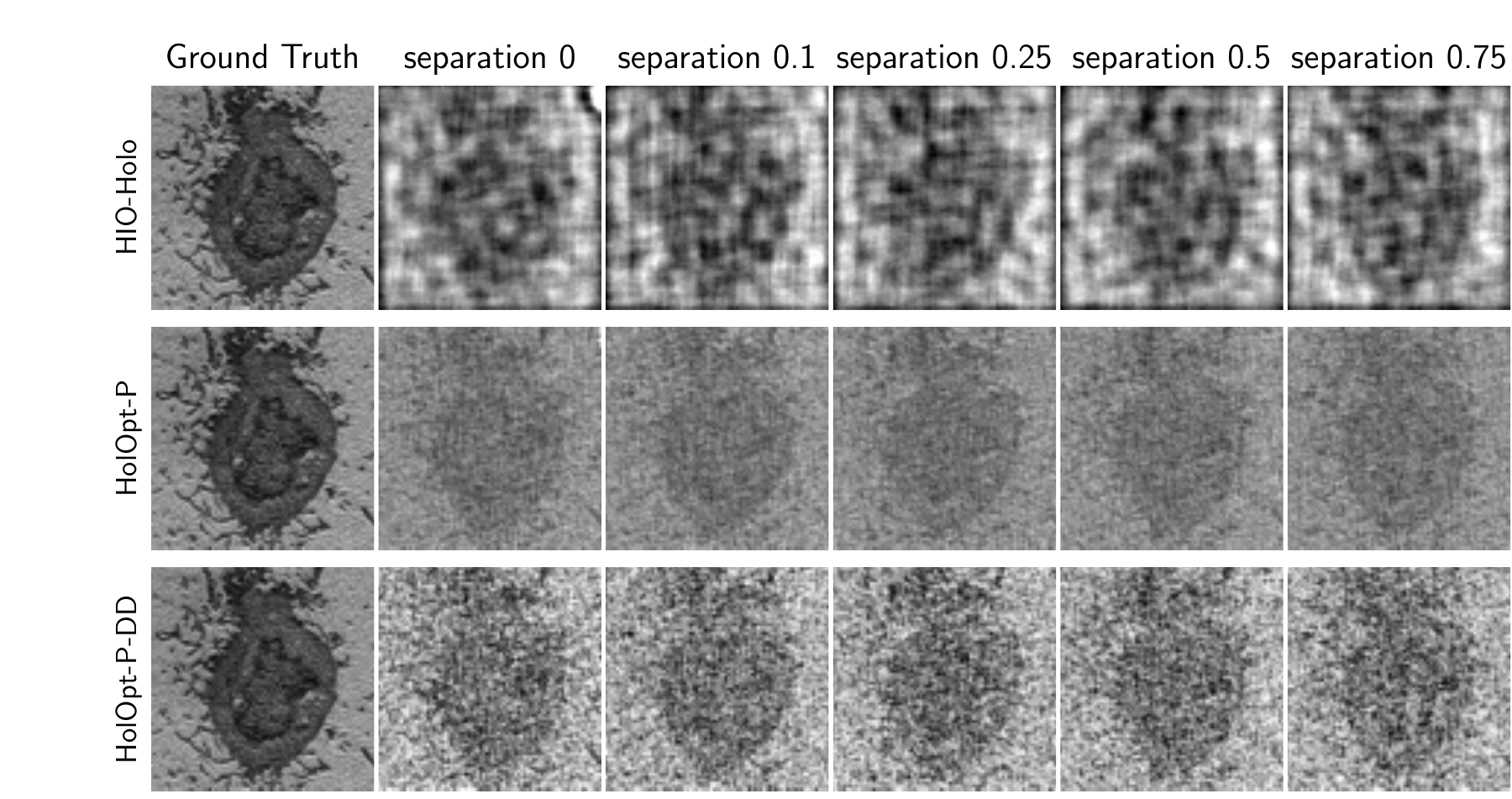}}
    \caption{\label{fig:sep-visual-np01-np100} {Reconstructed images as a function of the relative separation as described in caption of Figure \ref{fig:sep-ssim-graph}. Images correspond to the best of 10 runs in terms of residual error.}}
\end{figure}

\begin{figure}
    \centering
    {\includegraphics[trim = 0 0 0 15,  clip,width=0.49\textwidth]{figures/msml_rev/rev_msml_paper_sep_re_visual_cutoff_COIL13_noisepoisson10_0.pdf}}
    {\includegraphics[trim = 0 0 0 15,  clip,width=0.49\textwidth]{figures/msml_rev/rev_msml_paper_sep_re_visual_cutoff_COIL13_noisepoisson1_0.pdf}}\\
    \vspace{0.01\textwidth}
    {\includegraphics[trim = 0 0 0 15,  clip,width=0.49\textwidth]{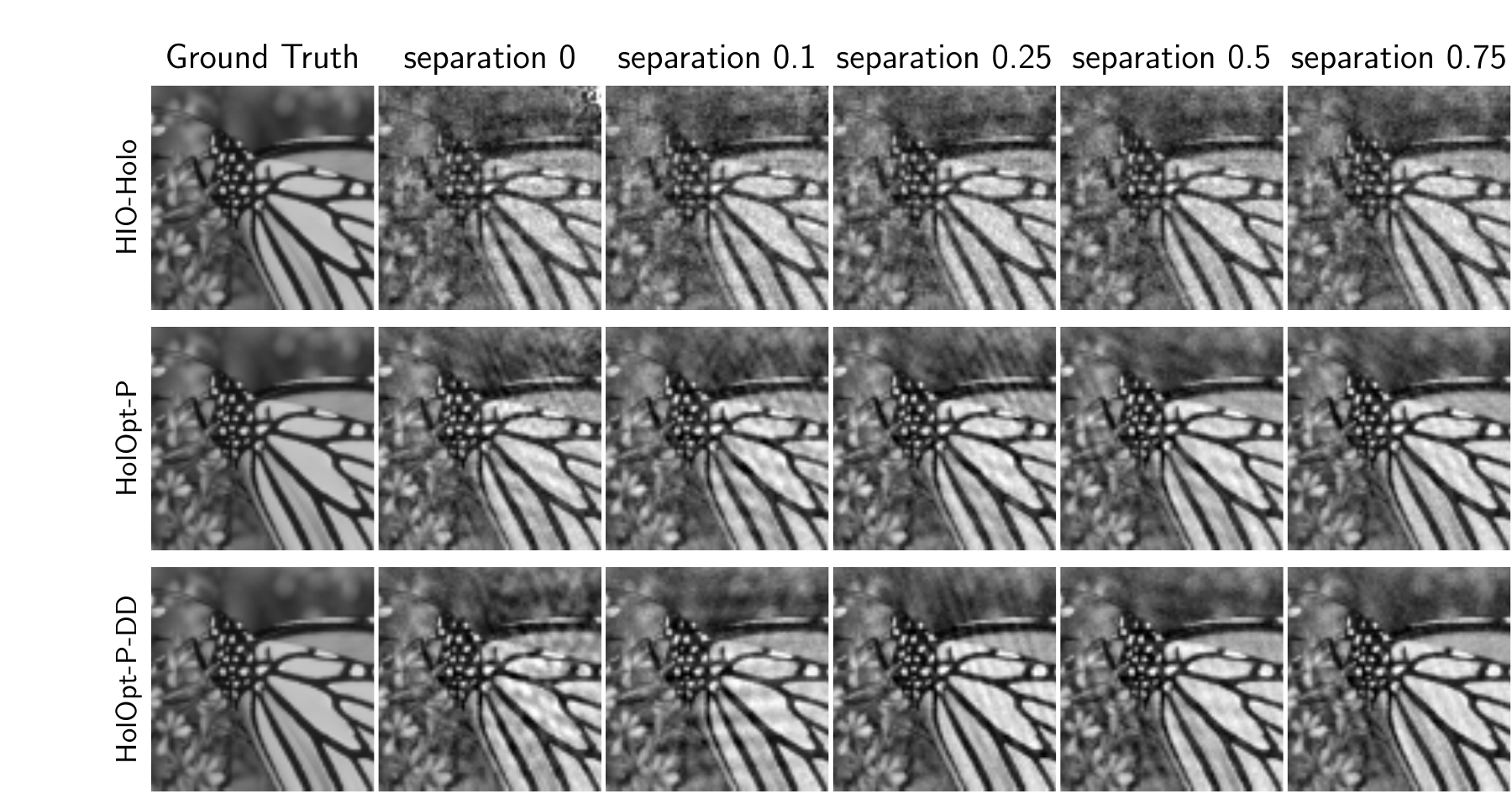}}
    {\includegraphics[trim = 0 0 0 15,  clip,width=0.49\textwidth]{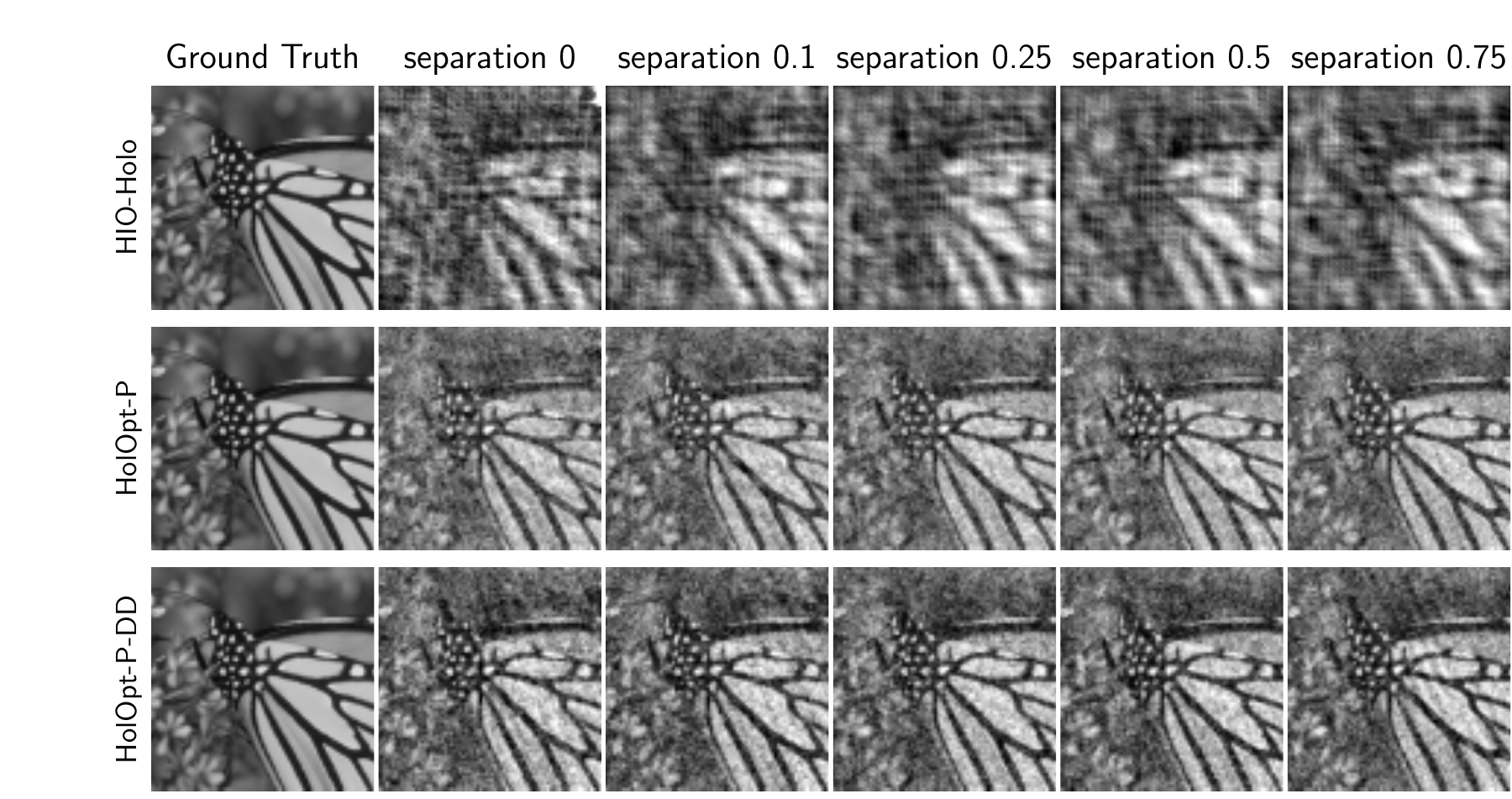}}\\
    \vspace{0.01\textwidth}
    \subfigure[$N_p = 10$ photons / pixel]
    {\includegraphics[trim = 0 0 0 15,  clip,width=0.49\textwidth]{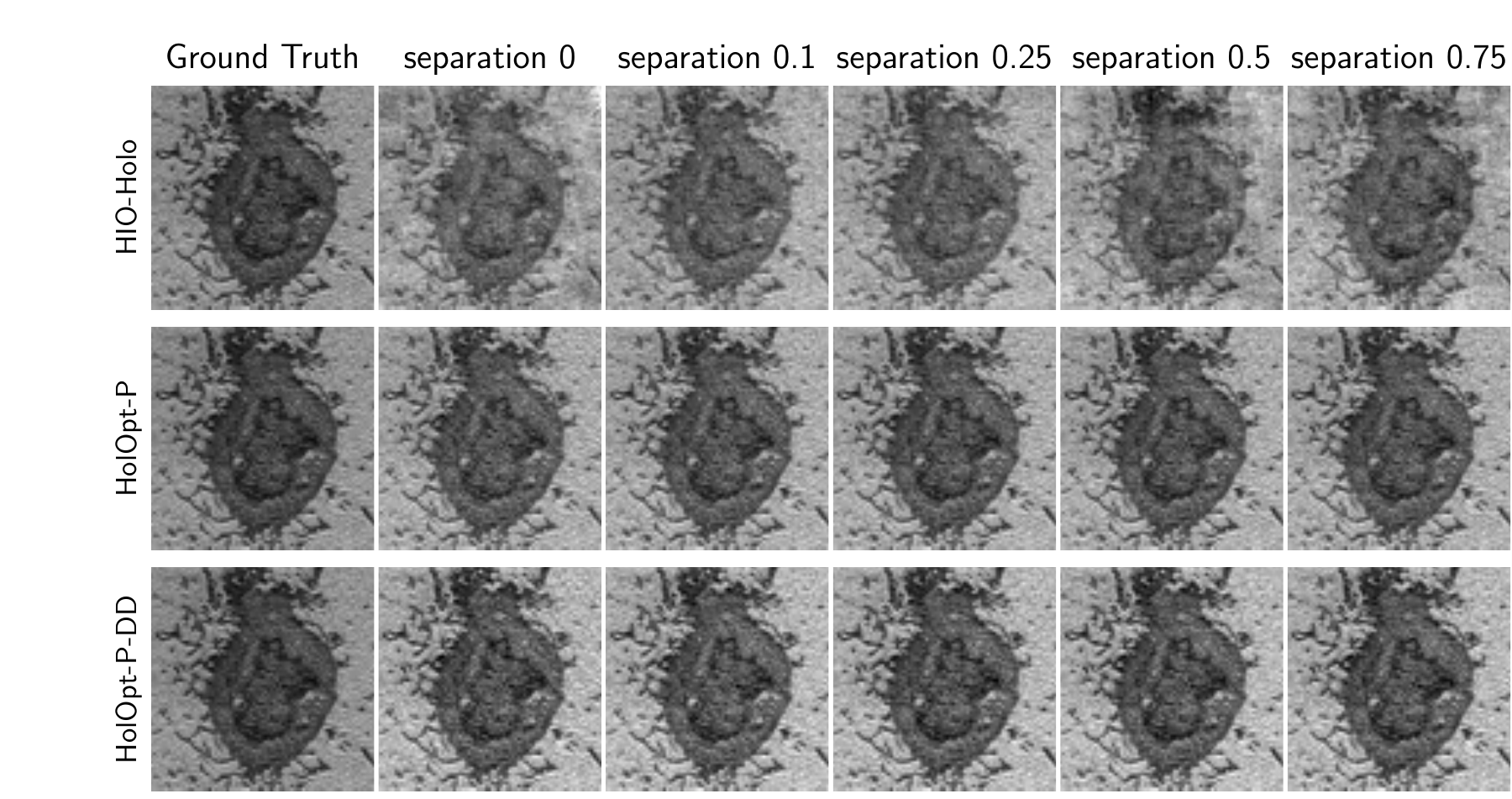}}
    \subfigure[$N_p = 1$ photons / pixel]
    {\includegraphics[trim = 0 0 0 15,  clip,width=0.49\textwidth]{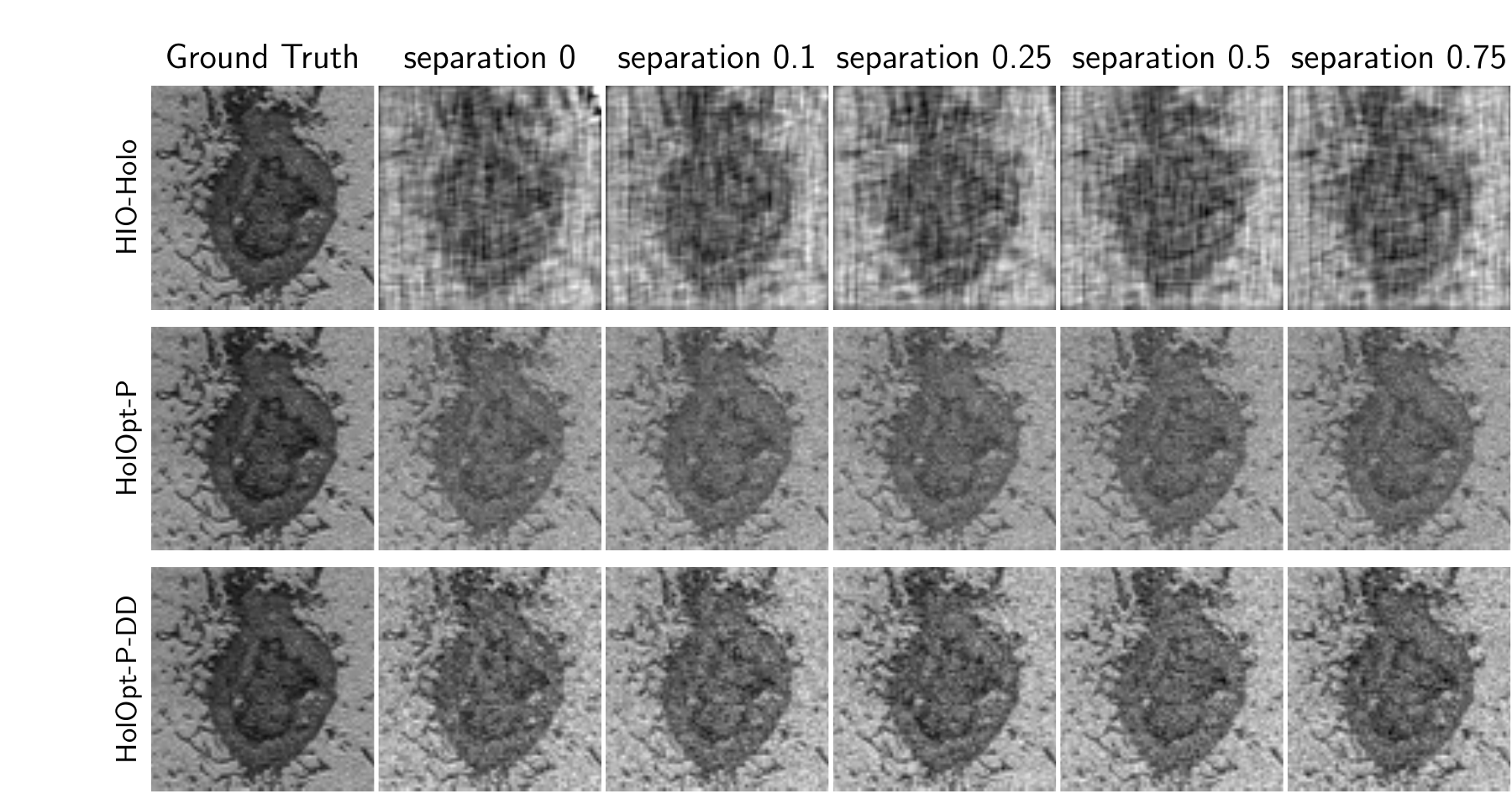}}
    \caption{\label{fig:sep-visual-np1-np10-all} {Reconstructions for photon counts $N_p = 10$ (left) and $N_p = 1$ (right) with a $0.1 m \times 0.1 m $ binary random reference as a function of the relative separation. A separation of $0.5$ implies that the left-most non-zero pixel of the reference is $0.5n$ pixels away form the image. 
    }}
\end{figure}

\begin{figure}
    {\includegraphics[trim = 0 0 0 0,  clip,width=1.\textwidth]{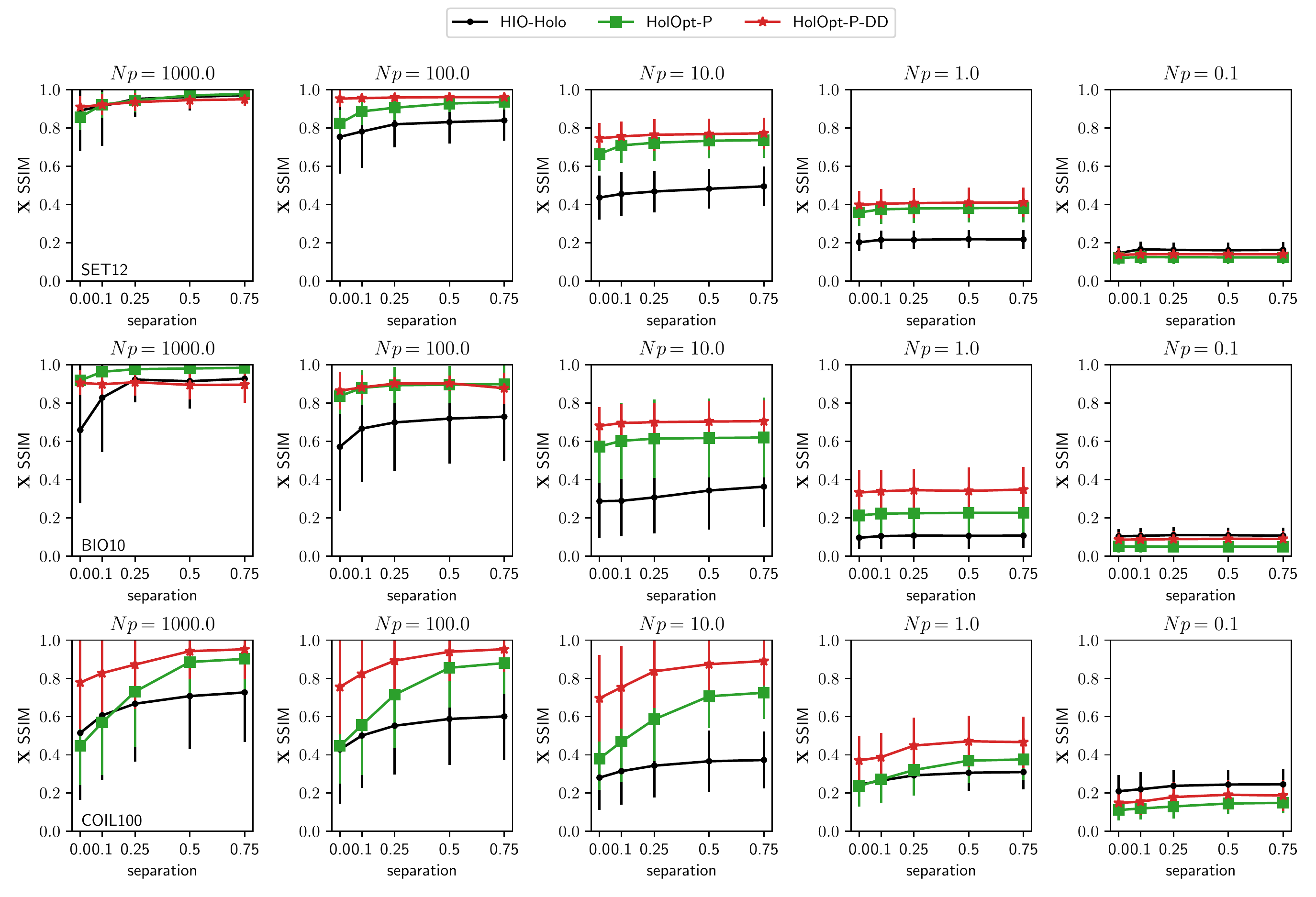}}
    \caption{\label{fig:sep-ssim-graph-all} Reconstruction SSIM for decreasing photon counts $N_p$ as a function of the relative separation (see caption of Figure \ref{fig:sep-visual-np1-np10-all}).
    }
    \vspace{-6pt}
\end{figure}

\begin{figure} 
    \centering
    \includegraphics[width=1.\textwidth]{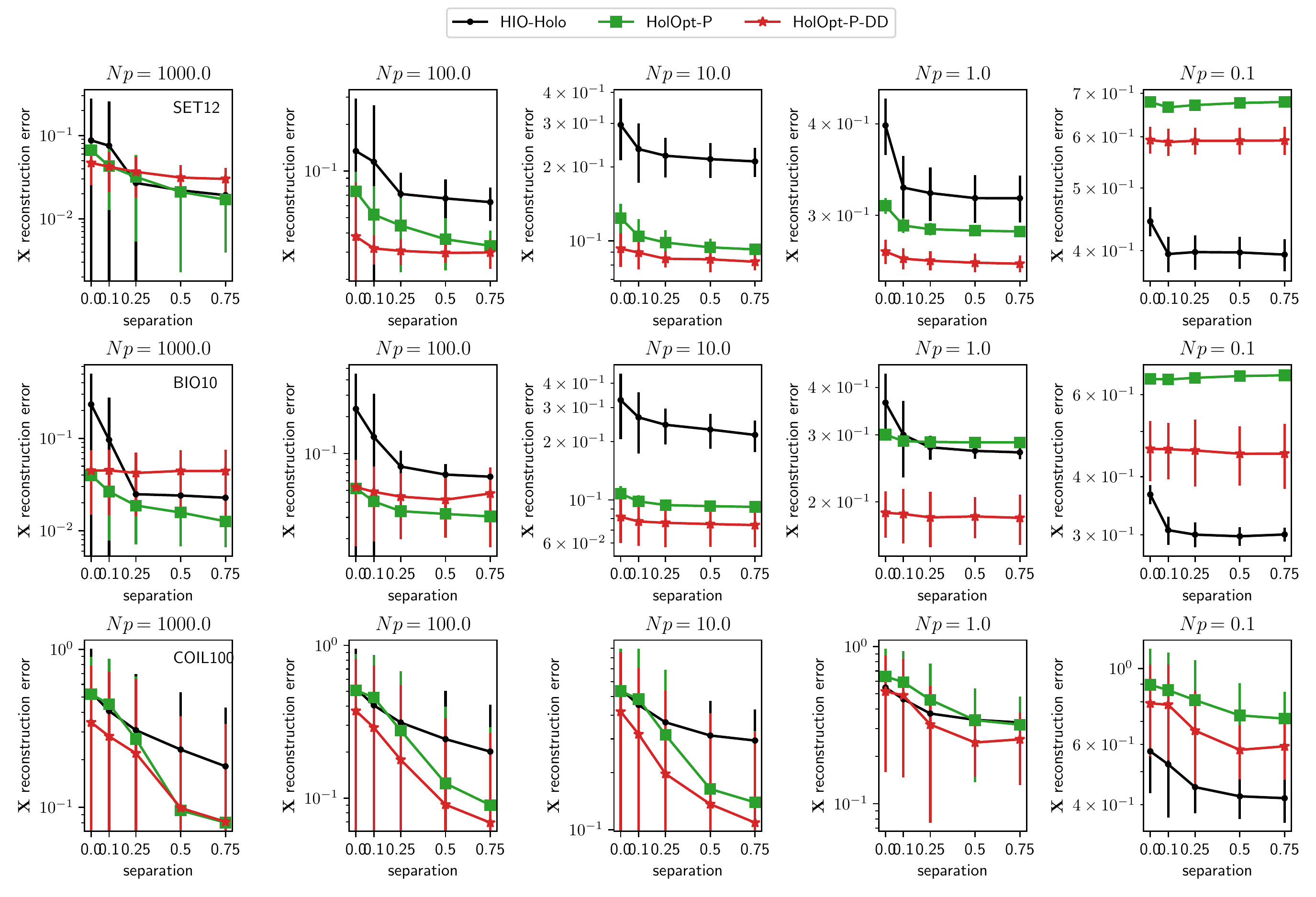}
    \caption{\label{fig:sep-graph} { Reconstruction errors for decreasing photon counts $N_p$ with a $0.1 m \times 0.1 m $ binary random reference as a function of the relative separation. A separation of $0.5$ implies that the left-most non-zero pixel of the reference is $0.5n$ pixels away form the image. Dashed lines corresponds to best run out of the 10 runs in terms of residual error.}}
\end{figure}

\subsection{Robustness to separation experiment}
\label{app:ref-distance}

Figure \ref{fig:sep-visual-np01-np100}, similar to \ref{fig:sep-visual-np1-np10}, displays visual of the reconstruction as a function of the separation between sample and reference for two additional noise levels. 
in Figure \ref{fig:sep-graph} we report MSEs averaged over the datasets as a function of the separation. 
Observations on the ordering of the methods and dependence with separation are comparable with the observations drawn from SSIM plots in the main text.

\subsection{Missing low-frequencies (beamstop) experiment}

\begin{figure} 
    \centering
    \includegraphics[width=\textwidth]{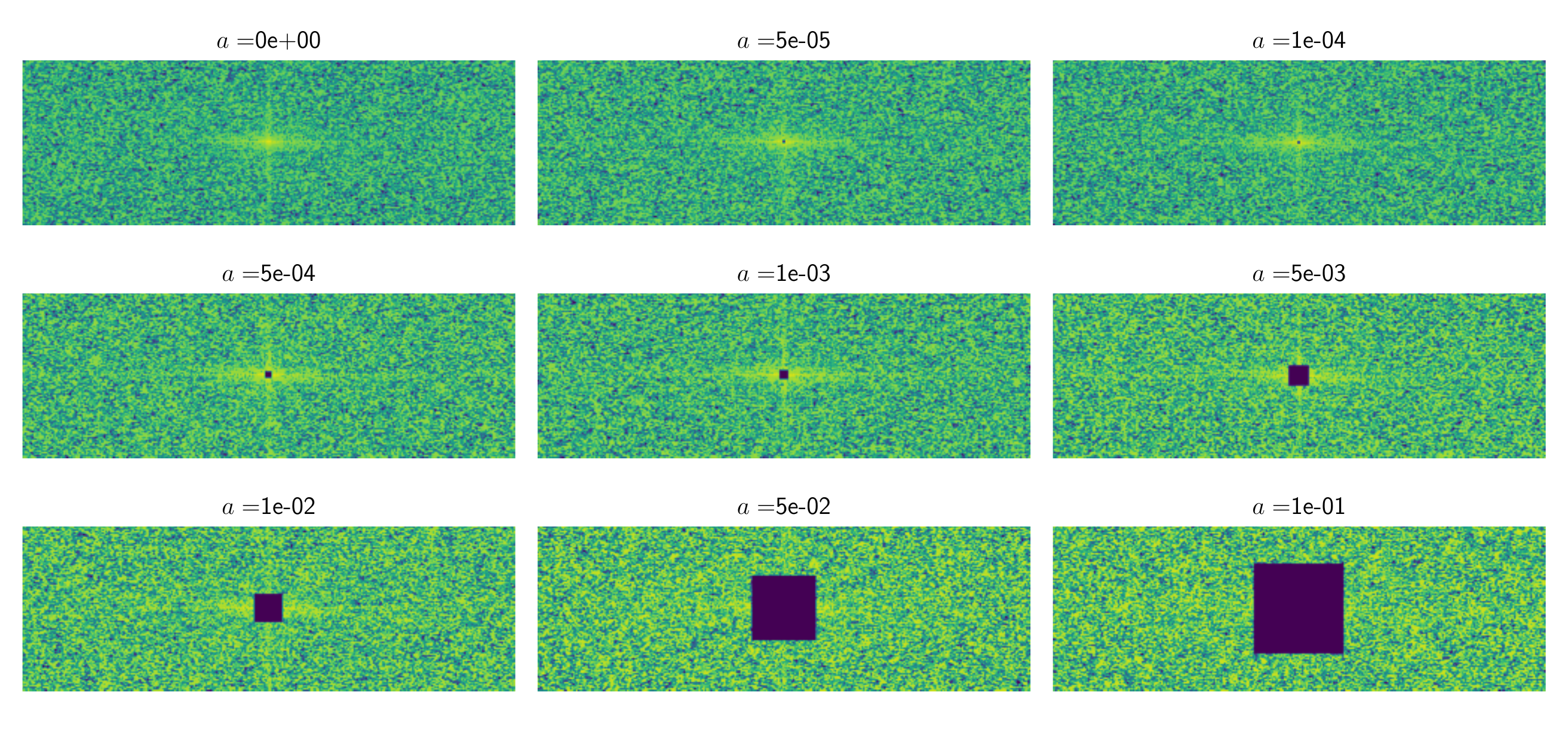}
    \caption{\label{fig:beamstop-area-fractions}Examples of the measured magnitudes corresponding to varying beamstop area fractions for the first image of SET12, CAMERA, with $N_p=10$.}
\end{figure}

\begin{figure}
    \centering
    \includegraphics[width=\textwidth]{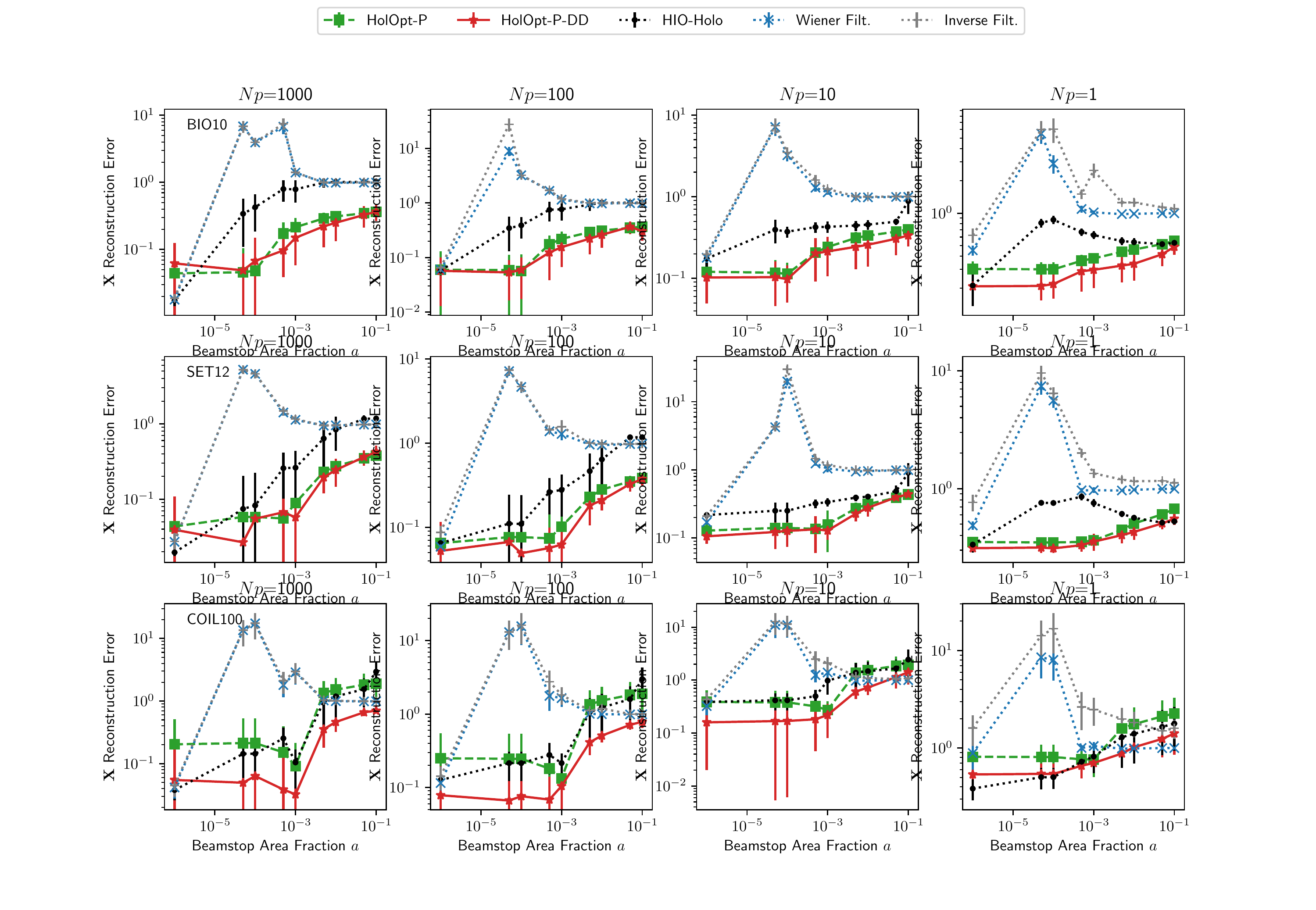}
    \caption{Reconstruction errors on all three datasets (one per row, as indicated by the leftmost row levels) as a function of beamstop area fraction. Errors and error bars are computed as in Figure \ref{fig:denoising-graph-ssim}. The leftmost datapoints at $a=1\mathrm{e
}{-6}$ correspond to no missing data,
i.e. 
$a=0$ at the leftmost points. \label{fig:beamstop-full-SSIM-err}}
\end{figure}

\begin{figure}
    \centering
    \includegraphics[width=\textwidth]{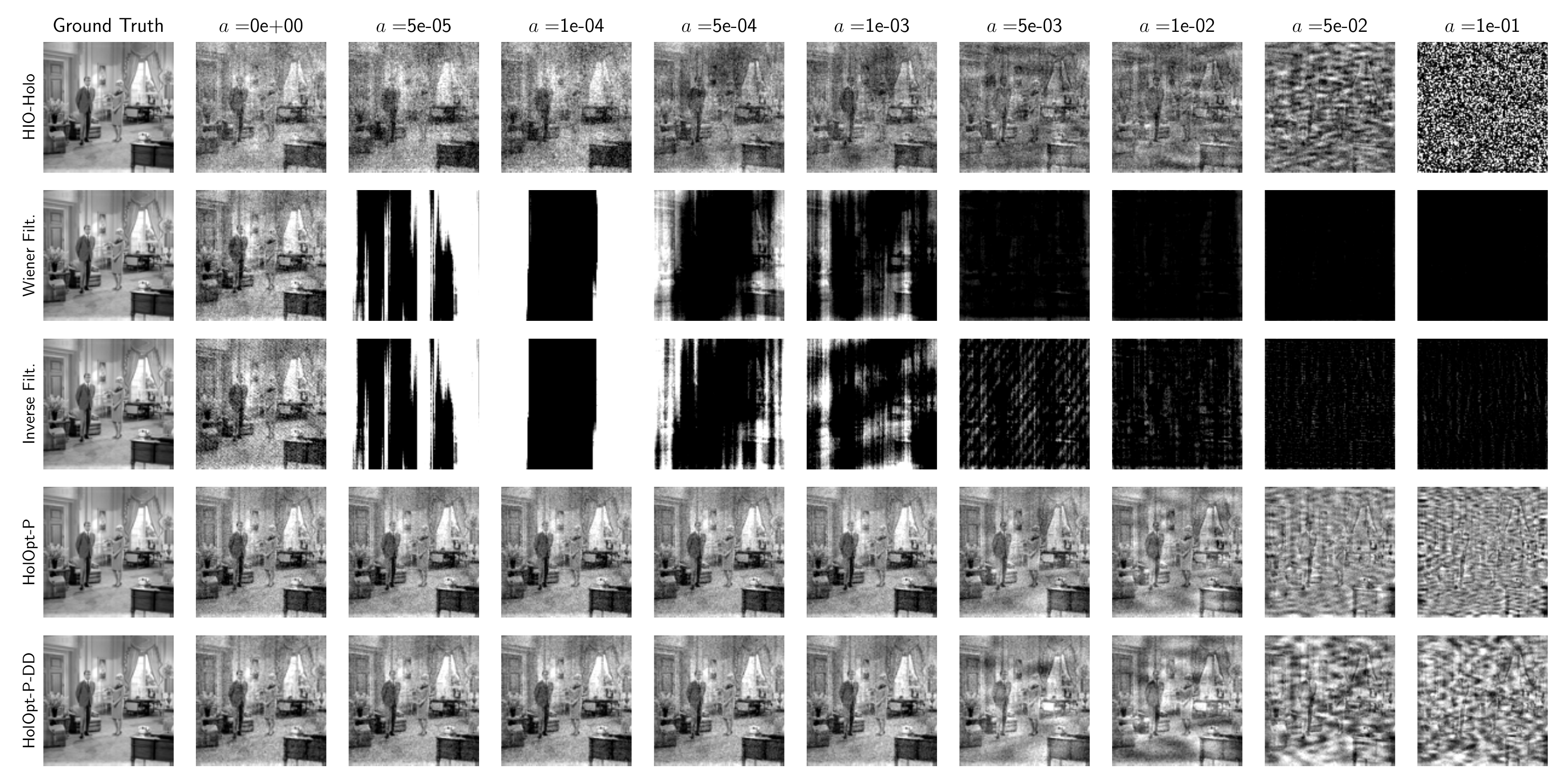}
    \caption{Reconstructed SET12 image with varying beamstop at $N_p=10$ photons.}
    \label{fig:beamstop-camera-10}
\end{figure}

\begin{figure}
    \centering
    \includegraphics[width=\textwidth]{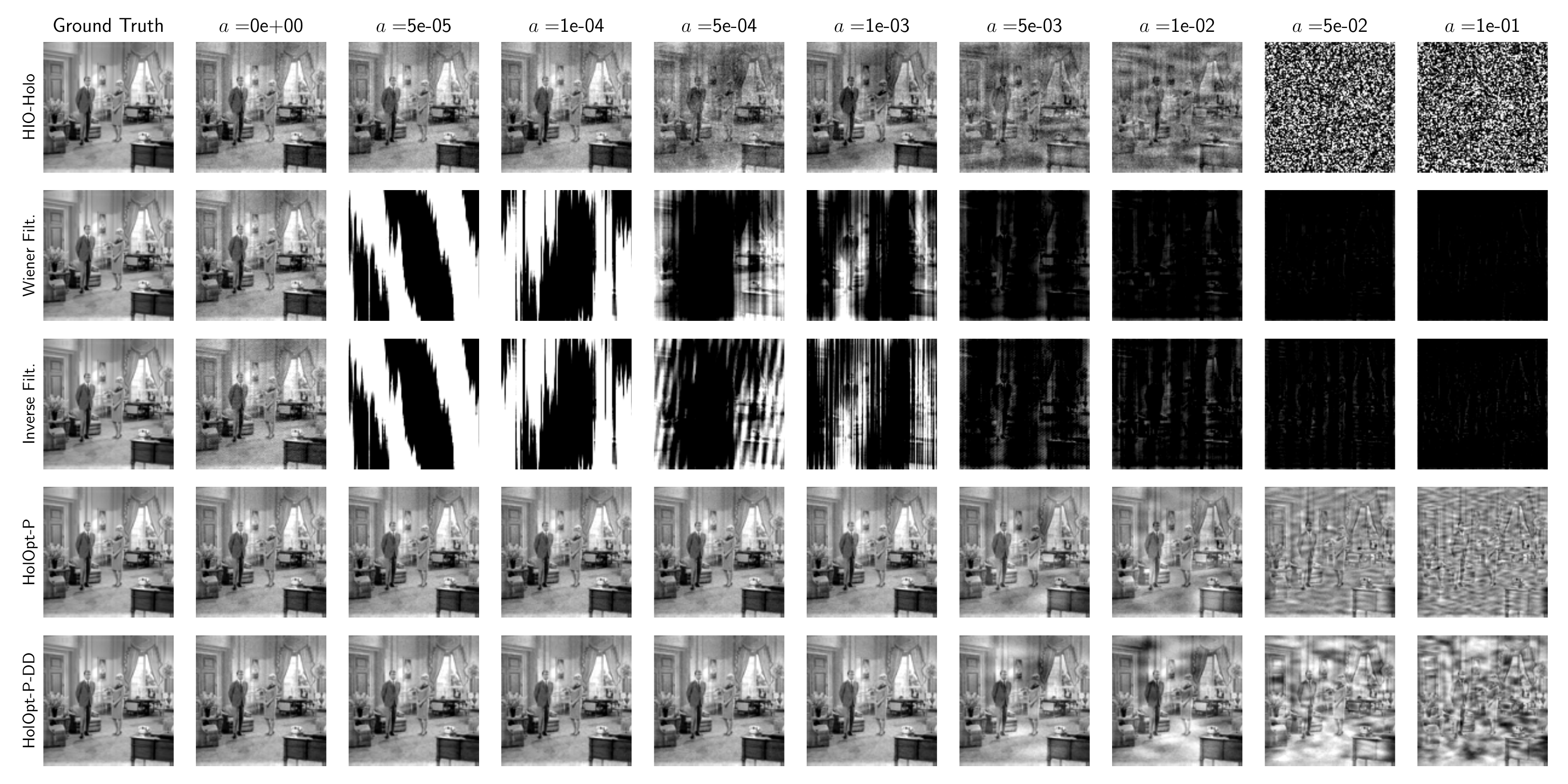}
    \caption{\label{fig:beamstop-CAMERA-1}Reconstructed SET12 image with varying beamstop at $N_p=100$ photons.}
    \label{fig:beamstop-camera-1}
\end{figure}

\begin{figure}
    \centering
    \includegraphics[width=\textwidth]{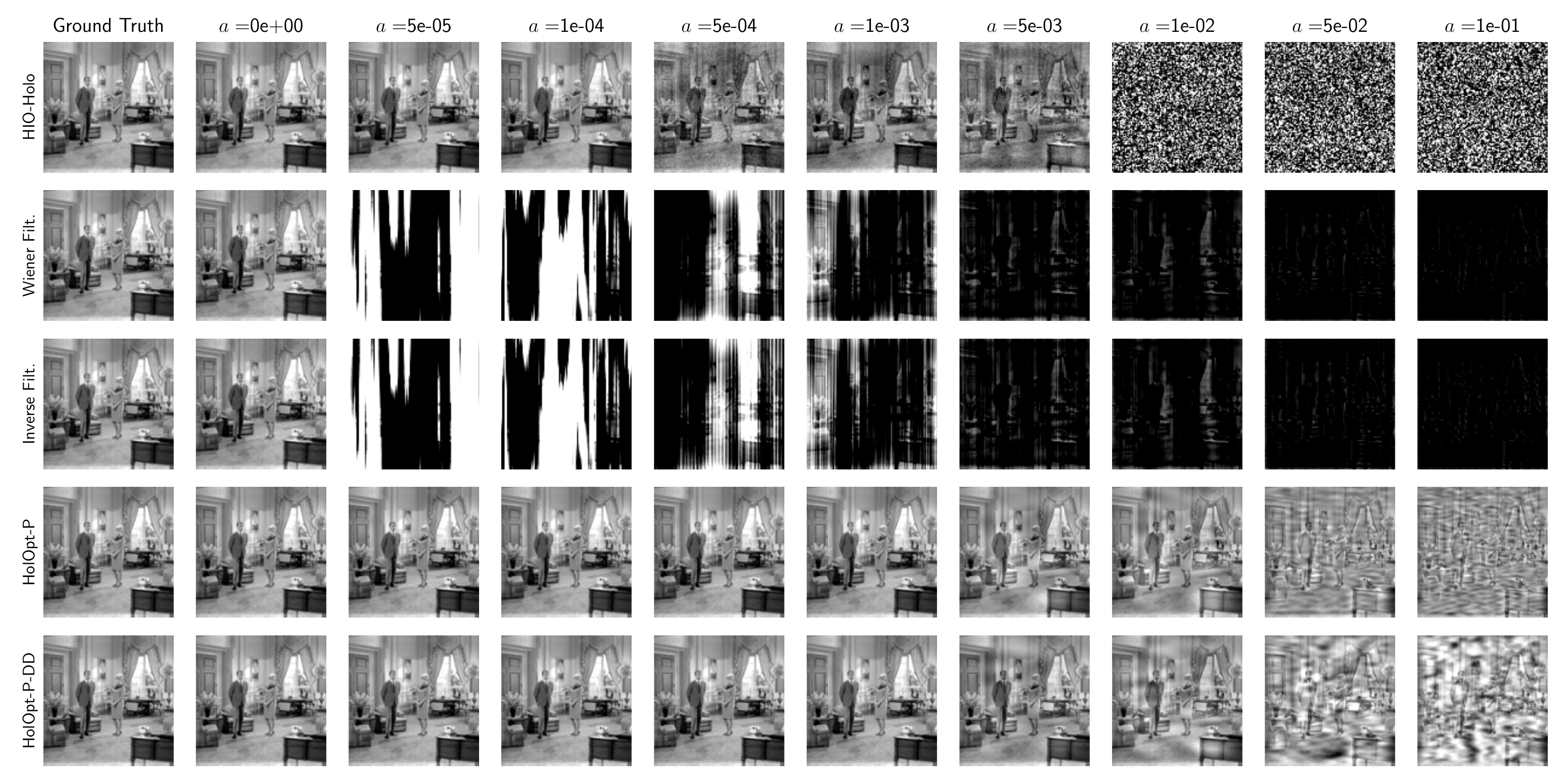}
    \caption{Reconstructed SET12 image with varying beamstop at $N_p=1000$ photons.}
    \label{fig:beamstop-camera-1000}
\end{figure}

\begin{figure}
    \centering
    \includegraphics[width=\textwidth]{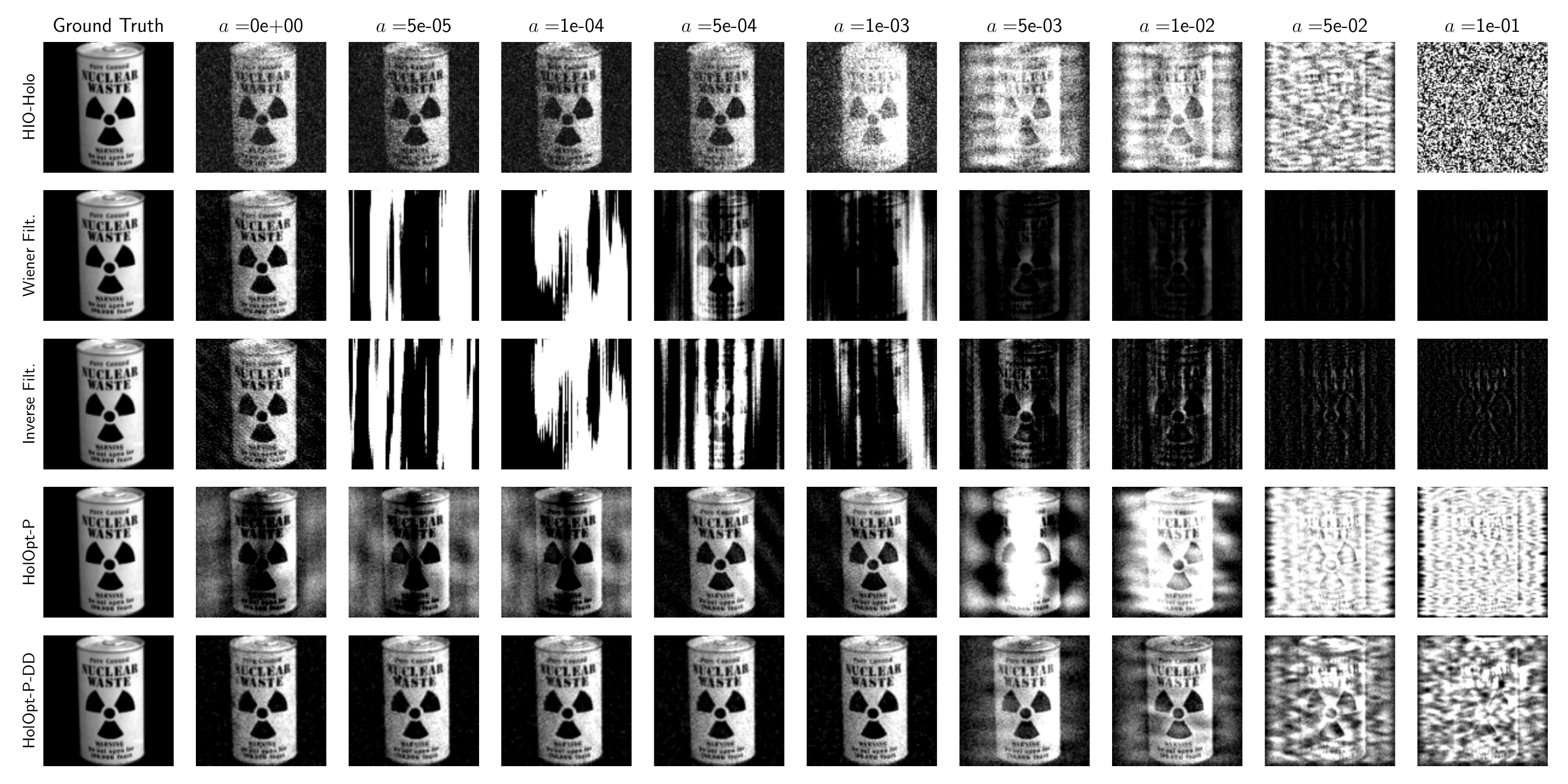}
    \caption{Reconstructed COIL image with varying beamstop at $N_p=10$ photons.}
    \label{fig:beamstop-coil-10}
\end{figure}

\begin{figure}
    \centering
    \includegraphics[width=\textwidth]{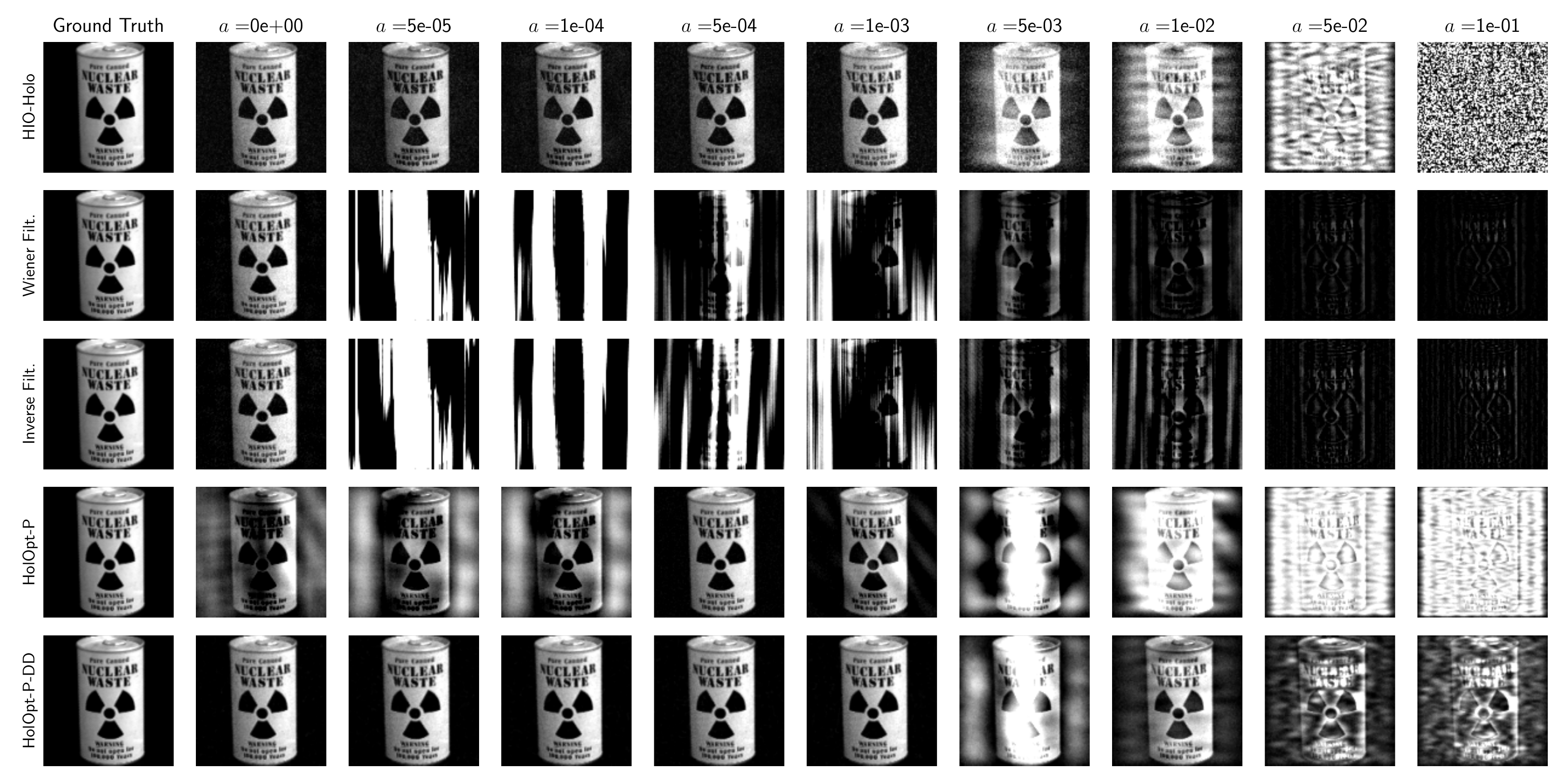}
    \caption{Reconstructed COIL image with varying beamstop at $N_p=100$ photons.}
    \label{fig:beamstop-coil-100}
\end{figure}

\begin{figure}
    \centering
    \includegraphics[width=\textwidth]{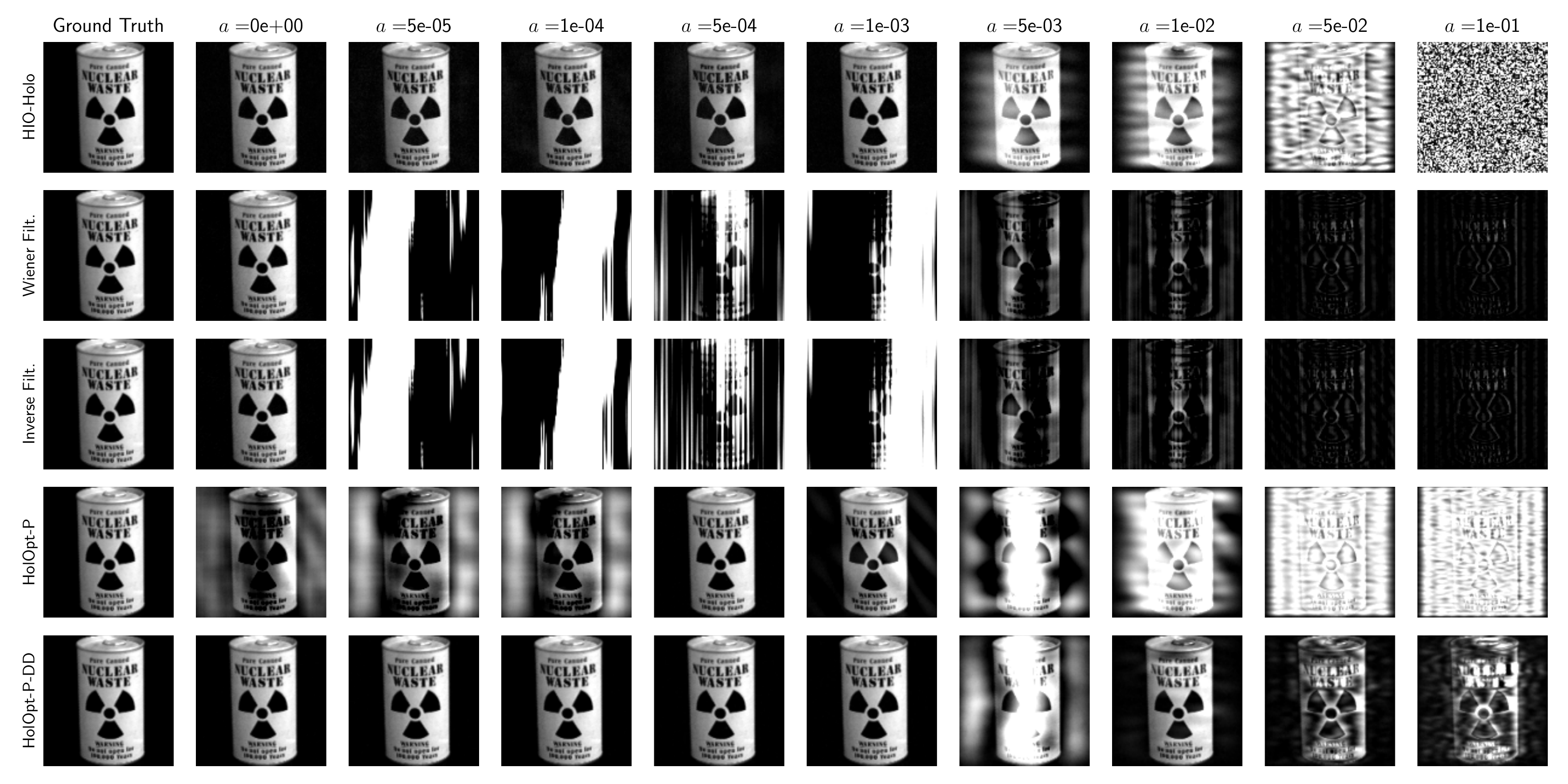}
    \caption{Reconstructed COIL image with varying beamstop at $N_p=1000$ photons.}
    \label{fig:beamstop-coil-1000}
\end{figure}

\begin{figure}
    \centering
    \includegraphics[width=\textwidth]{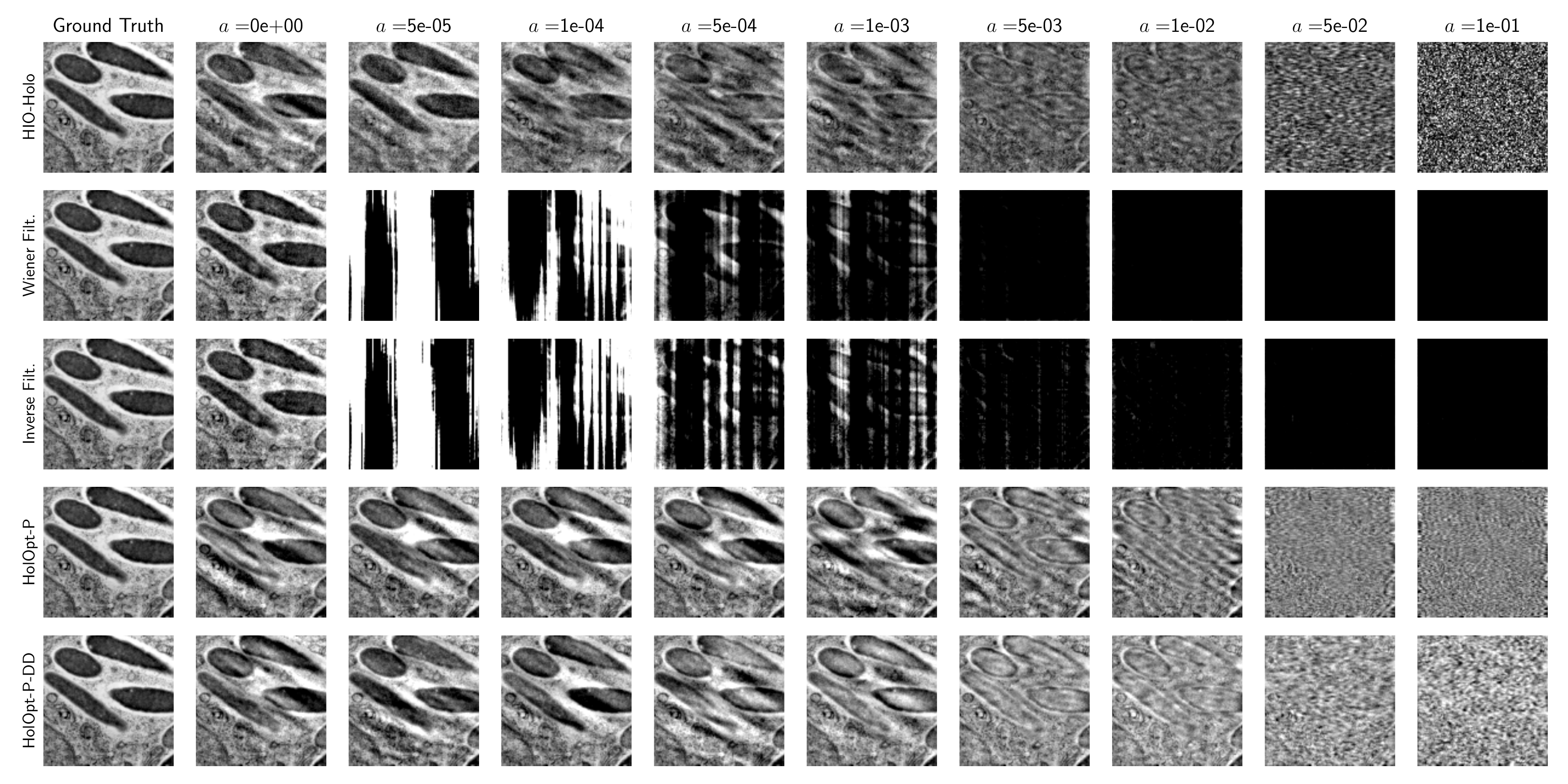}
    \caption{Reconstructed BIO10 image with varying beamstop at $N_p=10$ photons.}
    \label{fig:beamstop-virus-10}
\end{figure}

\begin{figure}
    \centering
    \includegraphics[width=\textwidth]{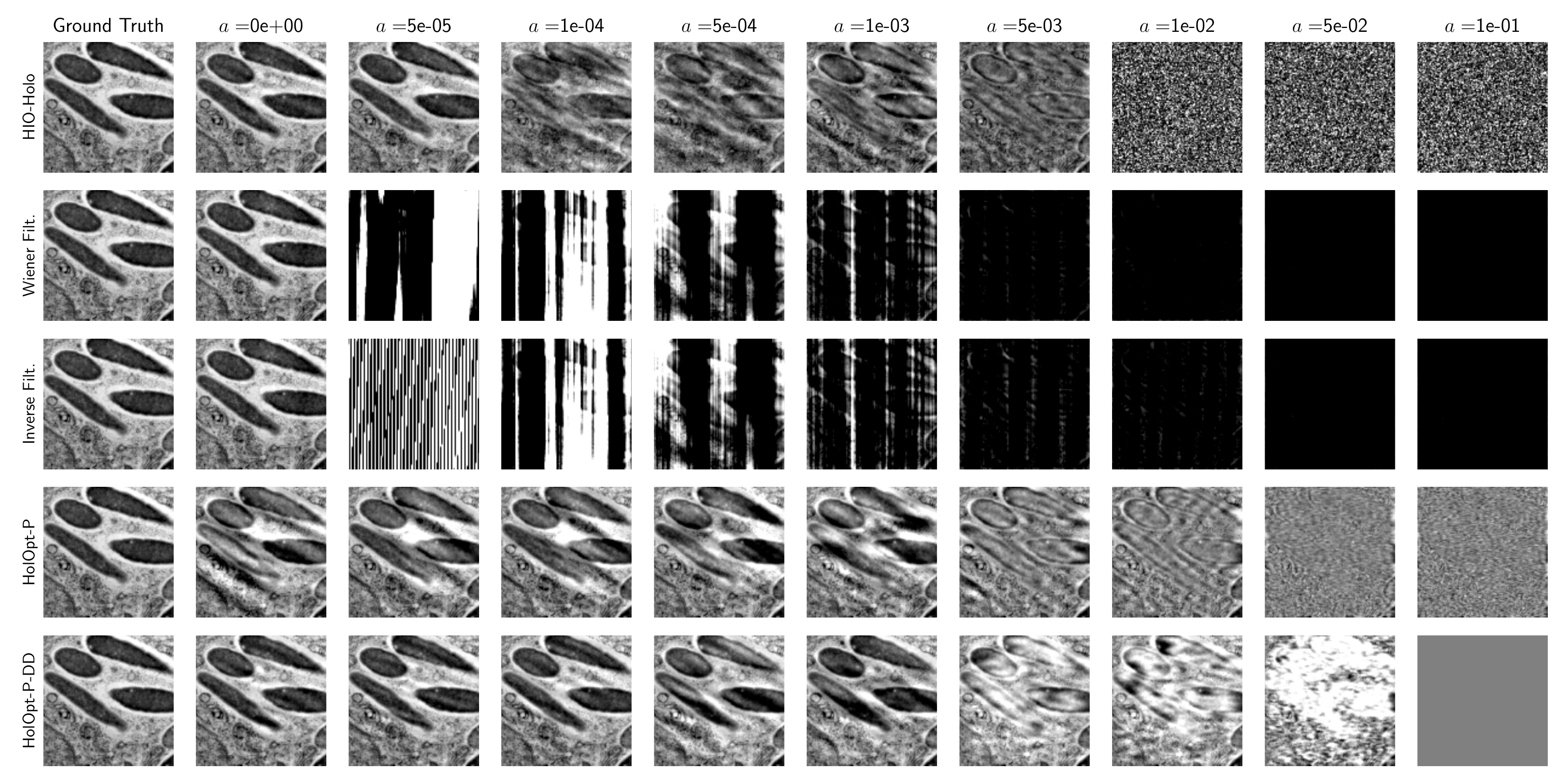}
    \caption{Reconstructed BIO10 image with varying beamstop at $N_p=100$ photons.}
    \label{fig:beamstop-virus-100}
\end{figure}

\begin{figure}
    \centering
    \includegraphics[width=\textwidth]{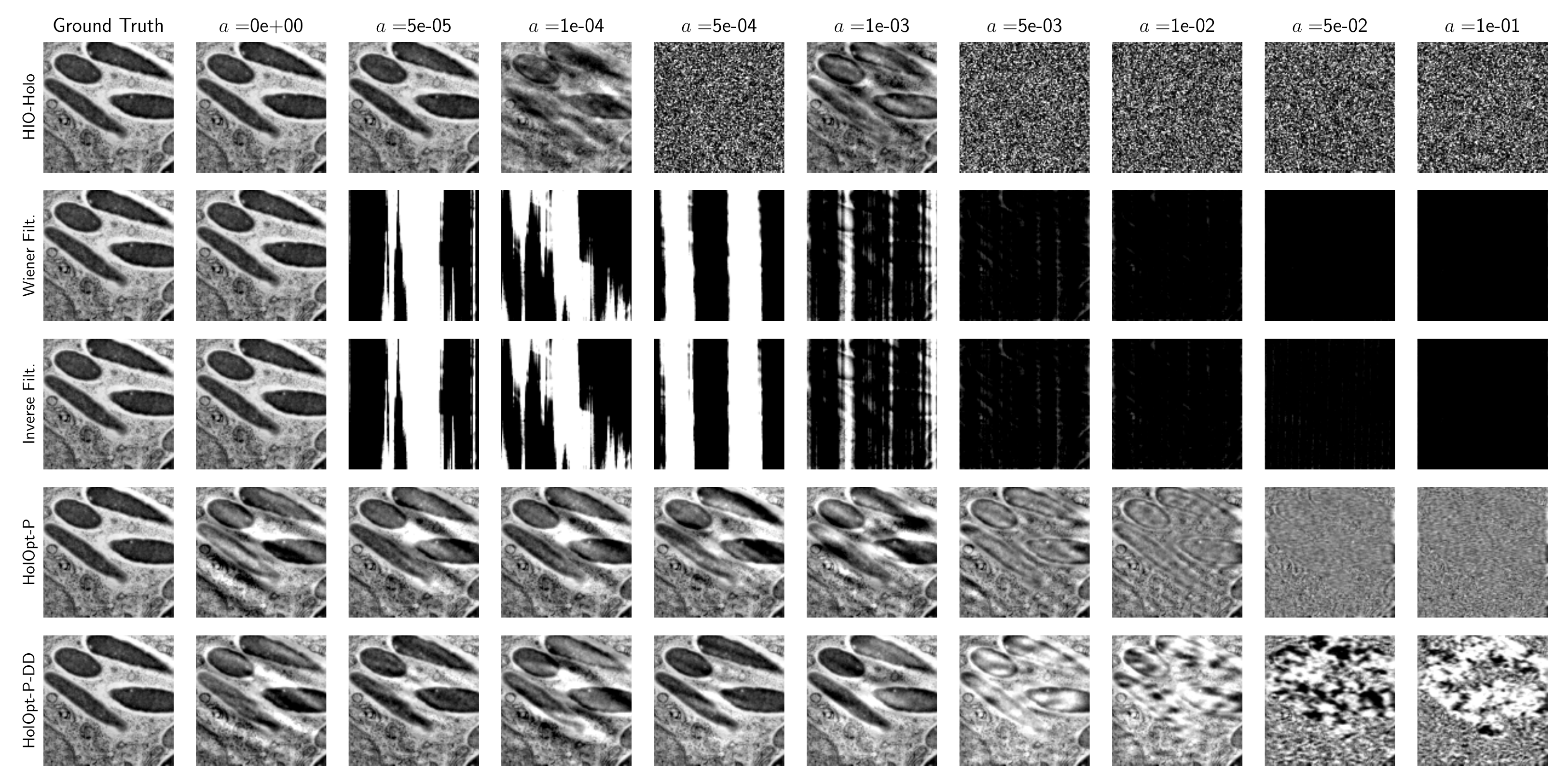}
    \caption{Reconstructed BIO10 image with varying beamstop at $N_p=1000$ photons.}
    \label{fig:beamstop-virus-1000}
\end{figure}

We now visualize the beamstop area fractions tested in Section \ref{sec:beamstop} (see Figure \ref{fig:beamstop-area-fractions}), as well as show analogous plots to Figures \ref{fig:beamstop-SET12-images} for all three test images and for all remaining noise levels, $N_p=10, 100, 1000$ (Figures\ref{fig:beamstop-camera-10}-\ref{fig:beamstop-virus-1000}). Figure \ref{fig:beamstop-full-SSIM-err} also plots all settings using $\ell-2$ errors.

\begin{figure} 
        \centering
        \includegraphics[trim= 0 0 0 0 , clip, width=\textwidth]{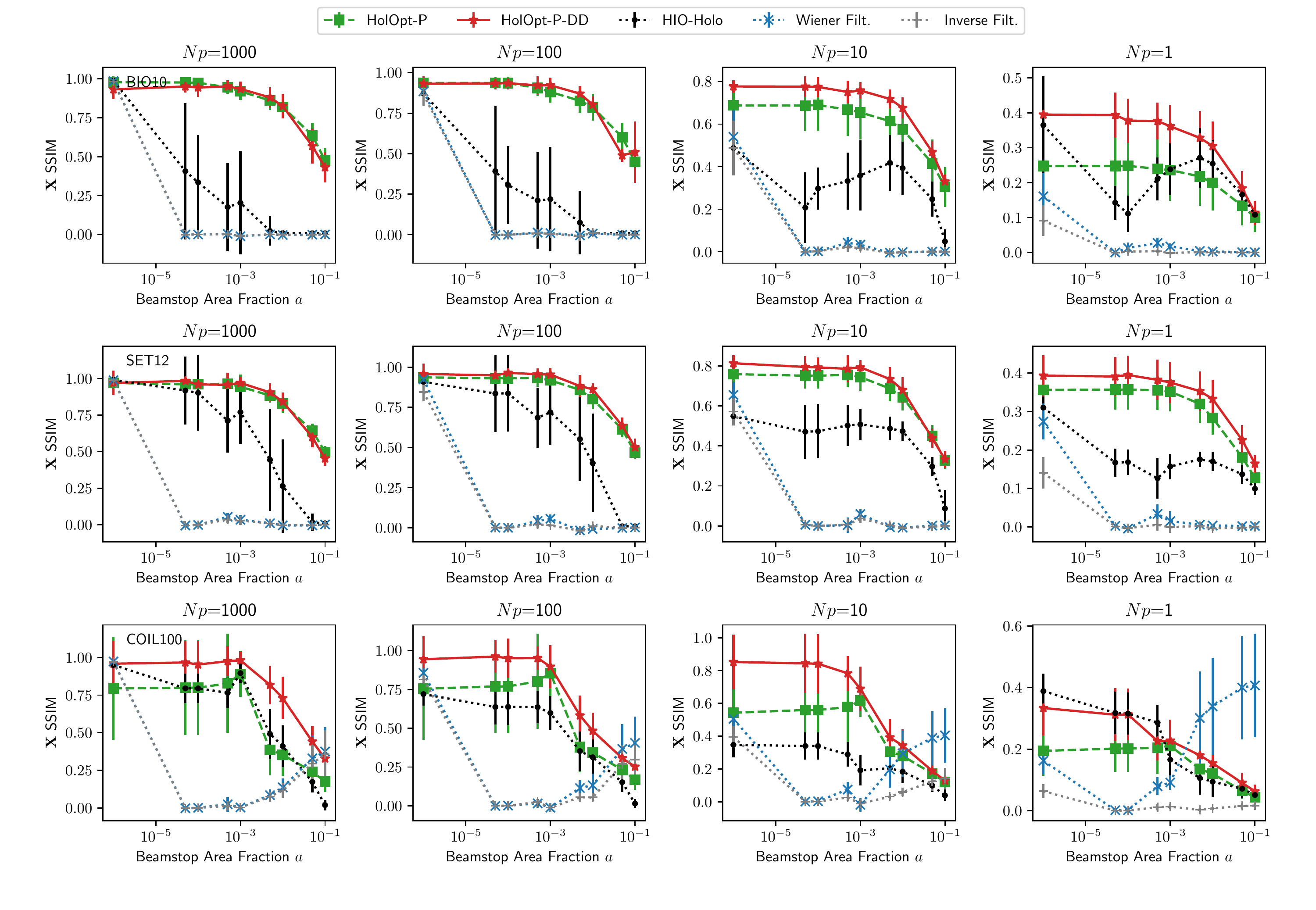}
        \caption{\label{fig:beamstop-error-plots-all} {Reconstruction SSIM as a function of beamstop area fraction. Baseline methods are run for 5 trials per image 
        across all datasets, 
        while for tractability our methods are run 
        for 2 trials each on BIO10 and SET12, and 
        1 trial each on COIL100. Average SSIM and one standard deviation error bars are shown. 
        The leftmost datapoints
        correspond to no missing data,
         i.e. formally $a=0$ at the leftmost points.}}
\end{figure}

\begin{figure} 
    \centering
    \includegraphics[width=0.8\textwidth]{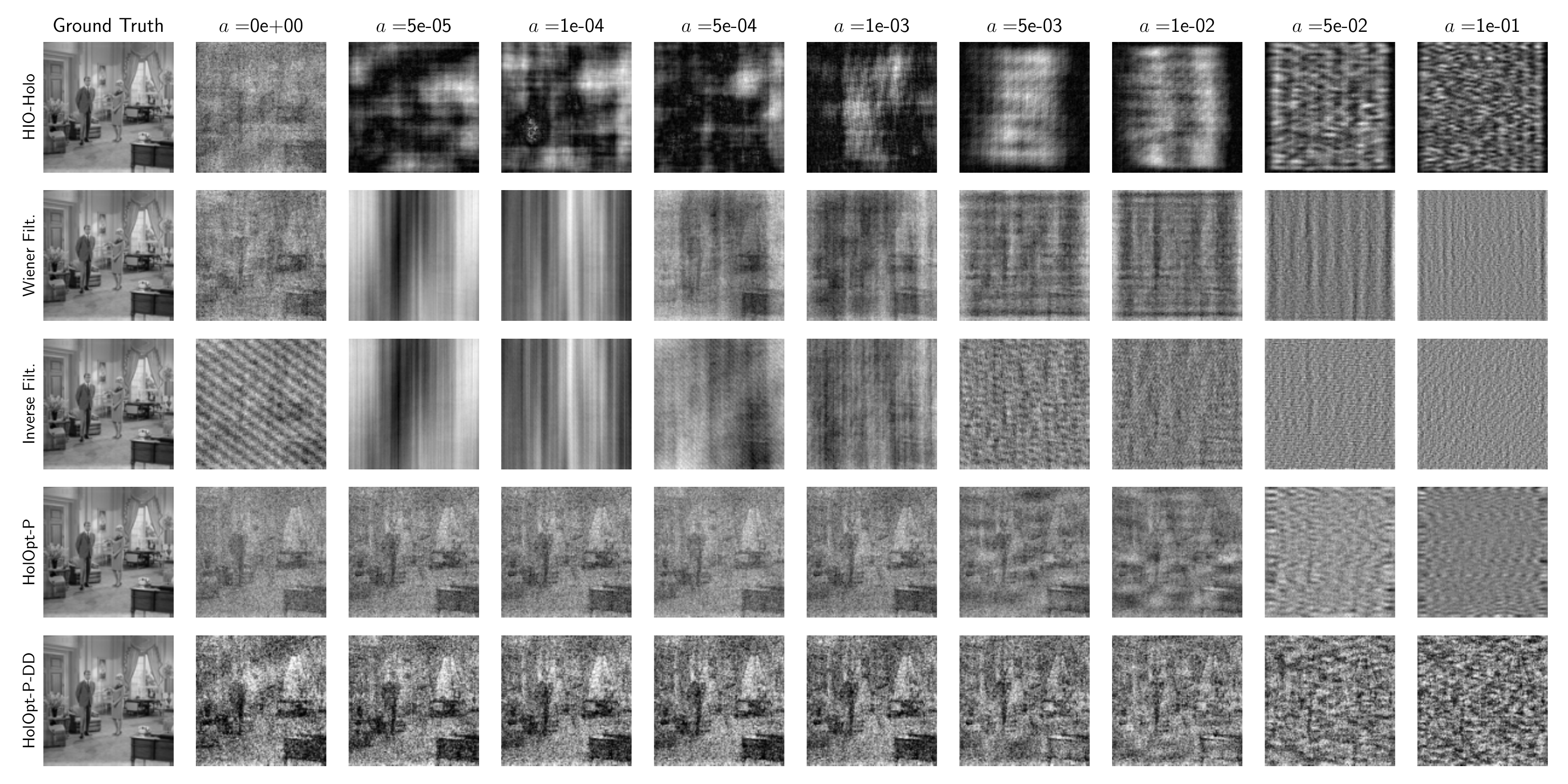}
    \includegraphics[width=0.8\textwidth]{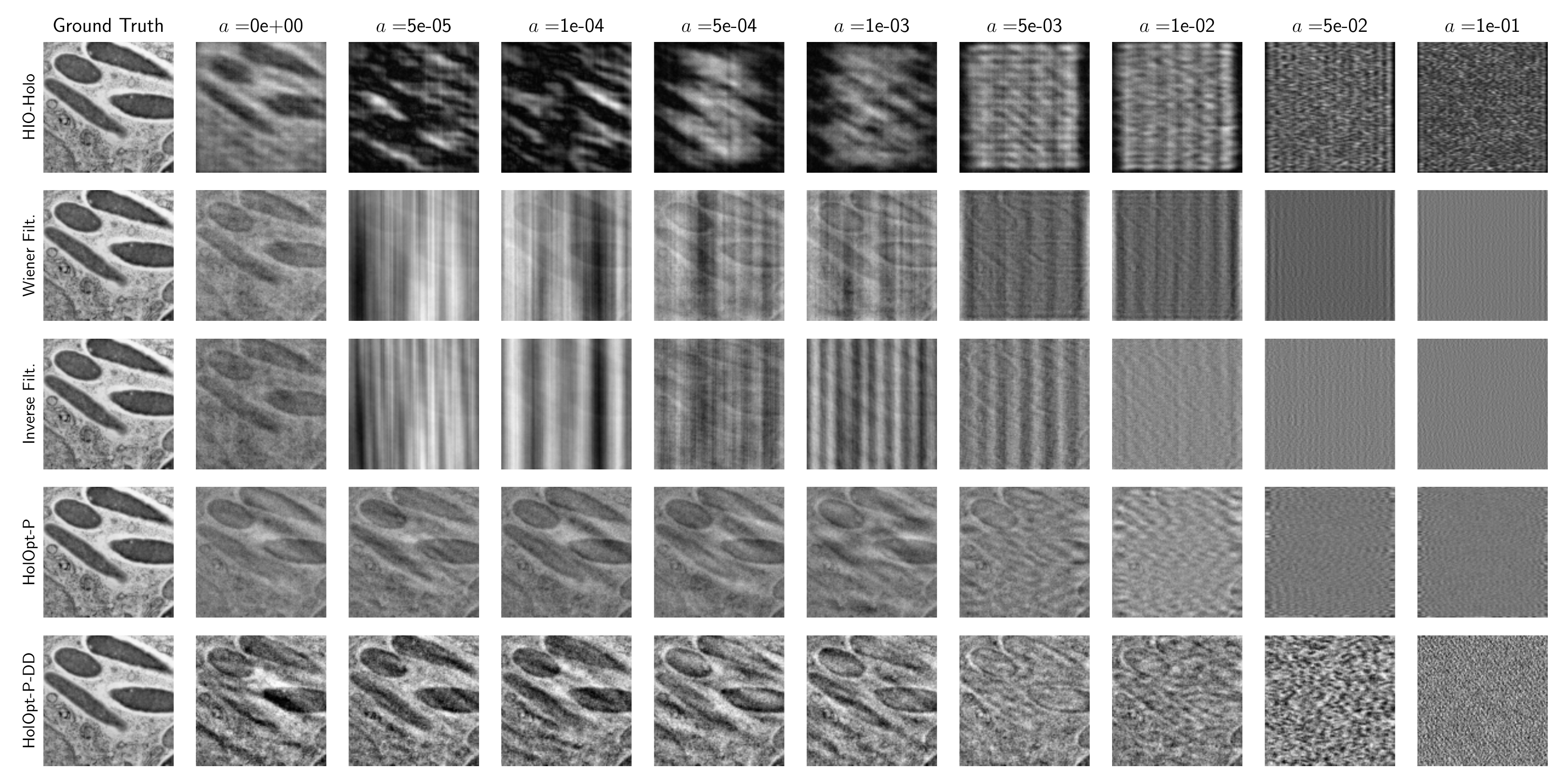}
    \includegraphics[width=0.8\textwidth]{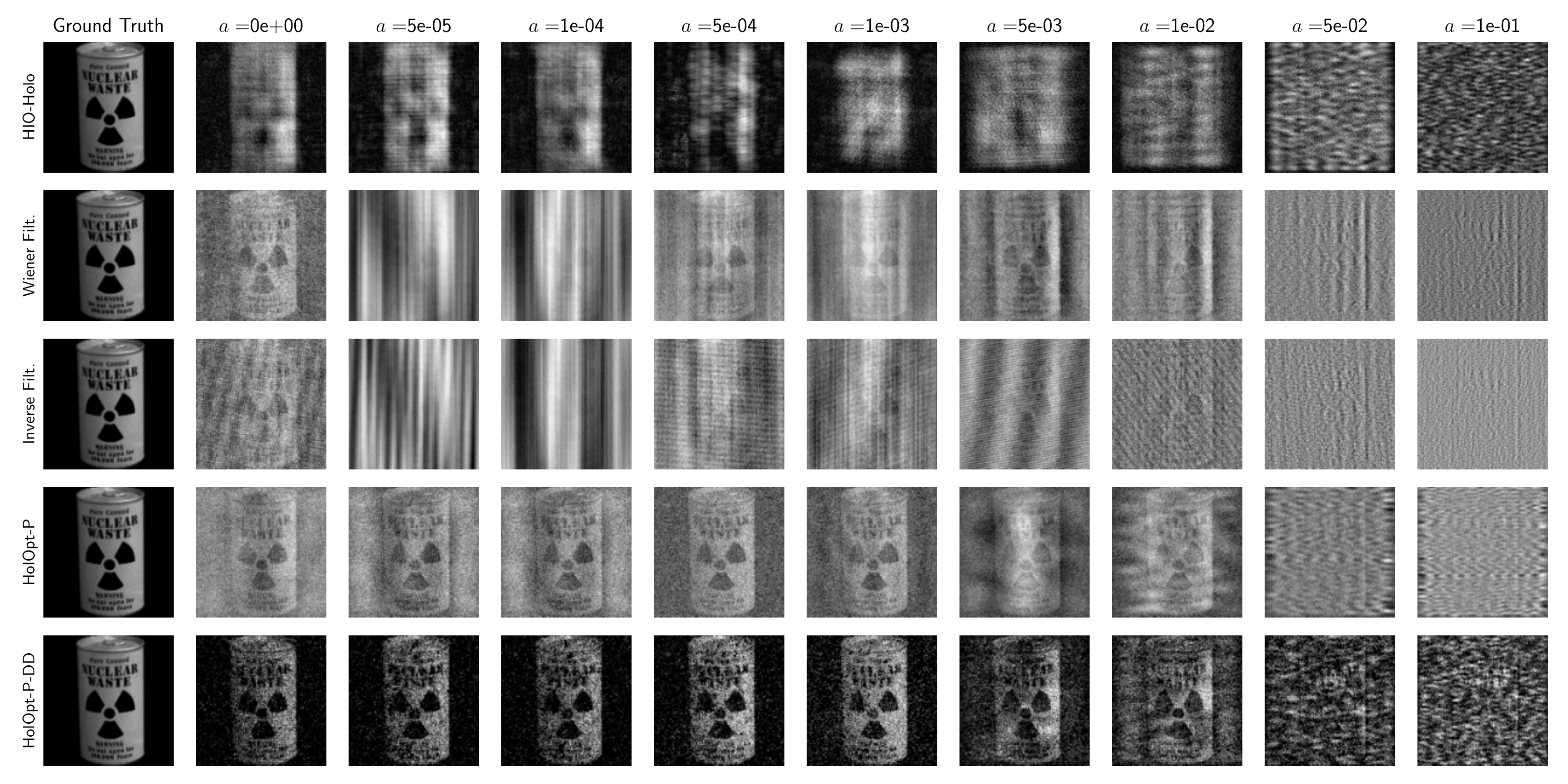}
    \caption{\label{fig:beamstop-SET12-images-orig} { Figure \ref{fig:beamstop-SET12-images} scaled without quantiles. An example reconstructed image from each of the SET12, BIO10, and COIL datasets, respectively, as a function of beamstop area fraction $a$ for fixed photon count $N_p = 1$. 
    }}
\end{figure}

In Figure \ref{fig:beamstop-SET12-images-orig}, for completeness we replicate exactly the reconstructions of Figure \ref{fig:beamstop-SET12-images} with an alternate image scaling technique. In particular, we map colors to pixel values according the absolute minimum and maximum value per reconstruction, instead of the more nuanced quantile approach of the main body (done to maintain fairness in intensity comparisons between methods). Hints of the underlying images are newly visible in the worst reconstructions, particularly in Wiener and inverse filtering, but the relative quality of the competing methods is wholly unchanged.

\begin{figure}[!htbp] 
    \centering
    \includegraphics[width=0.8\textwidth]{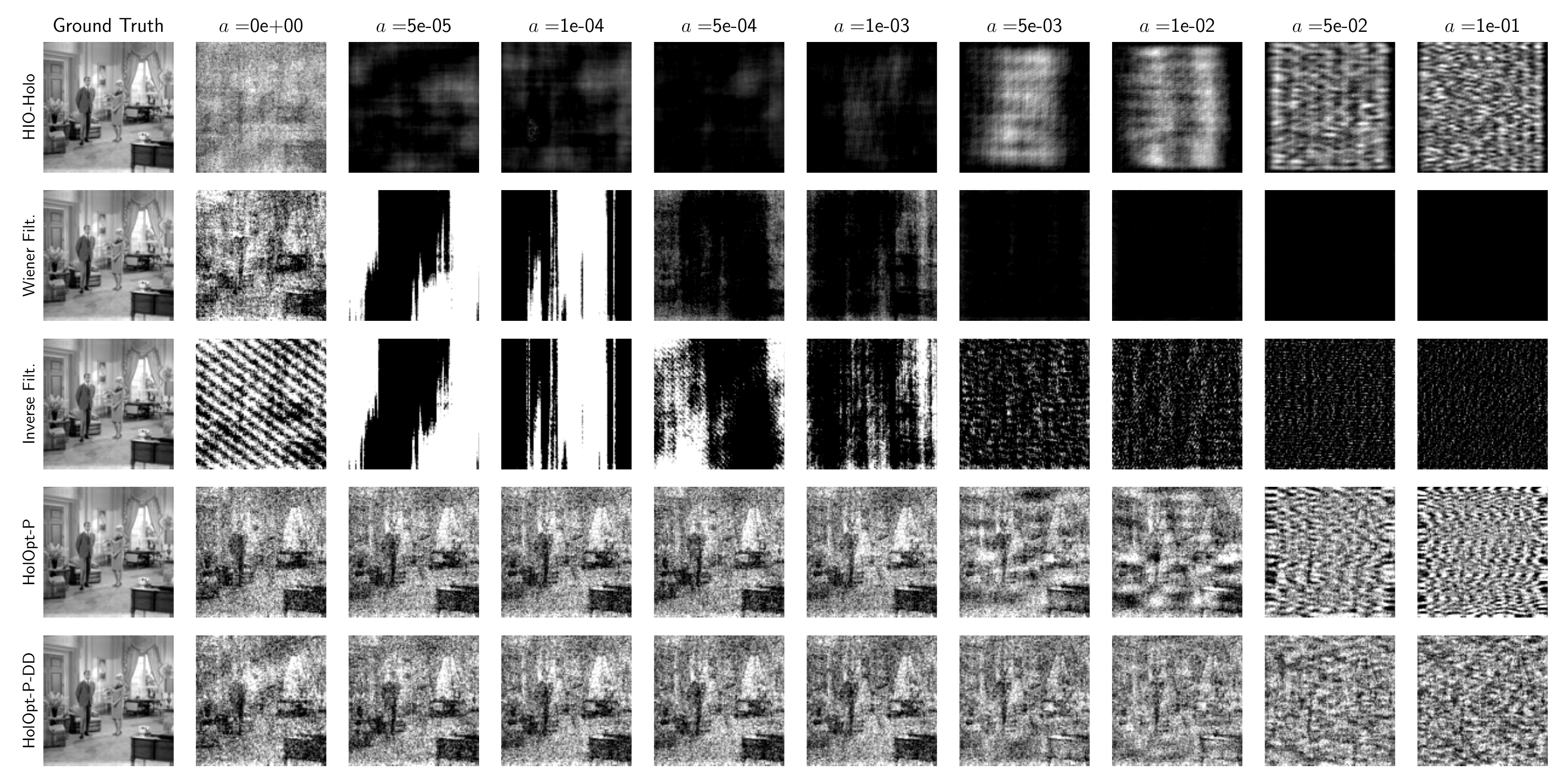}
    \includegraphics[width=0.8\textwidth]{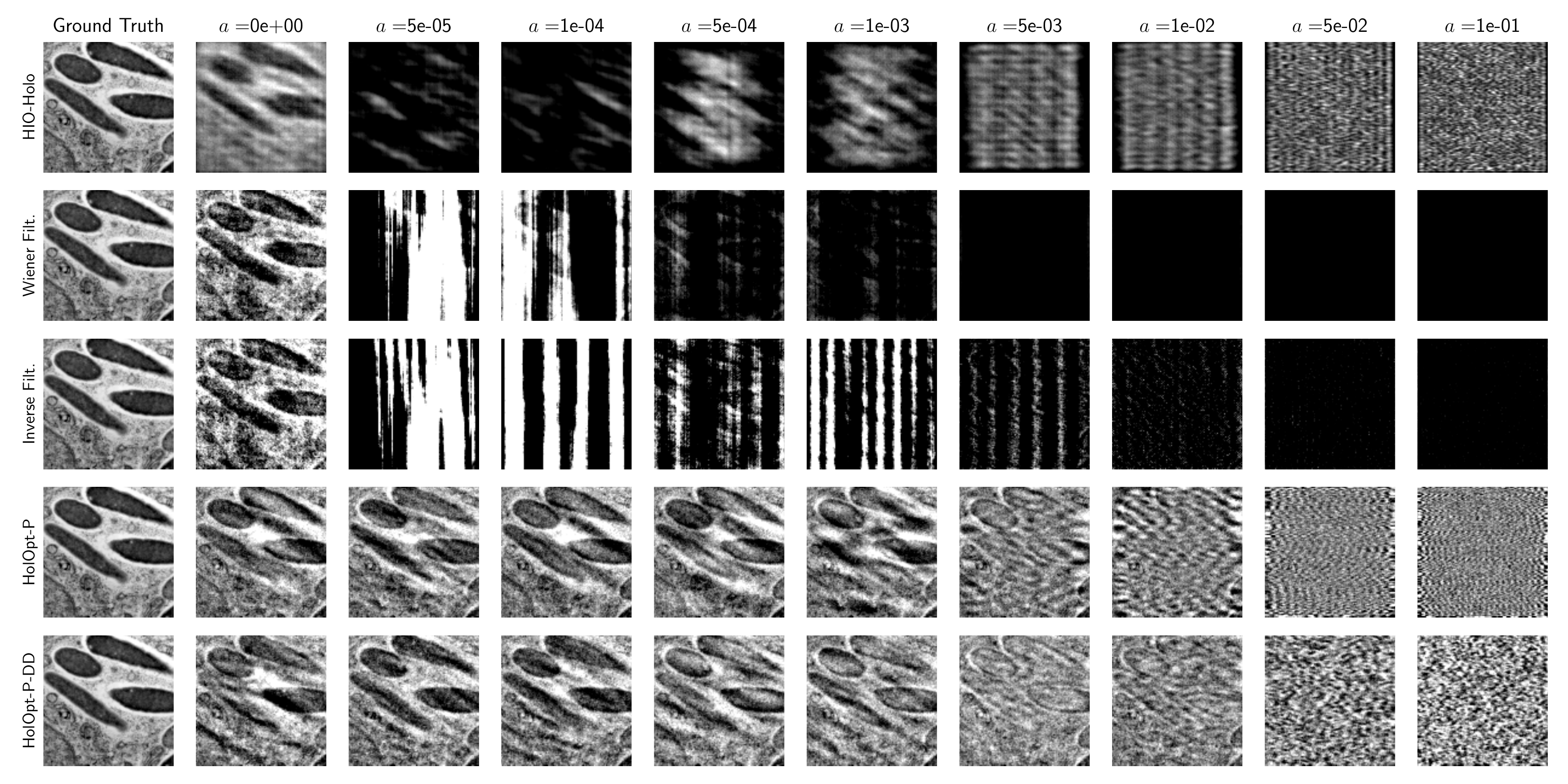}
    \includegraphics[width=0.8\textwidth]{figures/msml_scale_beamstop/COIL_Im_Photons_1.pdf}
    \caption{\label{fig:beamstop-SET12-images} { An example reconstructed image from 
    each of the SET12, BIO10, and
    COIL datasets, respectively, 
    as a function of beamstop area fraction $a$ for fixed photon count $N_p = 1$. See also Figure \ref{fig:beamstop-SET12-images-orig} for a differently scaled visualization.
    }}
\end{figure}

\begin{figure}
    \centering
    \includegraphics[width=\textwidth]{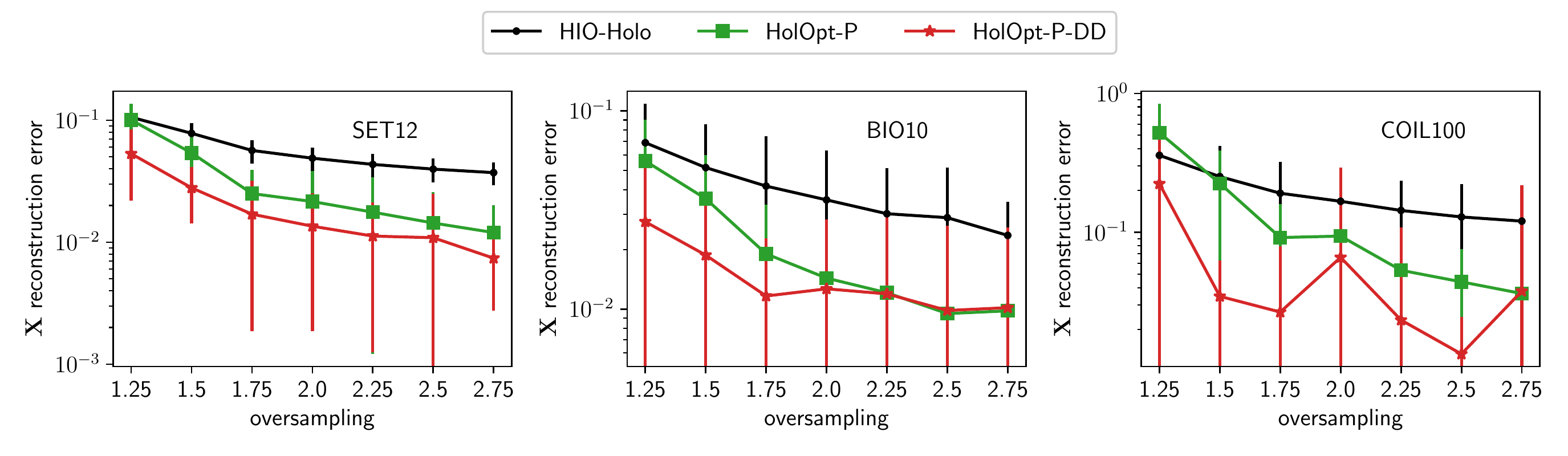}
    \caption{Reconstruction $\ell2$-errors for SET12, BIO10 and COIL100 with varying oversampling factor at $N_p=10$ photon/pixel. Corresponding SSIM and visuals are displayed in Figures \ref{fig:ovs-graph} and \ref{fig:ovs-visuals}.}
    \label{fig:ovs-graph-mse}
\end{figure}

\begin{figure}[!htbp] 
    \centering
    {\includegraphics[width=0.49\textwidth]{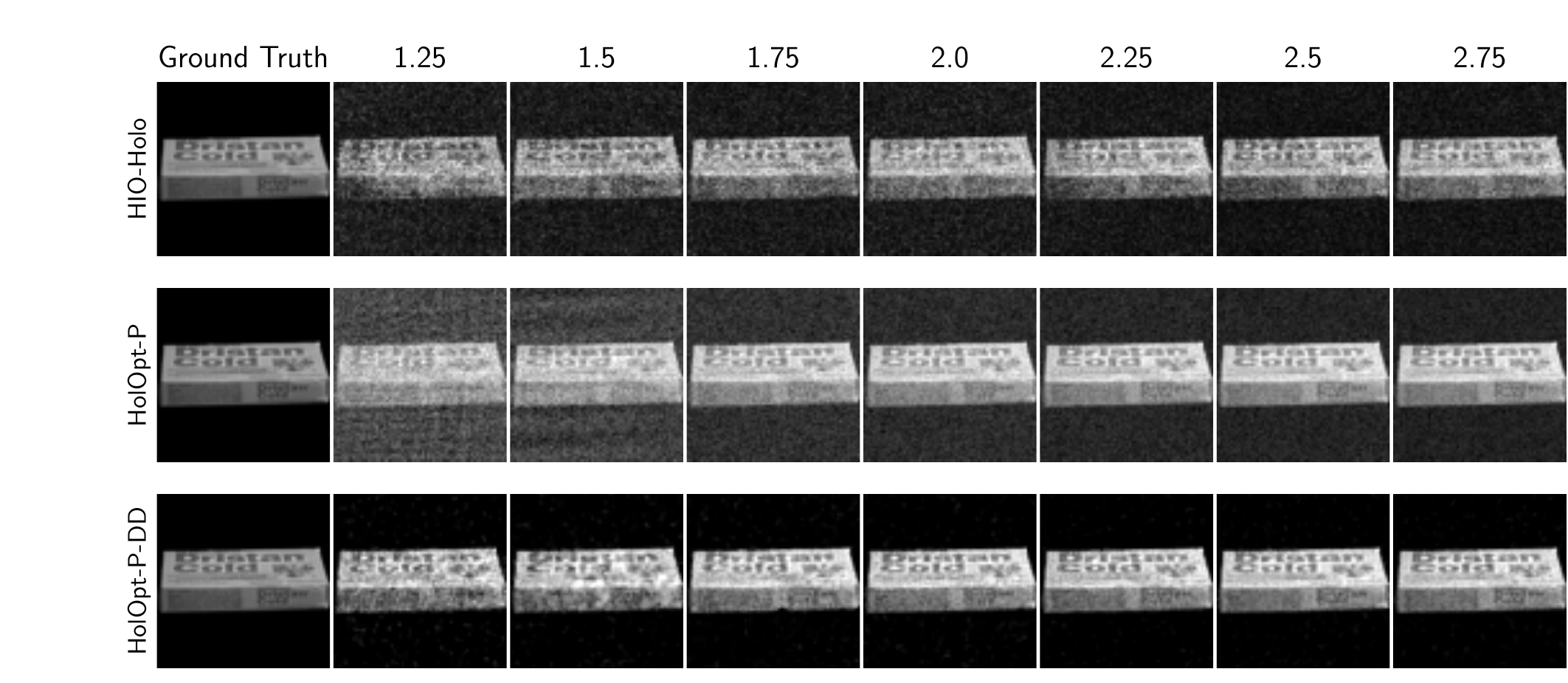}}
    {\includegraphics[width=0.49\textwidth]{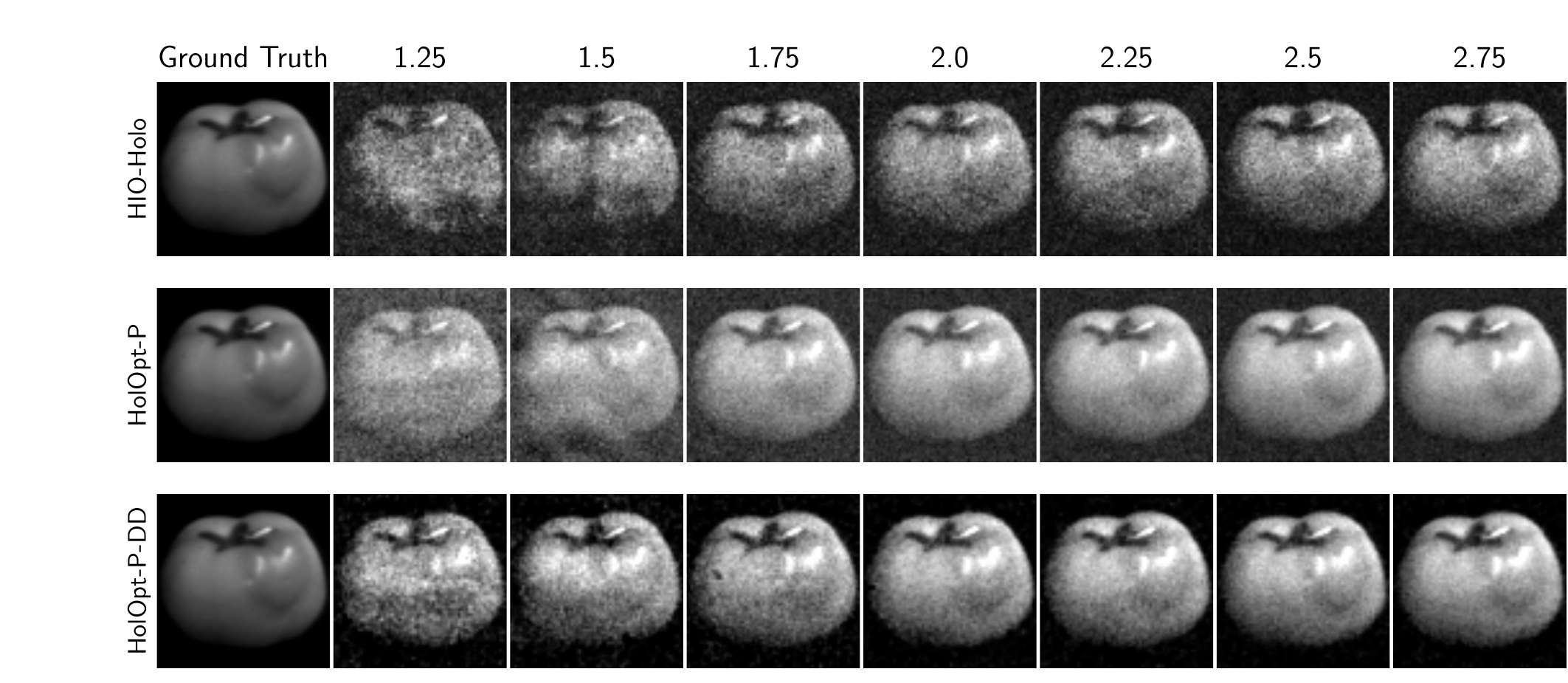}}
    \setlength{\belowcaptionskip}{-20pt}
    \caption{Reconstructed images for samples from COIL100 
     with varying oversampling factors (numbers above each column) at $N_p=10$ photon/pixel. Same as Figure \ref{fig:ovs-visuals}.}
    \label{fig:ovs-visuals-COIL}
\end{figure}

\begin{figure}
    \centering
    {\includegraphics[width=0.49\textwidth]{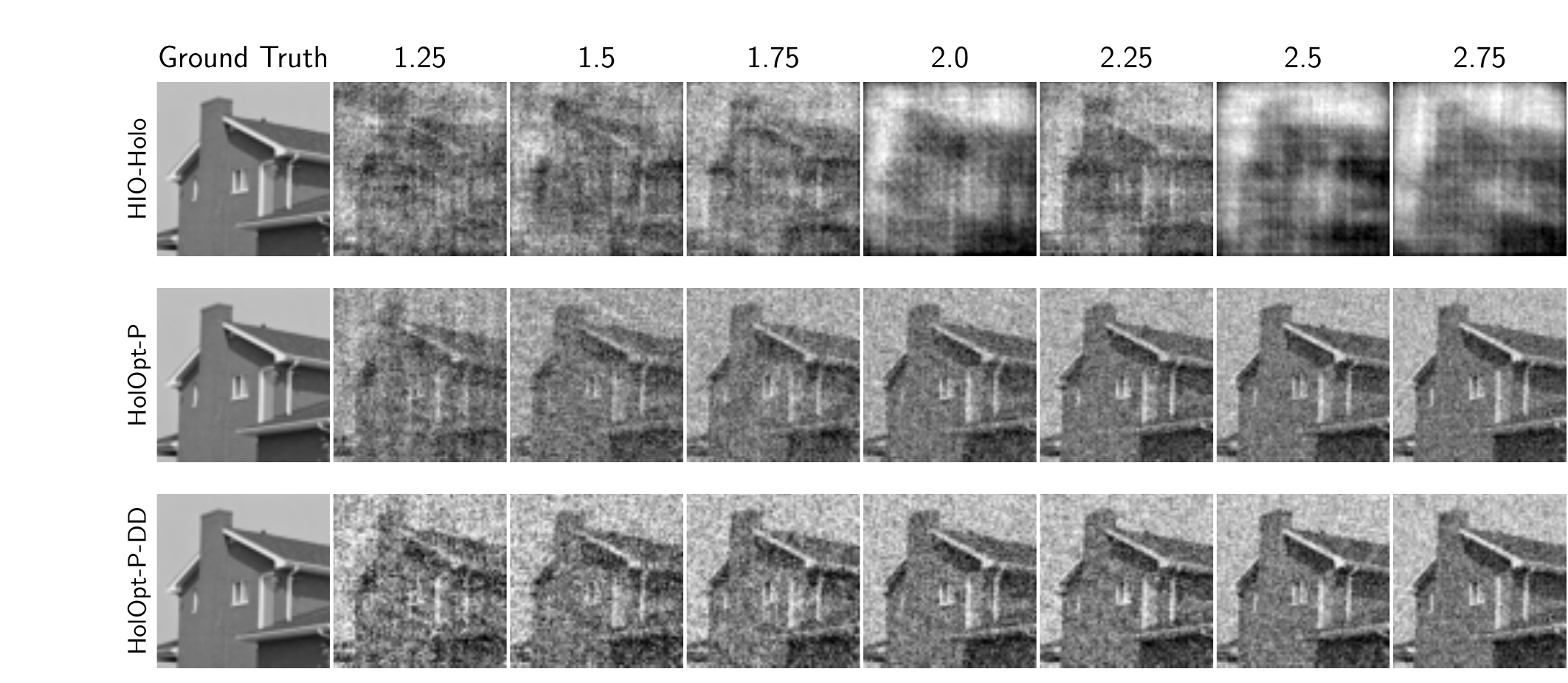}}
    {\includegraphics[width=0.49\textwidth]{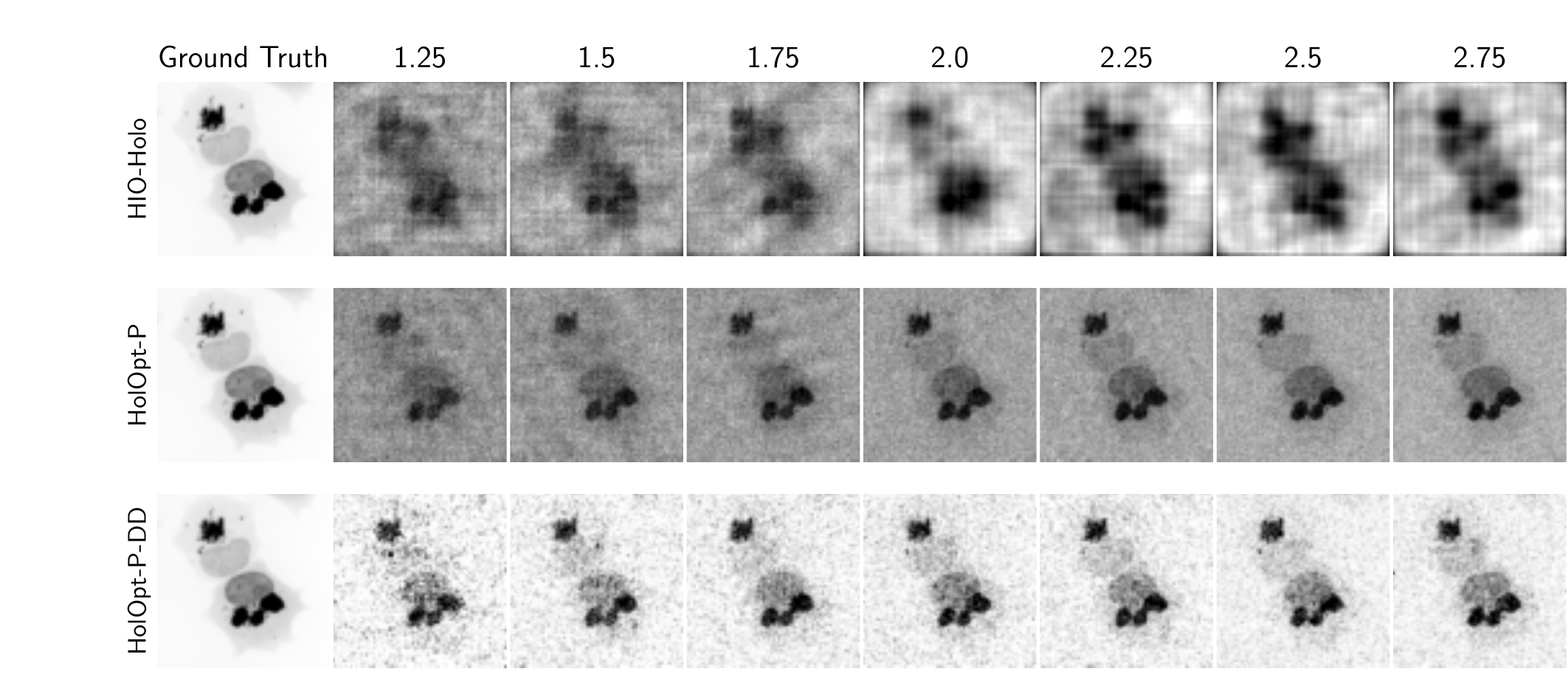}}\\
    {\includegraphics[width=0.49\textwidth]{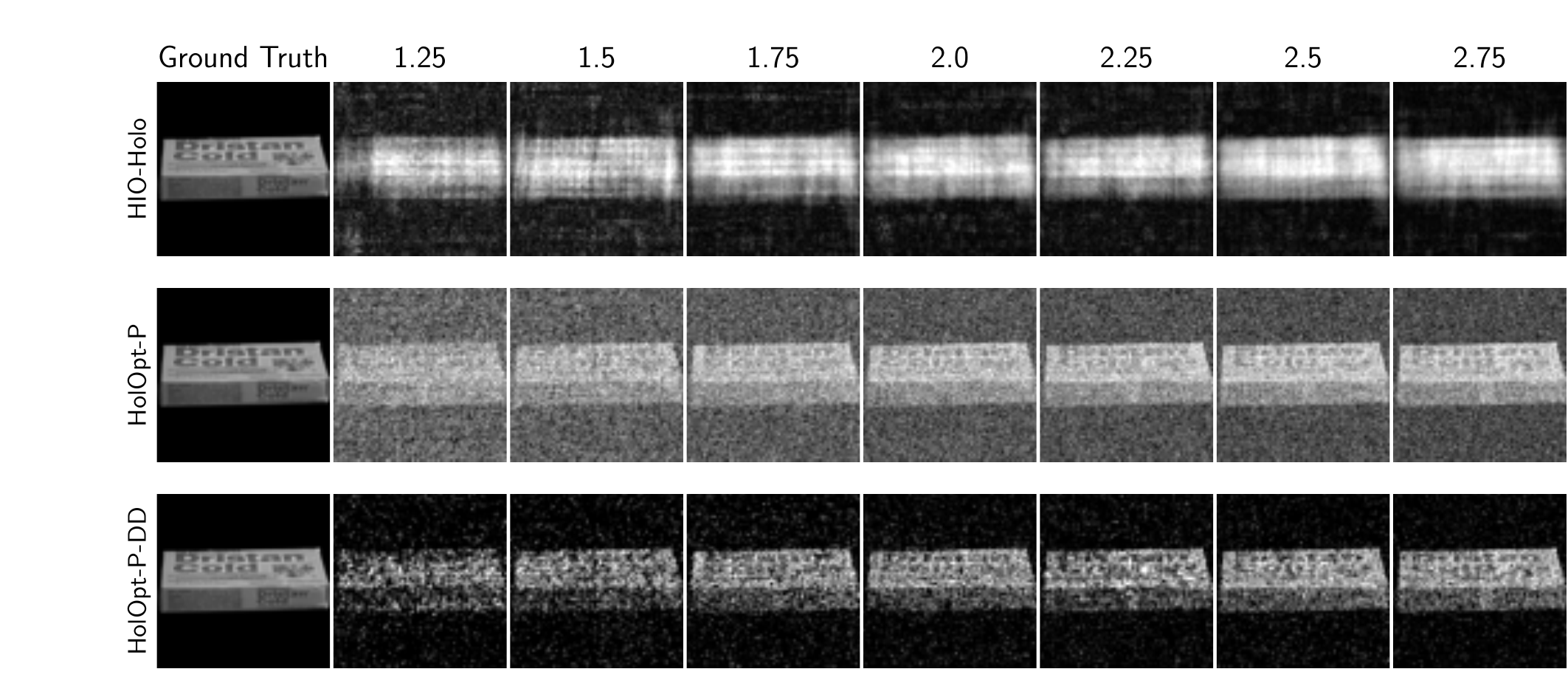}}
    {\includegraphics[width=0.49\textwidth]{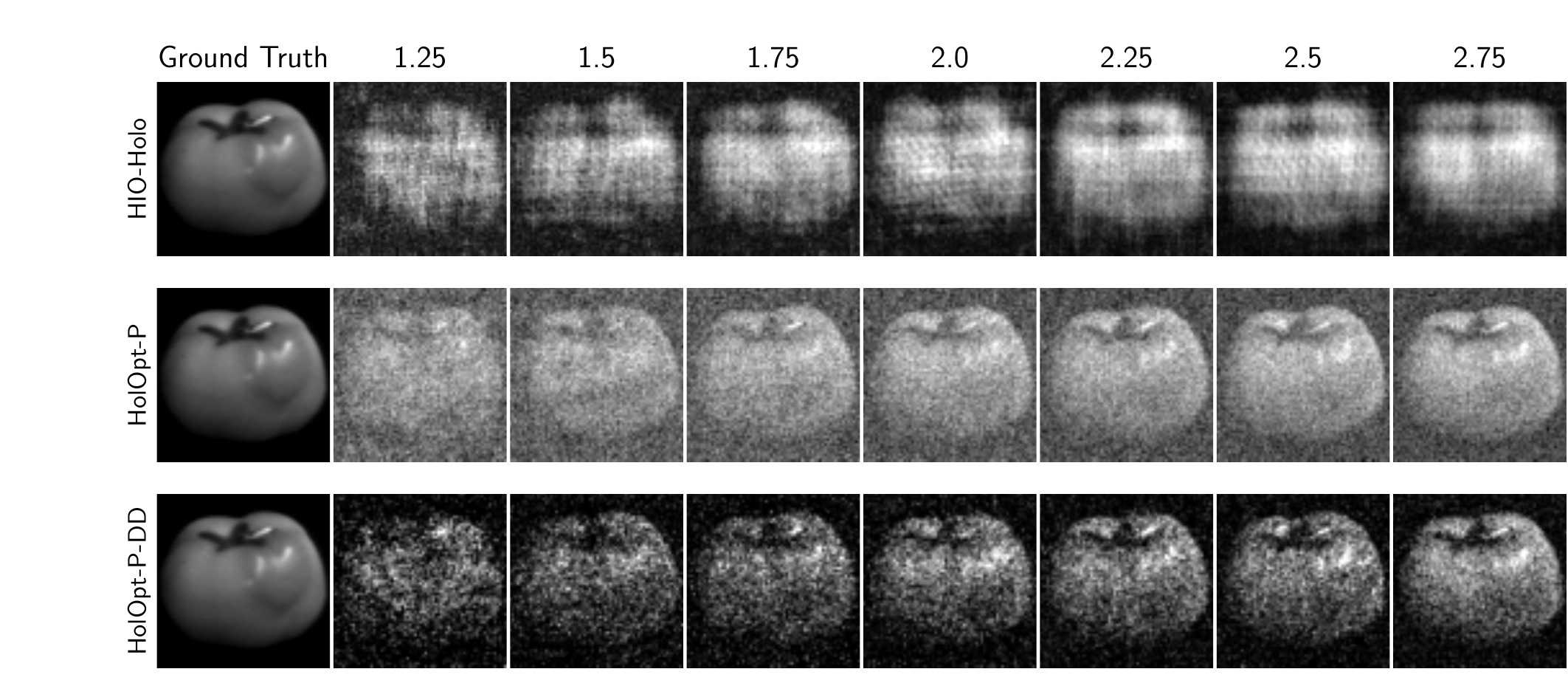}}
    \caption{Reconstructed images for SET12, BIO10 and COIL100 with varying oversampling factors (numbers above each column) at $N_p=1$ photon/pixel.}
    \label{fig:ovs-visuals-np1}
\end{figure}

\begin{figure}
    \centering
    \includegraphics[width=\textwidth]{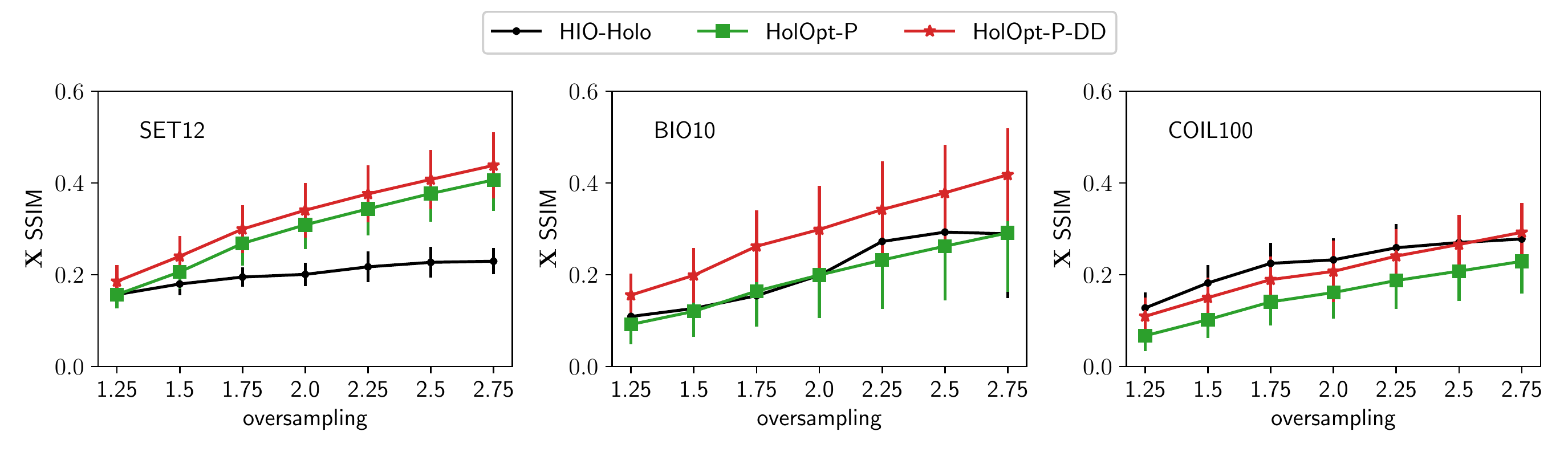}
    \includegraphics[width=\textwidth]{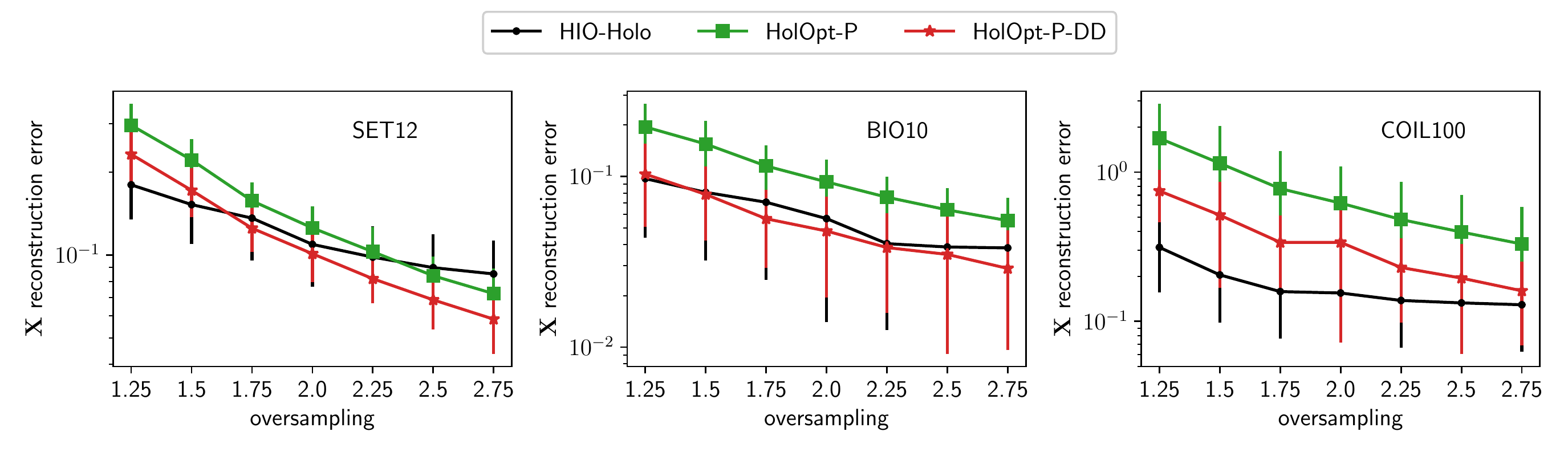}
    \caption{Reconstruction SSIM and $\ell2$-errors for SET12, BIO10 and COIL100 with varying oversampling factor at $N_p=1$ photon/pixel. Corresponding visuals are displayed in Figure \ref{fig:ovs-visuals-np1}.}
    \label{fig:ovs-graph-np1}
\end{figure}

\subsection{Oversampling experiment}
\label{app:ovs}
To complement results in the main text, Figure \ref{fig:ovs-graph-mse} displays reconstruction MSEs as a function of the oversampling rate as a function of the oversampling rate of the observations at $N_p = 10$.
Figure \ref{fig:ovs-visuals-np1} and Figure \ref{fig:ovs-graph-np1} focus on the noise level $N_p = 1$ photon / pixel and are similar to Figures already given at $N_p = 10$.

\end{document}